\documentclass{article}

     \PassOptionsToPackage{numbers, compress}{natbib}

\usepackage[final]{neurips_2019}

\usepackage[utf8]{inputenc} 
\usepackage[T1]{fontenc}    
\usepackage{hyperref}       
\usepackage{url}            
\usepackage{booktabs}       
\usepackage{amsfonts}       
\usepackage{nicefrac}       
\usepackage{microtype}      

\usepackage{graphicx}
\usepackage{subfigure}
\usepackage{multirow}
\usepackage{setspace}
\usepackage{amsmath,amssymb}
\usepackage{mathtools}
\usepackage{multicol}
\usepackage{float}
\usepackage{verbatim}
\usepackage{algorithm}
\usepackage{color}
\usepackage{algcompatible}
\usepackage[normalem]{ulem} 
\renewcommand{\algorithmiccomment}[1]{\bgroup\hfill\tiny//~#1\egroup}

\usepackage{symbols}
\usepackage{capt-of}
\usepackage{tabu}
\usepackage{tabularx}
\usepackage[font=small]{caption}

\title{Co-Generation with GANs using AIS based HMC}

\author{%
  Tiantian Fang  \\
  University of Illinois at Urbana-Champaign \\
  \texttt{tf6@illinois.edu} \\
  \And
  Alexander G. Schwing \\
  University of Illinois at Urbana-Champaign \\
  \texttt{aschwing@illinois.edu} 
}

\begin{document}

\maketitle

\begin{abstract}
Inferring the most likely configuration for a subset of variables of a joint distribution given the remaining ones -- which we refer to as co-generation -- is an important challenge that is computationally demanding for all but the simplest settings. This task has received a considerable amount of attention, particularly for classical ways of modeling distributions like structured prediction. In contrast, almost nothing is known about this task when considering recently proposed techniques for modeling high-dimensional distributions, particularly generative adversarial nets (GANs). Therefore, in this paper, we study the occurring challenges for co-generation with GANs. To address those challenges we develop an annealed importance sampling based Hamiltonian Monte Carlo co-generation algorithm. The presented approach significantly outperforms classical gradient based methods on a synthetic and on the CelebA and LSUN datasets. The code is available at \url{https://github.com/AilsaF/cogen_by_ais}.
\end{abstract}
\section{Introduction}
\label{sec:intro}

Finding a likely configuration for part of the variables of a joint distribution
given the remaining ones is a computationally
challenging problem with many applications in machine learning, computer vision
and natural language processing. 

Classical structured
prediction approaches~\cite{Lafferty2001,Taskar2003, Tsochantaridis2005} which
explicitly capture  correlations over an output space    of multiple
discrete random variables permit to formulate an energy function restricted to
the unobserved variables when conditioned on partly observed data. However, in many cases, it remains computationally
demanding to  find the most likely configuration or  to sample from the energy
restricted to the unobserved variables~\cite{Shimony1994,Valliant1979}. 

Alternatively, to model a joint probability distribution which implicitly
captures the correlations, generative adversarial nets
(GANs)~\cite{GoodfellowARXIV2014} and variational auto-encoders
(VAEs)~\cite{KingmaARXIV2013}  evolved as compelling tools  which exploit the
underlying manifold assumption: a latent `perturbation' is drawn from a simple
distribution which is subsequently transformed via a deep net
(generator/encoder) to the output space. Those methods have been used   for a
plethora of tasks, \eg, for domain
transfer~\cite{AnooshehARXIV2017,ChoiCVPR2018},
inpainting~\cite{PathakCVPR2016,YehChenCVPR2017}, image-to-image
translation~\cite{isola2017,LeeECCV2018,HuangECCV2018,LiuNIPS2017,RoyerARXIV2017,YiICCV2017,cyclegan,ZhuNIPS2017},
machine translation~\cite{conneau2017word} and health care~\cite{ShinARXIV2018}.

While GANs and VAEs permit easy sampling from the entire output space domain, it
also remains an open question of how to sample from part of the domain given the
remainder? We refer to this task as co-generation subsequently. 

Co-generation has been addressed in numerous works. For instance, for
image-to-image
translation~\cite{isola2017,LeeECCV2018,HuangECCV2018,LiuNIPS2017,RoyerARXIV2017,YiICCV2017,cyclegan,ZhuNIPS2017},
mappings between domains are learned directly via an encoder-decoder structure.
While such a formulation is convenient if we have two clearly separate domains,
this mechanism isn't scaleable if the number of  output space partitionings
grows, \eg, for image inpainting where missing regions are only specified at
test time.

To enable co-generation for a domain unknown at training time, for GANs,
optimization based algorithms have been proposed~\cite{YehChenCVPR2017,
LiuNIPS2016}. Intuitively, they aim at finding that latent sample that
accurately matches the observed part. \citet{dinhiclr2015} maximize the log-likelihood of the missing part given the observed one. However, we find that successful training of a GAN
leads to an increasingly ragged energy landscape, making the search for an
appropriate latent variable via back-propagation through the generator harder
and harder until it eventually fails.  

To deal with this ragged energy landscape for co-generation, we develop  an 
annealed importance sampling (AIS)~\cite{NealAIS2001} based Hamiltonian Monte Carlo (HMC)
algorithm~\cite{DuanePLB1987,NealHMC2010}. The proposed approach leverages the benefits of AIS, \ie, gradually
annealing a complex probability distribution, and HMC, \ie, avoiding a localized
random walk. 

We evaluate the proposed approach on synthetic data and imaging data (CelebA and LSUN), showing compelling results via MSE and MSSIM  metrics.
\vspace{-0.2cm}
\section{Related Work}
\label{sec:rel}
\vspace{-0.2cm}
In the following, we briefly discuss generative adversarial nets before providing
background on co-generation with adversarial nets.

\noindent\textbf{Generative adversarial nets} (GANs)~\cite{GoodfellowNIPS2014}
have originally been proposed as a non-cooperative two-player game, pitting a
generator against a discriminator. The discriminator is tasked to tell apart
real data from samples produced by the generator, while the generator is asked
to make differentiation for the discriminator as hard as possible. For a dataset
of samples $x\in\cX$ and random perturbations $z$ drawn from a simple
distribution, this intuitive formulation results in the saddle-point objective
$$
\max_\theta\min_w -\mathbb{E}_{{x}}[\ln D_w({x})] - \mathbb{E}_{{z}}[\ln (1-D_w(G_\theta({z})))],
$$
where $G_\theta$ denotes the generator parameterized by $\theta$ and $D_w$
refers to the discriminator parameterized by $w$. The discriminator assesses the probability of  its input argument
being real data. We let $\cX$ denote the output space. 
Subsequently, we refer to this formulation as the `Vanilla GAN,' and note that
its loss is related to the Jensen-Shannon divergence. Many other divergences and
distances have been proposed recently~\cite{arjovsky2017wasserstein,
li2017mmd,gulrajani2017improved,kolouri2017sliced,deshpande2018generative,
cully2017magan,mroueh2017mcgan,berthelot2017began,mroueh2017fisher,
lin2017pacgan,heusel2017gans,salimans2018improving} to improve the stability of the saddle-point
objective optimization during training and to address mode-collapse, some theoretically founded
and others empirically motivated. It is
beyond the scope to review all those variants.

\noindent\textbf{Co-generation}, is \emph{the task of obtaining a sample for a subset of the output space domain, given as input the remainder of the output
space domain}. This task is useful for applications like image
inpainting~\cite{PathakCVPR2016,YehChenCVPR2017}  or image-to-image
translation~\cite{isola2017,LeeECCV2018,HuangECCV2018,LiuNIPS2017,RoyerARXIV2017,YiICCV2017,cyclegan,ZhuNIPS2017}.
Many formulations for co-generation have been considered in the past. However,
few meet the criteria that \emph{any} given a subset of the output space could be
provided to generate the remainder.

Conditional GANs~\cite{MirzaARXIV2014} have been used to generate output space
objects based on a given input signal~\cite{WangCVPR2018}. The output space
object is typically generated as a whole and, to the best of our knowledge, no
decomposition into multiple subsets is considered.

Co-generation is  related to multi-modal Boltzmann
machines~\cite{SrivastavaNIPS2012,NgiamICML2011}, which learn a shared
representation for video and audio~\cite{NgiamICML2011} or image and
text~\cite{SrivastavaNIPS2012}. Restricted Boltzmann Machine based
encoder-decoder architectures are used to reconstruct either video/audio or
image/text given one of the representations. 
%
Co-generation is also related to deep net based joint embedding space
learning~\cite{KirosARXIV2014}. Specifically, a joint embedding of images and
text into a single vector space is demonstrated using deep net encoders. After performing vector operations in the embedding space, a new sentence can be
constructed using a decoder.  
%
%
Co-generation is also related to cross-domain image
generation~\cite{YimCVPR2015,ReedNIPS2015,DosovitskiyCVPR2015,AnooshehARXIV2017}.
Those techniques use  an encoder-decoder style deep net to transform rotation of
faces, to learn the transfer of style properties like rotation and translation to
other objects, or to encode class, view and transformation parameters into
images. 

Image-to-image translation is related in that a transformation between two
domains is learned either via an Image Transformation Net or an Encoder-Decoder
architecture. Early works in this direction tackled supervised image-to-image
translation~\cite{JohnsonECCV2016,isola2017,LedigCVPR2017,ChenICCV2017,LiangARXIV2017,ChoiCVPR2018}
followed by unsupervised
variants~\cite{TaigmanICLR2017,ShrivastavaCVPR2017,YiICCV2017,RoyerARXIV2017,BousmalisCVPR2017,WolfICCV2017,TauICLR2018,HoshenICLR2018}.
Cycle-consistency was discovered as a convenient regularization mechanism
in~\cite{KimICML2017,cyclegan,LiuNIPS2017,AlmahairiARXIV2018} and a distance
preserving regularization was shown by~\citet{BenaimNIPS2017}. Disentangling of
image representations was investigated recently~\cite{HuangECCV2018,LeeECCV2018}
and ambiguity in the task was considered by \citet{ZhuNIPS2017}. Other
losses such as a `triangle formulation' have been investigated
in~\cite{GanNIPS2017,LiXuNIPS2017}.

Attribute transfer~\cite{LaffontTOG2014}, analogy
learning~\cite{HertzmannSIGGRAPH2001,ReedNIPS2015} and many style transfer
approaches~\cite{TenenbaumNIPS1997,BousmalisNIPS2016,VillegasICLR2017,DentonNIPS2017,MathieuNIPS2016,UlyanovICML2016,HuangICCV2017,LiFangNIPS2017,LiECCV2018,TulyakovCVPR2018,DonahueICLR2018,ShenNIPS2017,DumoulinICLR2017,GhiasiBMVC2017,ZhuECCV2016}
just like feature learning via inpainting~\cite{PathakCVPR2016} are also using
an encoder-decoder formulation, which maps entire samples from one domain to
entire samples in another domain.

Co-generation is at least challenging if not impossible for all the
aforementioned works since decoders need to be trained for every subset of the
output space domain. This is not scalable unless we know ahead of time the few
distinct subsets of interest. 

Hence, to generate arbitrary sub-spaces, other techniques need to be considered.
Some applicable exceptions from the encoder-decoder style training are work on
style transfer by \citet{GatysCVPR2016},  work on image inpainting
by \citet{YehChenCVPR2017}, and coupled generative adversarial nets
(CoGANs) by \citet{LiuNIPS2016}. In all three formulations, a loss is optimized  
to match observations to parts of the generated data 
by iteratively computing gradient updates for a latent space sample. 
%
In particular, \citet{LiuNIPS2016} learn a joint distribution over
multiple domains by coupling multiple generators and possibly discriminators via
weight-sharing. 
\citet{LiuNIPS2016}
briefly discuss co-generation when talking about ``cross-domain image
transformation,'' report to observe coverage issues and state that they leave a
detailed study to ``future work.'' Instead of an optimization based procedure, we
propose to use an annealed importance sampling based  Hamiltonian Monte Carlo
approach. We briefly review both techniques subsequently.

\noindent\textbf{Annealded importance sampling (AIS)}~\cite{NealAIS2001} is an algorithm
typically used to estimate (ratios of) the partition function~\cite{Wuiclr2017,dickhais}. Specifically, it
gradually approaches the partition function for a distribution of interest by
successively refining samples which were initially obtained from a `simple'
distribution, \eg, a multivariate Gaussian. 
Here, we are not interested in the partition function itself, but rather in the
ability of AIS to accurately draw samples from complex distributions, which makes AIS a great tool for co-generation. 

\noindent\textbf{Hamiltonian Monte Carlo (HMC)}~\cite{DuanePLB1987} originally
referred to as ``Hybrid Monte Carlo'' united the Markov Chain Monte
Carlo~\cite{MetropolisCP1953} technique with molecular dynamics
approaches~\cite{Alder1959}. Early on they were used for neural net
models~\cite{NealHMCNeuralNets} and a seminal review by~\citet{NealHMC2010}
provides a detailed account. In short, a Hamiltonian combines the potential
energy, \ie, the log probability that we are interested in sampling from with auxiliary kinetic energy. The latter typically follows a Gaussian distribution.
HMC alternates updates for the kinetic energy with Metropolis updates computed
by following a trajectory of constant value along the Hamiltonian to compute a
new proposal. HMC is useful for co-generation because of its reduced random-walk behavior as we will explain next.

\vspace{-0.2cm}
\section{AIS based HMC for Co-Generation}
\label{sec:app}
\vspace{-0.2cm}
In the following we first motivate the problem of co-generation before we
present an overview of our proposed approach and discuss the details of the
employed Hamiltonian Monte Carlo method.

\vspace{-0.2cm}
\subsection{Motivation}
\vspace{-0.2cm}

\begin{figure}[t]
\vspace{-0.2cm}
\begin{center}
\setlength{\tabcolsep}{0pt}

\begin{tabular}{cccc}
  \multicolumn{1}{c}{{\small 500th iteration}} & \multicolumn{1}{c}{{\small 1500th iteration}} & \multicolumn{1}{c}{{\small 2500th iteration}} & \multicolumn{1}{c}{{\small 15000th iteration}} 
\\
  \includegraphics[width=.2\textwidth]{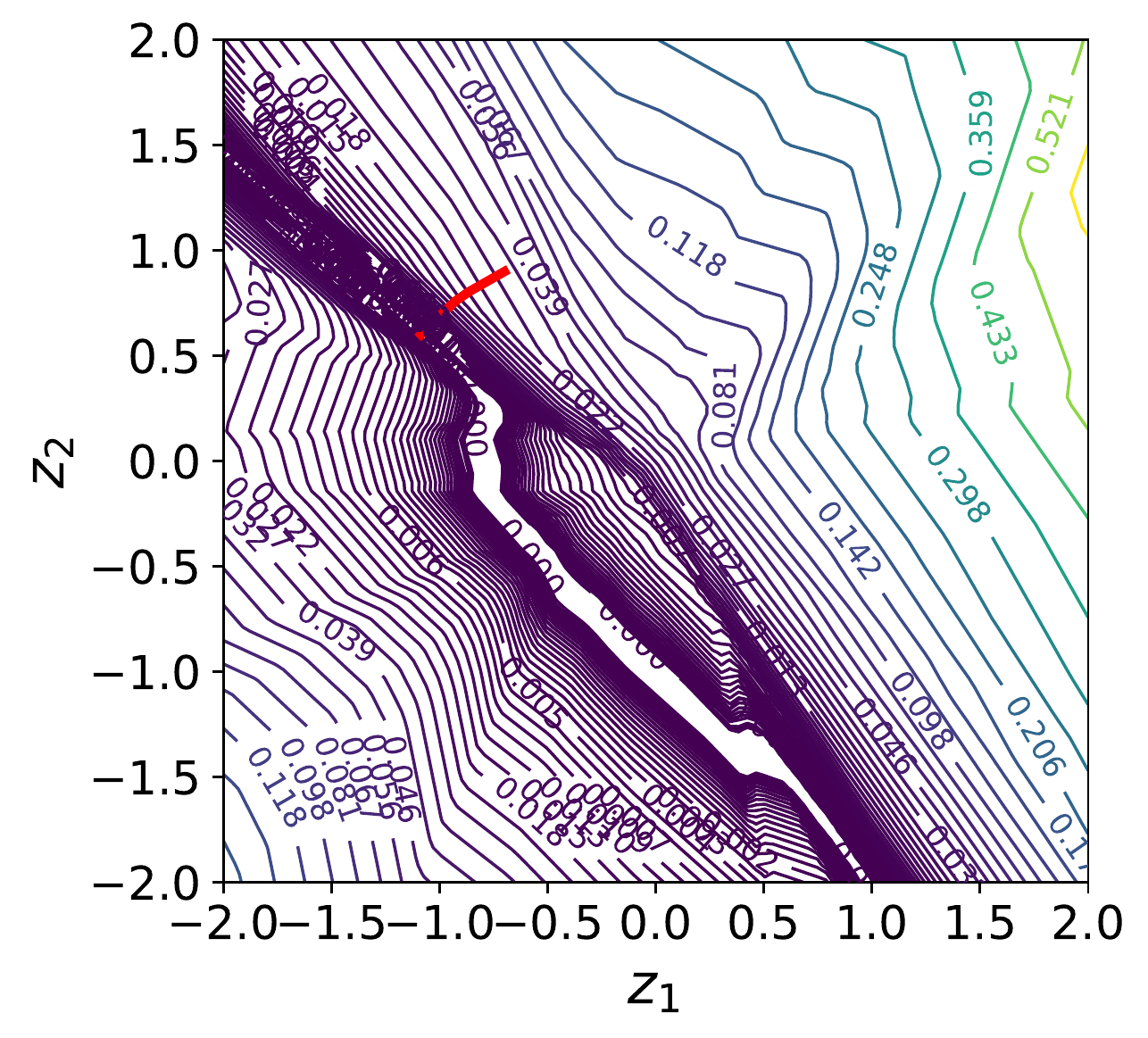}&\includegraphics[width=.2\textwidth]{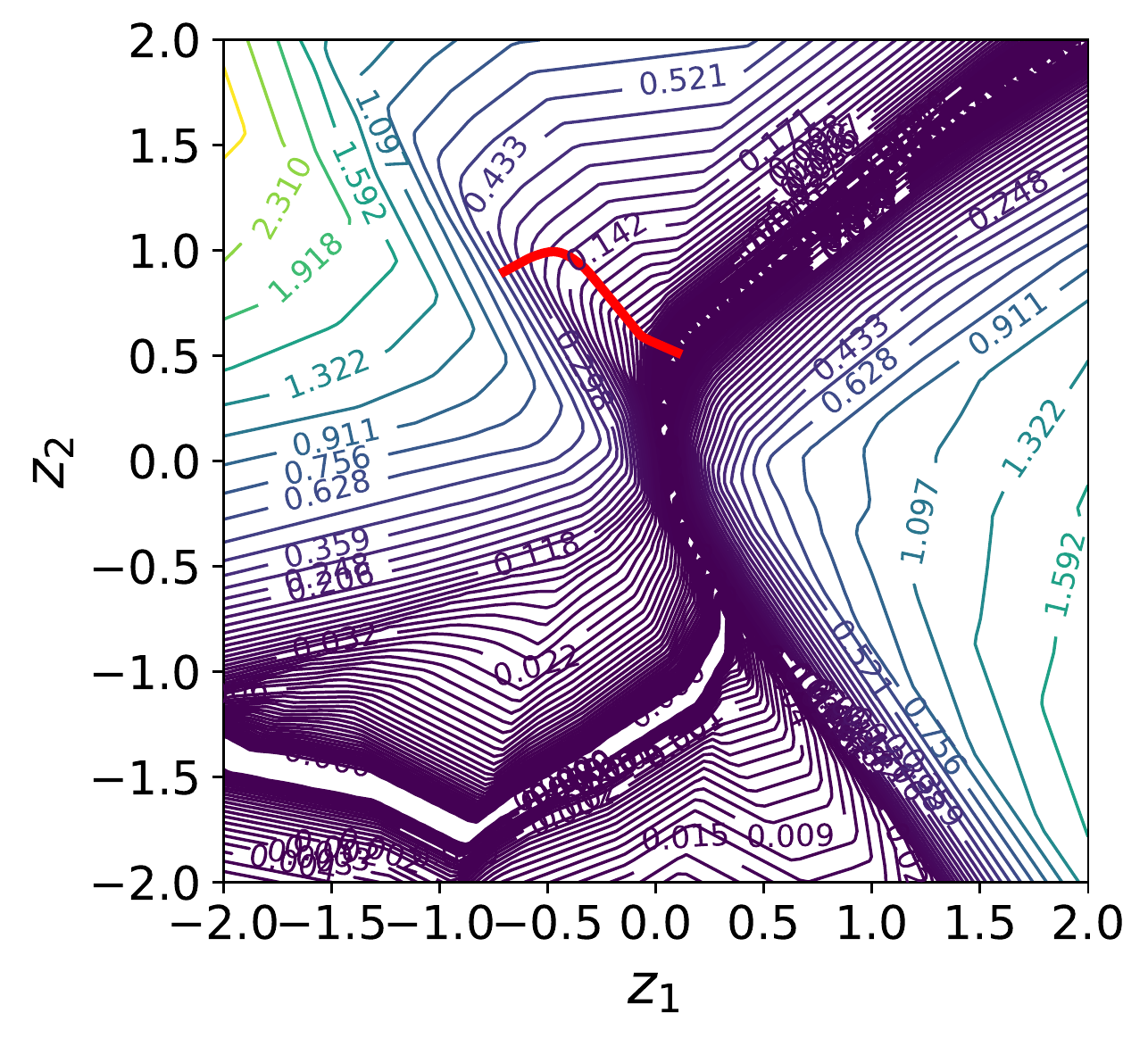}&\includegraphics[width=.2\textwidth]{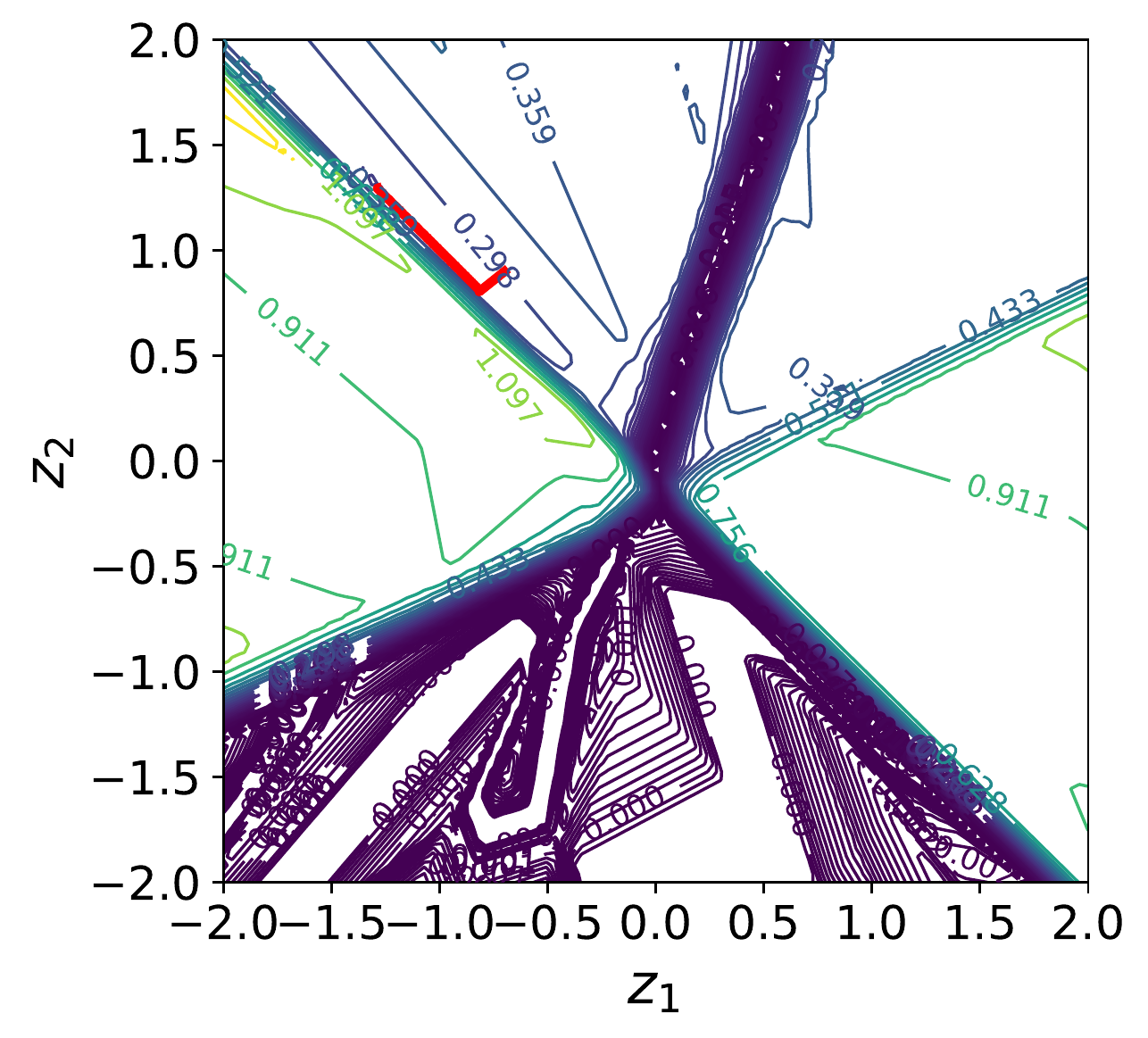}&\includegraphics[width=.2\textwidth]{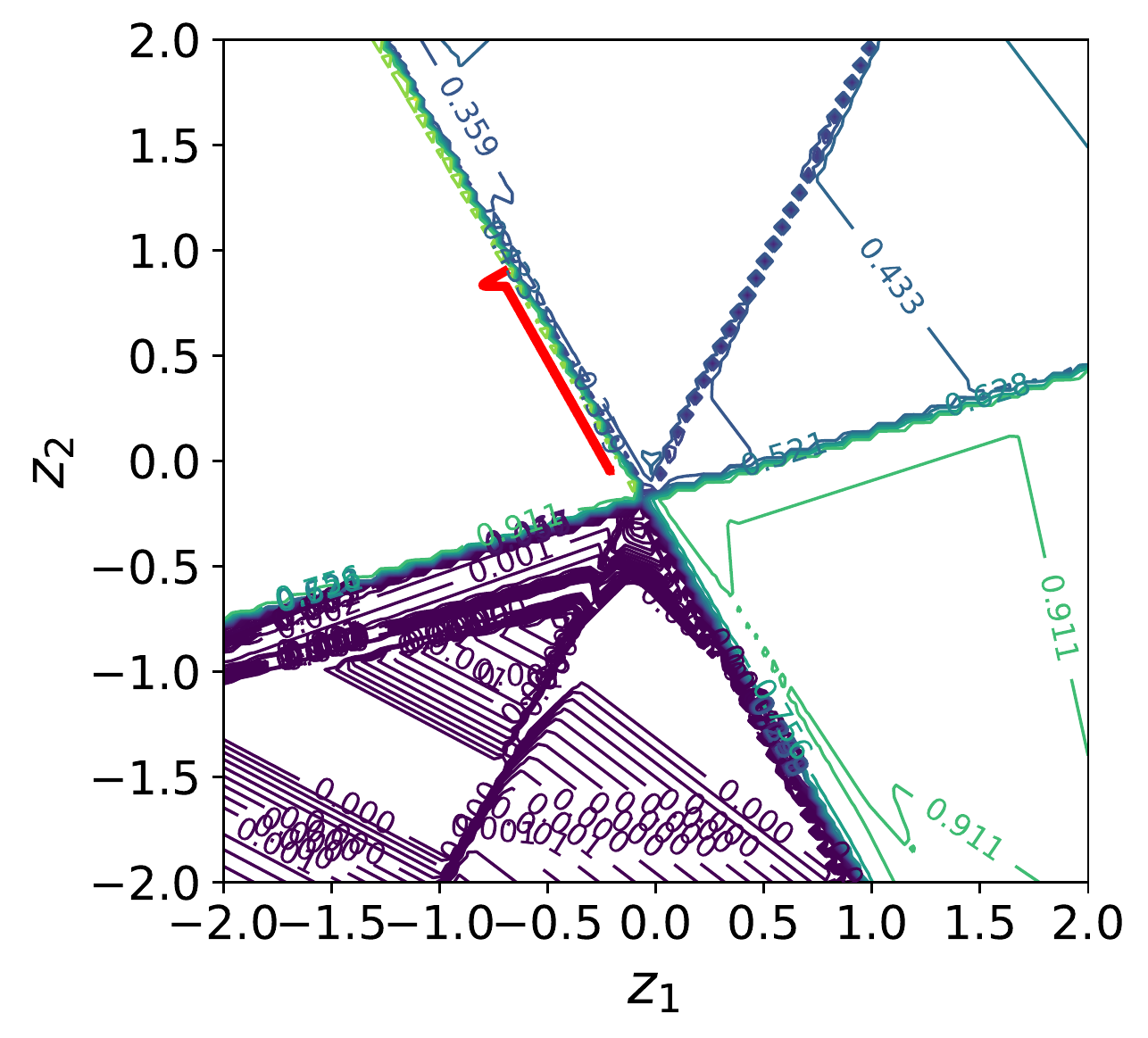} \\
   \includegraphics[width=.2\textwidth]{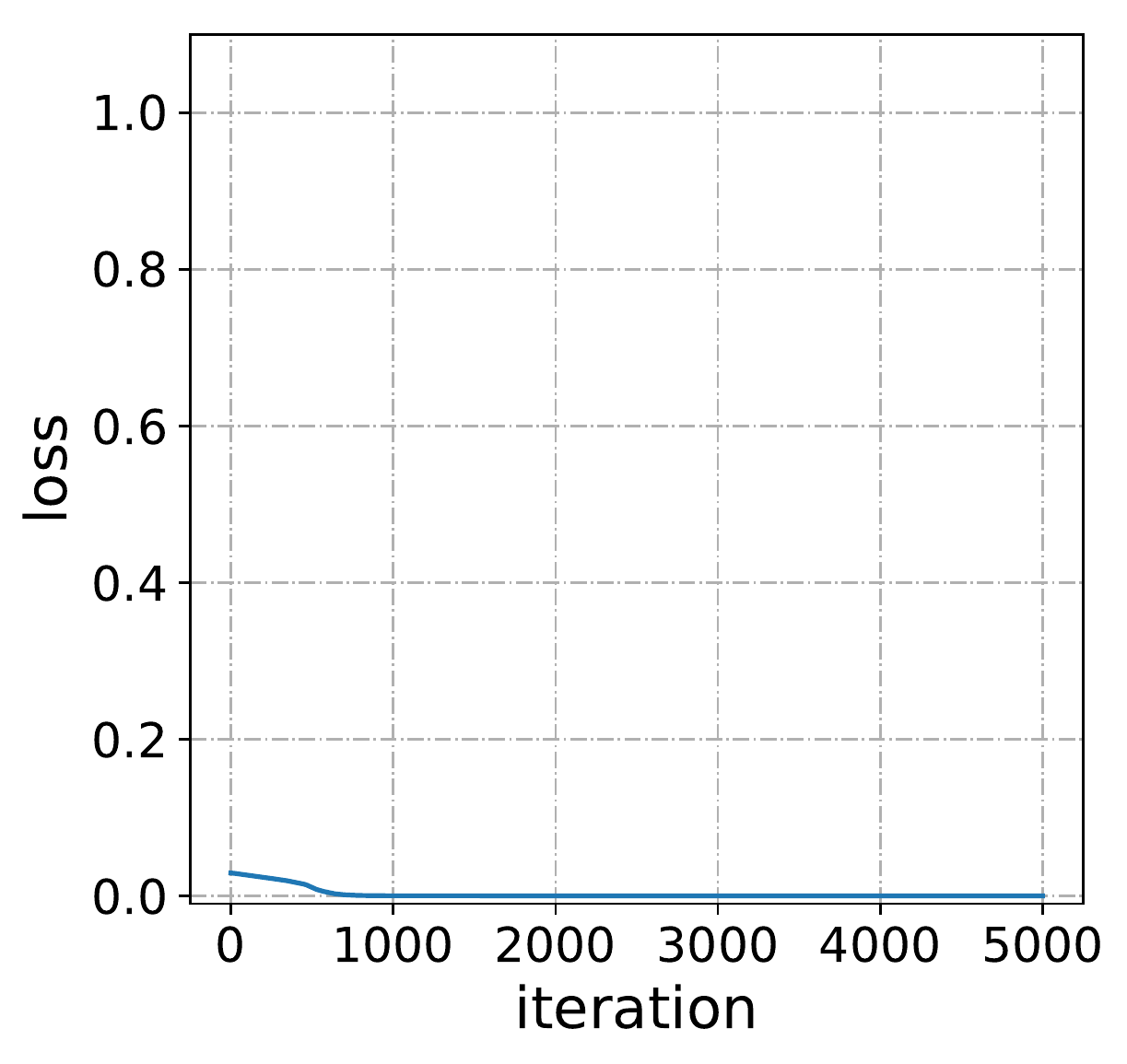}&\includegraphics[width=.2\textwidth]{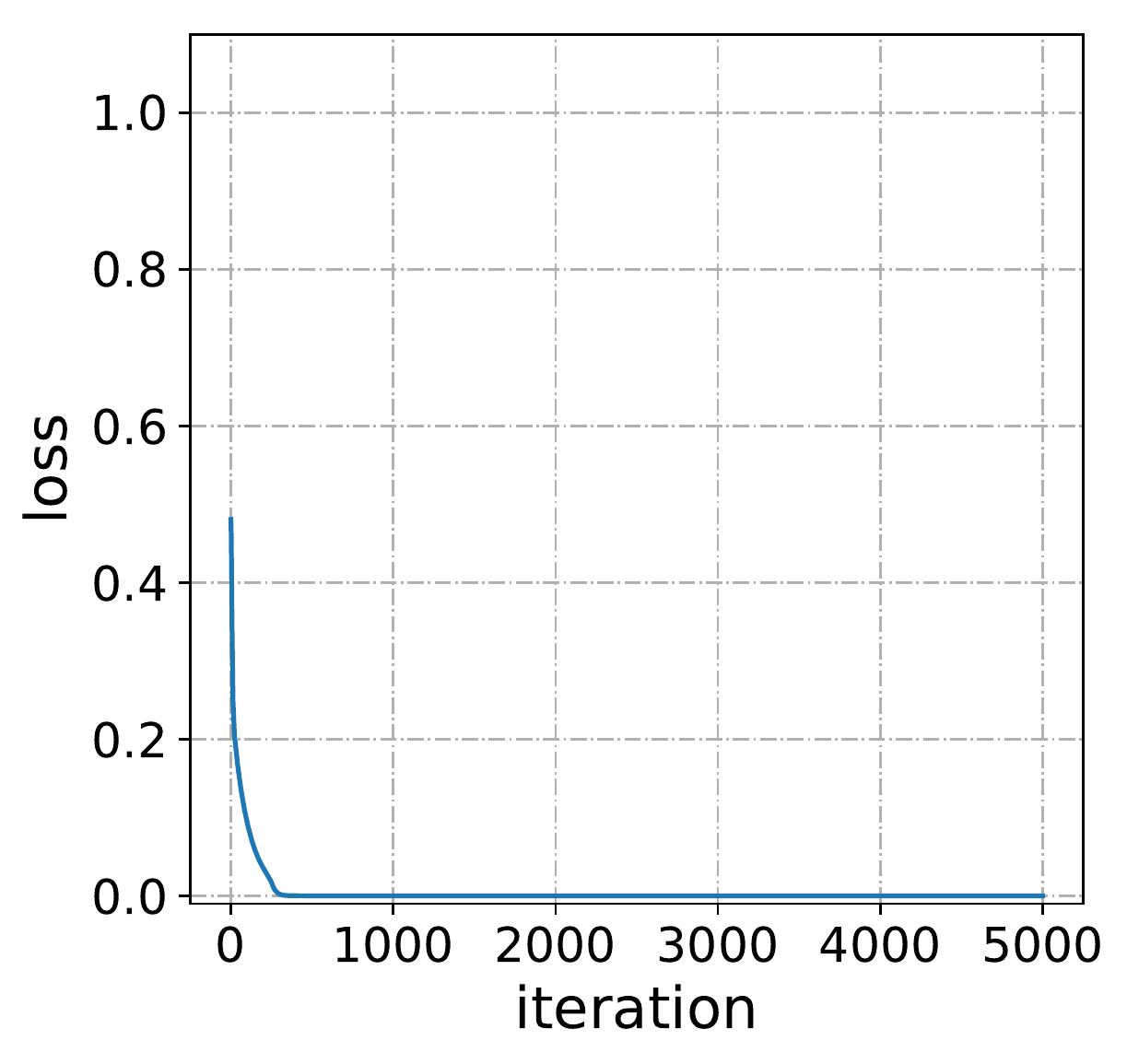}&\includegraphics[width=.2\textwidth]{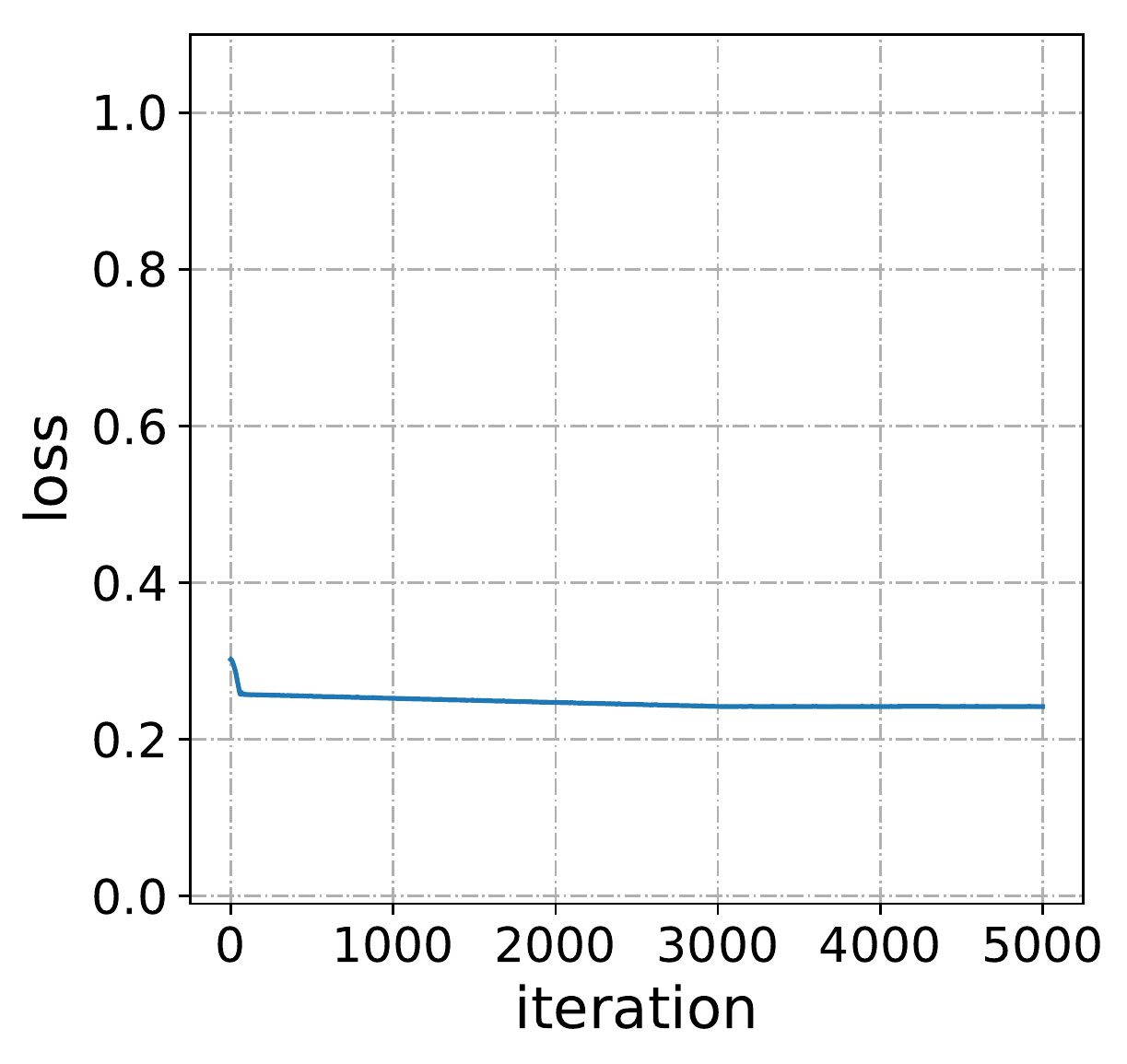}&\includegraphics[width=.2\textwidth]{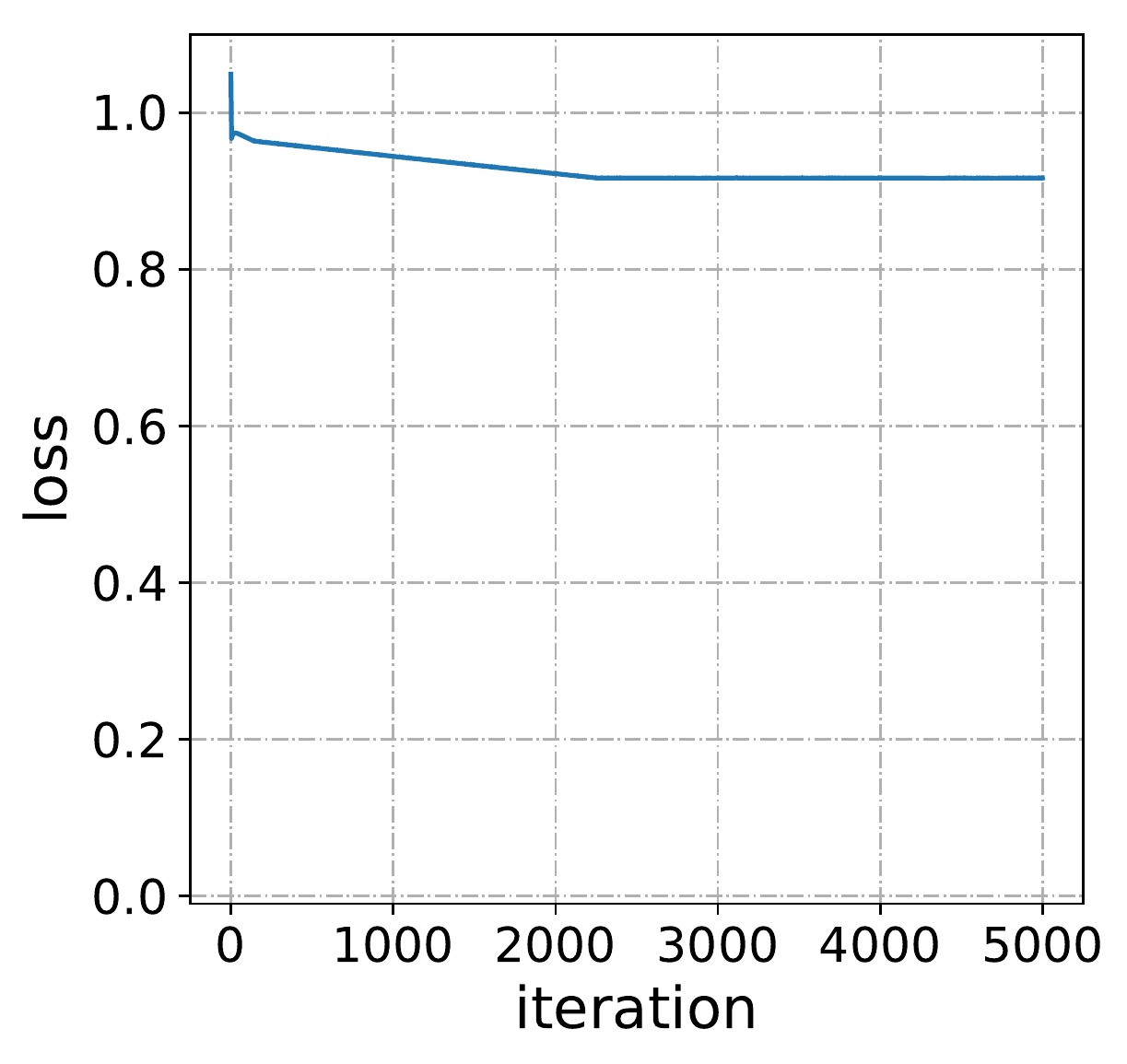} \\
\end{tabular}
\end{center}
\vspace{-0.4cm}
\caption{Vanilla GAN loss in $\cZ$ space (top) and gradient descent (GD) reconstruction error for 500, 1.5k, 2.5k and 15k generator training epochs.}
\label{fig:toysgddiff}
\vspace{-0.7cm}
\end{figure}

Assume we are given a well trained generator $\hat x = G_\theta(z)$, parameterized by $\theta$, which is
able to produce samples $\hat x$ from an implicitly modeled distribution
$p_G(x|z)$ via a transformation of embeddings $z$~\cite{GoodfellowARXIV2014,arjovsky2017wasserstein,LiNIPS2017,deshpande2018generative,IDeshpandeCVPR2019}. Further assume we are given partially observed data $x_o$ while the
remaining part $x_h$ of the data $x = (x_o, x_h)$ is latent, \ie, hidden. Note that during  training of the generator parameters $\theta$ we don't assume information about which part of the data is missing to be available. 

To reconstruct the latent parts of the data $x_h$ from available observations $x_o$, a program is often formulated as follows:
\be
z^\ast = \arg\min_z \|x_o - G_\theta(z)_o\|_2^2,
\label{eq:backprop}
\ee
where $G_\theta(z)_o$ denotes the restriction of the generated sample $G_\theta(z)$ to the observed part. We focus on the $\ell_2$ loss here but note that any other function measuring the fit of $x_o$ and $G_\theta(z)_o$ is equally applicable. Upon solving the program given in \equref{eq:backprop}, we easily obtain an estimate for the missing data $\hat x_h = G(z^\ast)_h$. 

Although the program given in \equref{eq:backprop} seems rather straightforward, it turns out to be really hard to solve, particularly if the generator $G_\theta(z)$ is very well trained. To see this, consider as an example a generator operating on a 2-dimensional latent space $z = (z_1, z_2)$ and 2-dimensional data $x = (x_1, x_2)$ drawn from a mixture of five equally weighted Gaussians with a variance of $0.02$, the means of which are spaced equally on the unit circle. For this example we use $h = 1$ and let $x_o = x_2 = 0$.
In the first row of \figref{fig:toysgddiff} we illustrate the loss surface of the objective given in \equref{eq:backprop} obtained when using a generator $G_\theta(z)$ trained on the original 2-dimensional data for 500, 1.5k, 2.5k, and 15k iterations (columns in \figref{fig:toysgddiff}). 

Even in this simple 2-dimensional setting, we observe the latent space to become increasingly ragged, exhibiting folds that clearly separate different data regimes. First or second order optimization techniques cannot cope easily with such a loss landscape and likely get trapped in local optima. To illustrate this we highlight in \figref{fig:toysgddiff} (first row) the trajectory of a sample $z$ optimized via gradient descent (GD) using red color and provide the corresponding loss over the number of GD updates for the objective given in \equref{eq:backprop} in \figref{fig:toysgddiff} (second row). We observe optimization to get stuck in a local optimum as the loss fails to decrease to zero once the generator better captures the data.

To prevent those local-optima issues for co-generation, we propose an annealed importance-sampling (AIS) based Hamiltonian Monte Carlo (HMC) method in the following. 

\vspace{-0.2cm}
\subsection{Overview}
\vspace{-0.2cm}
 In
order to reconstruct the hidden portion $x_h$ of the data $x = (x_o, x_h)$ we are
interested in drawing samples $\hat z$ such that $\hat x_o = G_\theta(\hat z)_o$
has a high probability under $\log p(z|x_o) \propto -\|x_o -
G_\theta(z)_o\|_2^2$. Note that  the proposed approach is not restricted to this
log-quadratic posterior $p(z|x_o)$ just like the objective in \equref{eq:backprop} is not restricted to the $\ell_2$ norm.

To obtain samples $\hat z$ following the posterior distribution $p(z|x_o)$, the sampling-importance-resampling framework provides a
mechanism which only requires access to samples and doesn't need computation of
a normalization constant. Specifically, for sampling-importance-resampling, we
first draw latent points  $z \sim p(z)$ from a simple prior distribution $p(z)$, \eg, a Gaussian. 
We then compute weights according to $p(z|x_o)$ in a second step and finally resample
in a third step from the originally drawn set according to the computed weights.

However, sampling-importance-resampling is particularly challenging in even modestly high-dimensional settings since many samples are required to adequately cover
the  space to a reasonable degree. As expected, empirically, we found this
procedure to not work very well. To address this concern, here, we propose an
annealed importance sampling (AIS) based Hamiltonian Monte Carlo (HMC)
procedure. Just like sampling-importance-resampling, the proposed approach
only requires access to samples and no normalization constant needs to be computed. 

More specifically, we use annealed importance sampling to gradually approach
the complex and often high-dimensional posterior distribution $p(z|x_o)$ by simulating a Markov Chain starting from
the prior distribution $p(z) = \cN(z|0,I)$, a standard normal distribution with
zero mean and unit variance. With the increasing number of updates, we gradually approach the true posterior. 
Formally, we define an annealing schedule for the
parameter $\beta_t$ from $\beta_0 = 0$ to $\beta_T = 1$. At every time step $t
\in \{1, \ldots, T\}$ we
refine the  samples drawn at the previous timestep $t-1$ so as to represent the
distribution $\hat p_t(z|x_o) = p(z|x_o)^{\beta_t}p(z)^{1-\beta_t}$. 
Intuitively and
following the spirit of annealed importance sampling, it is easier to gradually
approach sampling from $p(z|x_o) = \hat p_T(z|x_o)$ by successively refining the
samples. Note the notational difference between the posterior of interest $p(z|x_o)$, and the annealed posterior $\hat p_t(z|x_o)$.

To successively refine the samples we use Hamilton Monte Carlo (HMC) sampling
because a proposed update can be far from the current sample while still having
a high acceptance probability. Specifically, HMC  enables  to bypass to some extent
slow exploration of the space when using classical Metropolis updates based on a
random walk proposal distribution.

\begin{figure}[t]
    \begin{minipage}[t]{\linewidth}
    \begin{algorithm}[H]
        \caption{AIS based HMC}
            \label{alg:ours}
    \begin{algorithmic}[1]
        \STATE {\bfseries Input:} $p(z|x_o)$, $\beta_t$ $\forall t\in\{1,
        \ldots, T\}$
        \STATE Draw set of samples $z \in \cZ$ from prior distribution $p(z)$
        \FOR[AIS loop]{$t = 1, \ldots, T$}
            
            \STATE Define $\hat p_t(z|x_o) = p(z|x_o)^{\beta_t}p(z)^{1-\beta_t}$ 
            \FOR[HMC loop]{$m = 1, \ldots, M$}

                \STATE $\forall
                z\in \cZ$ initialize Hamiltonian and momentum variables $v\sim\cN(0,I) $
                \STATE $\forall
            z\in \cZ$ compute new proposal sample using leapfrog integration on Hamiltonian
            \STATE $\forall z\in \cZ$ use Metropolis Hastings  to check whether to accept the proposal and
            update $\cZ$ 
            \ENDFOR
        \ENDFOR
        \STATE {\bfseries Return:} $\cZ$
    \end{algorithmic}
    \end{algorithm}
\end{minipage}
\vspace{-0.6cm}
\end{figure}

Combining both AIS and HMC, the developed approach summarized in
\algref{alg:ours} iteratively proceeds as follows after having drawn initial
samples from $p(z) = \hat p_0(z|x_o)$: (1) define the desired proposal distribution;
and (2) for $K$ iterations compute new proposals using leapfrog integration and
check whether to replace the previous sample with the new proposal. 
Subsequently, we discuss how to compute proposals and how to check acceptance.

\vspace{-0.2cm}
\subsection{Hamilton Monte Carlo}
\vspace{-0.2cm}

Hamilton Monte Carlo (HMC) explores the latent space much more quickly than a classical random walk algorithm. 
Moreover,  HMC methods are particularly suitable for co-generation because
they are capable of traversing folds in an energy landscape. 
To this end, HMC methods
trade potential energy $U_t(z) = -\log \hat p_t(z|x_o)$ with kinetic energy $K_t(v)$. 
Hereby the dimension $d$ of the momentum variable $v\in\mathbb{R}^d$ is
identical to that of the latent samples $z\in\mathbb{R}^d$. 
For readability, we drop the dependence on the time index $t$ from here on. 

Specifically, HMC defines a Hamiltonian $H(z,v) = U(z) + K(v)$ or conversely
a joint probability distribution $\log p(z,v) \propto -H(z,v)$ and proceeds by
iterating 
three steps $M$ times.

In a first step, the Hamiltonian is initialized by randomly
sampling the momentum variable $v$, typically using a standard Gaussian. Note
that this step leaves the joint distribution $p(z,v)$ corresponding to the Hamiltonian invariant as the momentum $v$
is independent of samples $z$ and as we sample from the correct pre-defined
distribution for the momentum variables. 

In a second step, we compute proposals $(z^\ast, v^\ast)$ via leapfrog
integration to move along a hypersurface of the Hamiltonian, \ie, the value of
the Hamiltonian does not change. However note, in this step, kinetic energy
$K(v)$ can be traded for potential energy $U(z)$ and vice versa.  

In the final third step we decide whether to accept the proposal $(z^\ast,
v^\ast)$ computed via leapfrog integration. Formally, we accept the proposal
with probability
\be
\min\{1, \exp\left(-H(z^\ast, v^\ast) + H(z, v)\right)\}.
\label{eq:acc}
\ee
If the proposed state $(z^\ast, v^\ast)$ is rejected, the $m+1$-th iteration
reuses $z$, otherwise $z$ is replaced with $z^\ast$ in the $m+1$-th iteration. 

Note that points $(z,v)$ with different probability density are only obtained
during the first step, \ie, sampling of the moment variables $v$. Importantly, resampling of $v$ can change the probability
density by a large amount. As evident from \equref{eq:acc}, a low value for the
Hamiltonian obtained after resampling $v$ increases the chances of accepting
this proposal, \ie, we gradually increase the number of samples with a low value for the Hamiltonian, conversely a high probability. 

\vspace{-0.2cm}
\subsection{Implementation Details}
\vspace{-0.2cm}
We use a sigmoid schedule for the parameter $\beta_t$, \ie, we linearly space $T-1$ temperature values within a range and apply a sigmoid function to these values to obtain $\beta_t$. This schedule, emphasizes locations where the distribution changes drastically. We use $0.01$ as the leapfrog step size and employ $10$ leapfrog updates per HMC loop for the synthetic 2D dataset and $20$ leapfrog updates for the real dataset at first. The acceptance rate is $0.65$, as recommended by \citet{NealHMC2010}. Low acceptance rate means the leapfrog step size is too large in which case the step size will be decreased by 0.98 automatically. In contrast, a high acceptance rate will increase the step size by 1.02\footnote{We  adapt the AIS implementation from \url{https://github.com/tonywu95/eval_gen}}.


\begin{figure}[t]
\centering
 \begin{tabularx}{\textwidth} {c@{\hskip1pt}c@{\hskip3pt}c@{\hskip3pt}c@{\hskip3pt}c@{\hskip3pt}c}

  \rotatebox{90}{\hspace{0.8cm}{\small 500}} & \includegraphics[width=25mm]{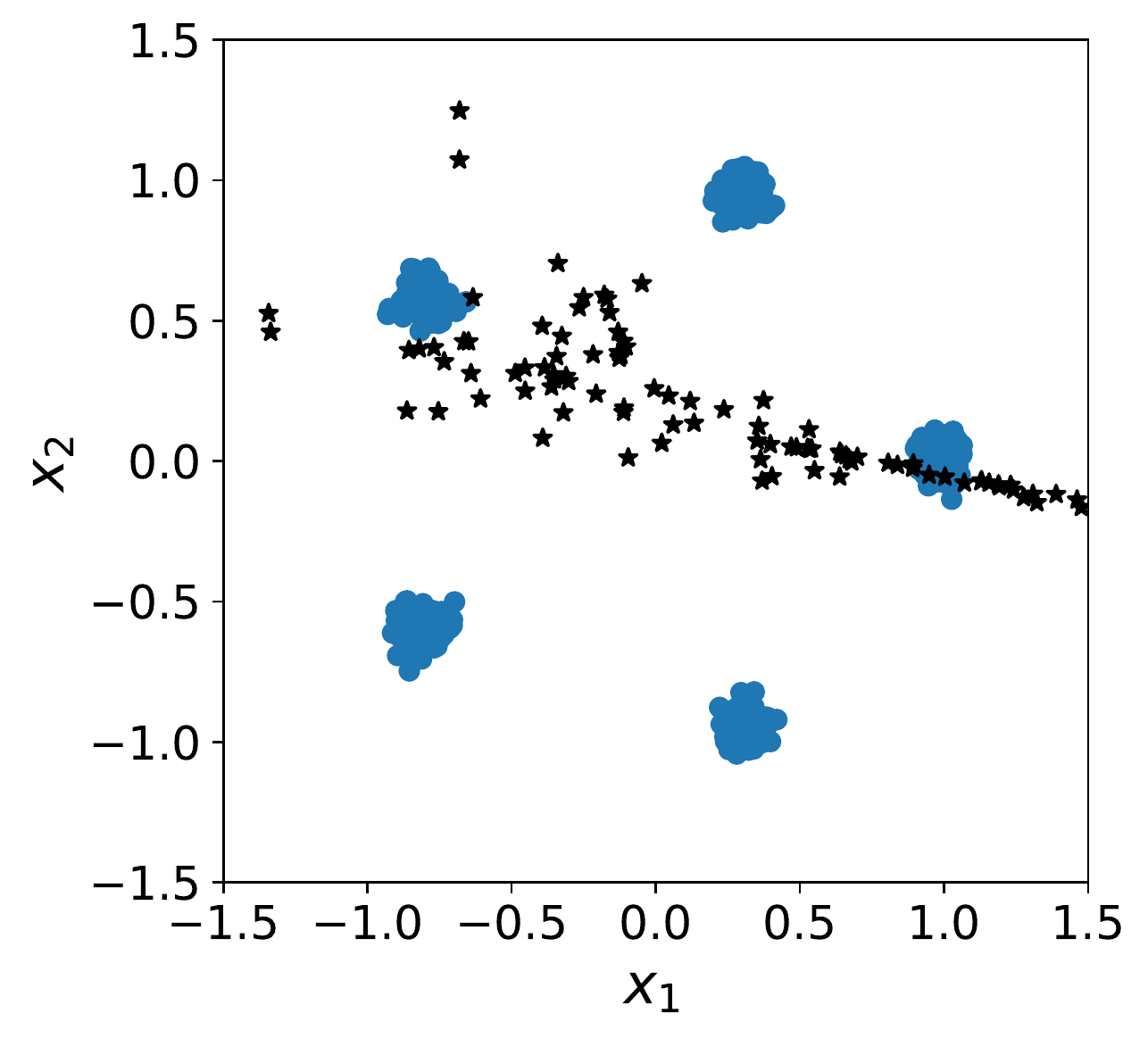}&\includegraphics[width=25mm]{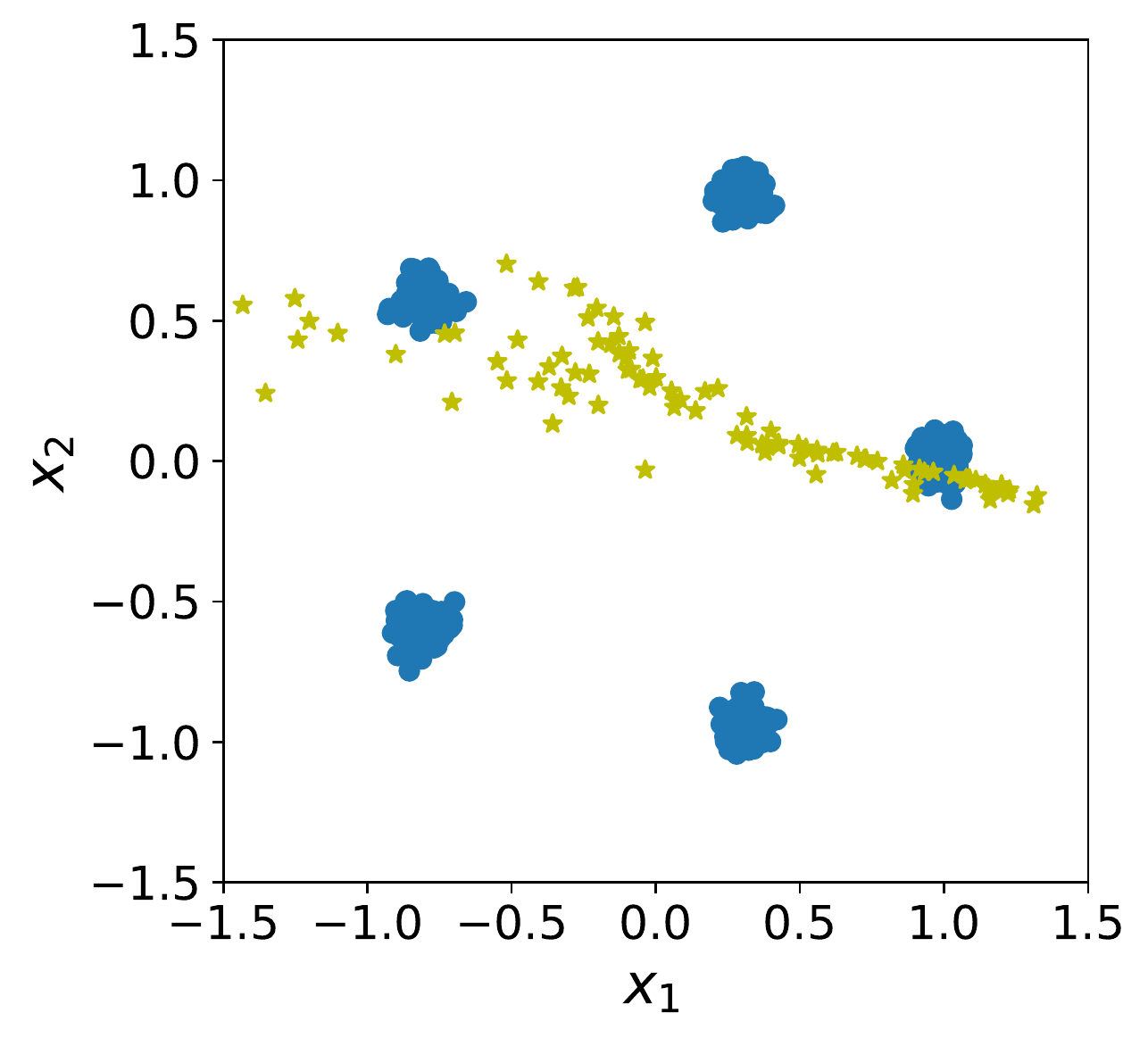}&\includegraphics[width=25mm]{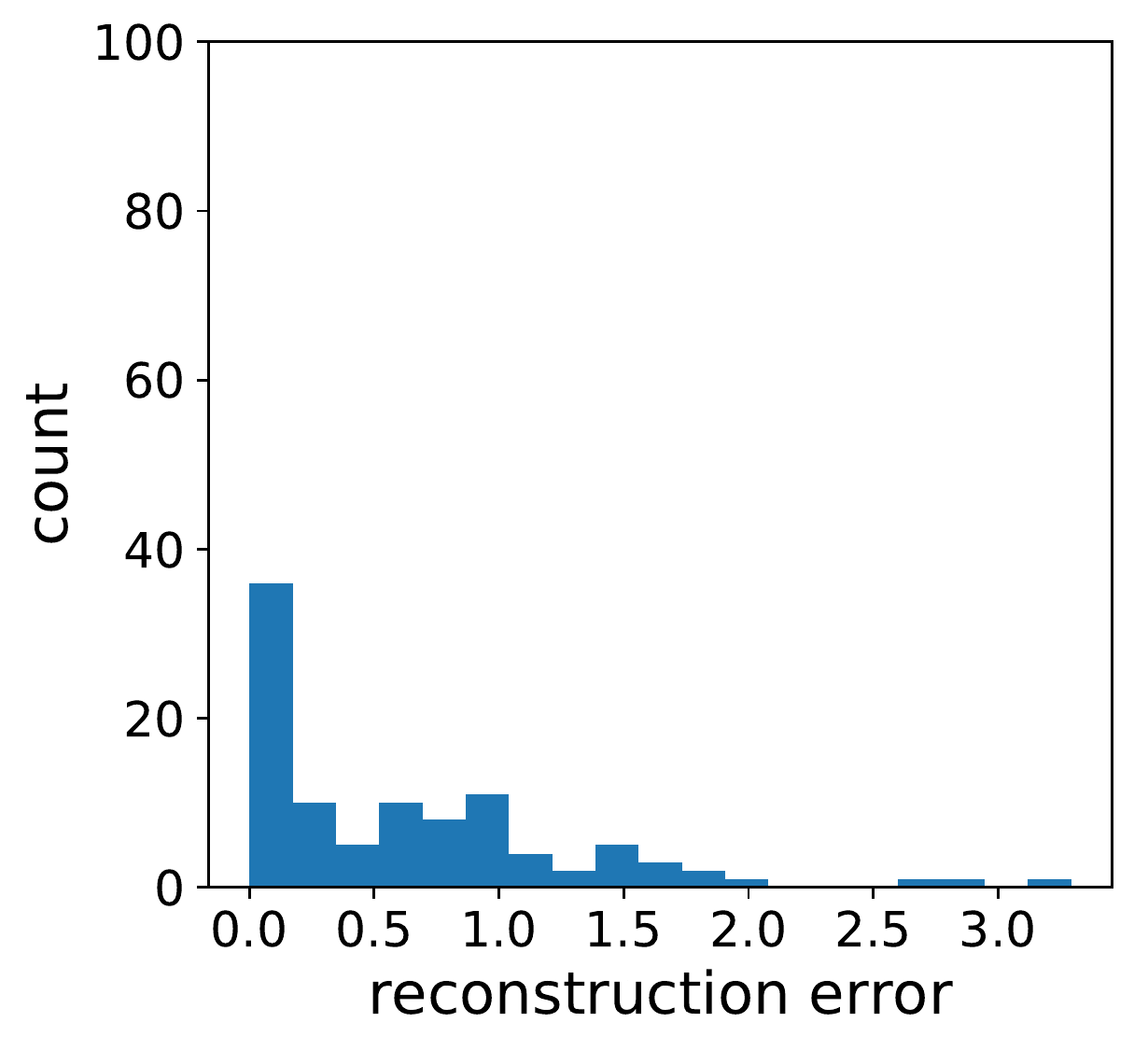}&\includegraphics[width=25mm]{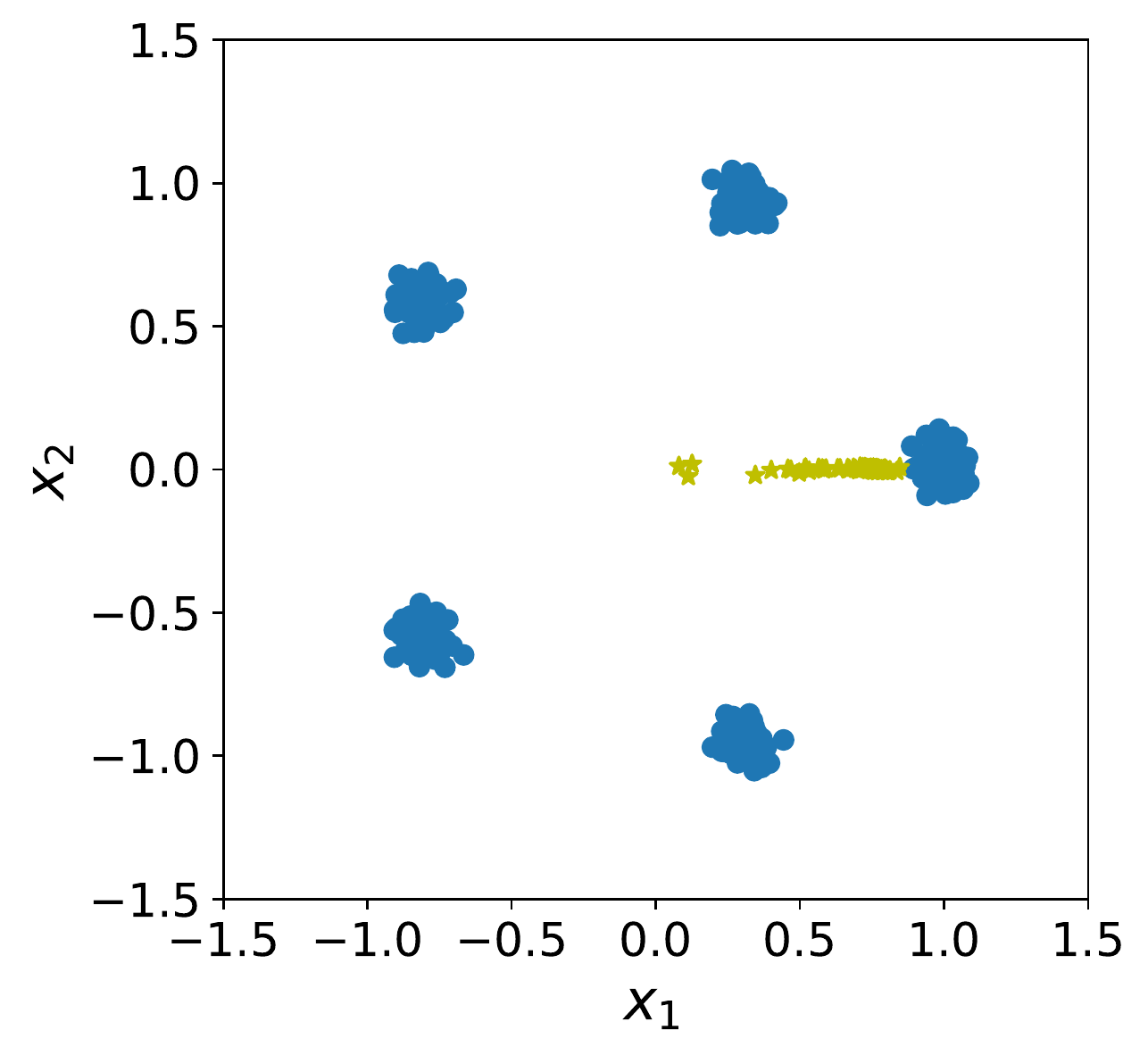}& \includegraphics[width=25mm]{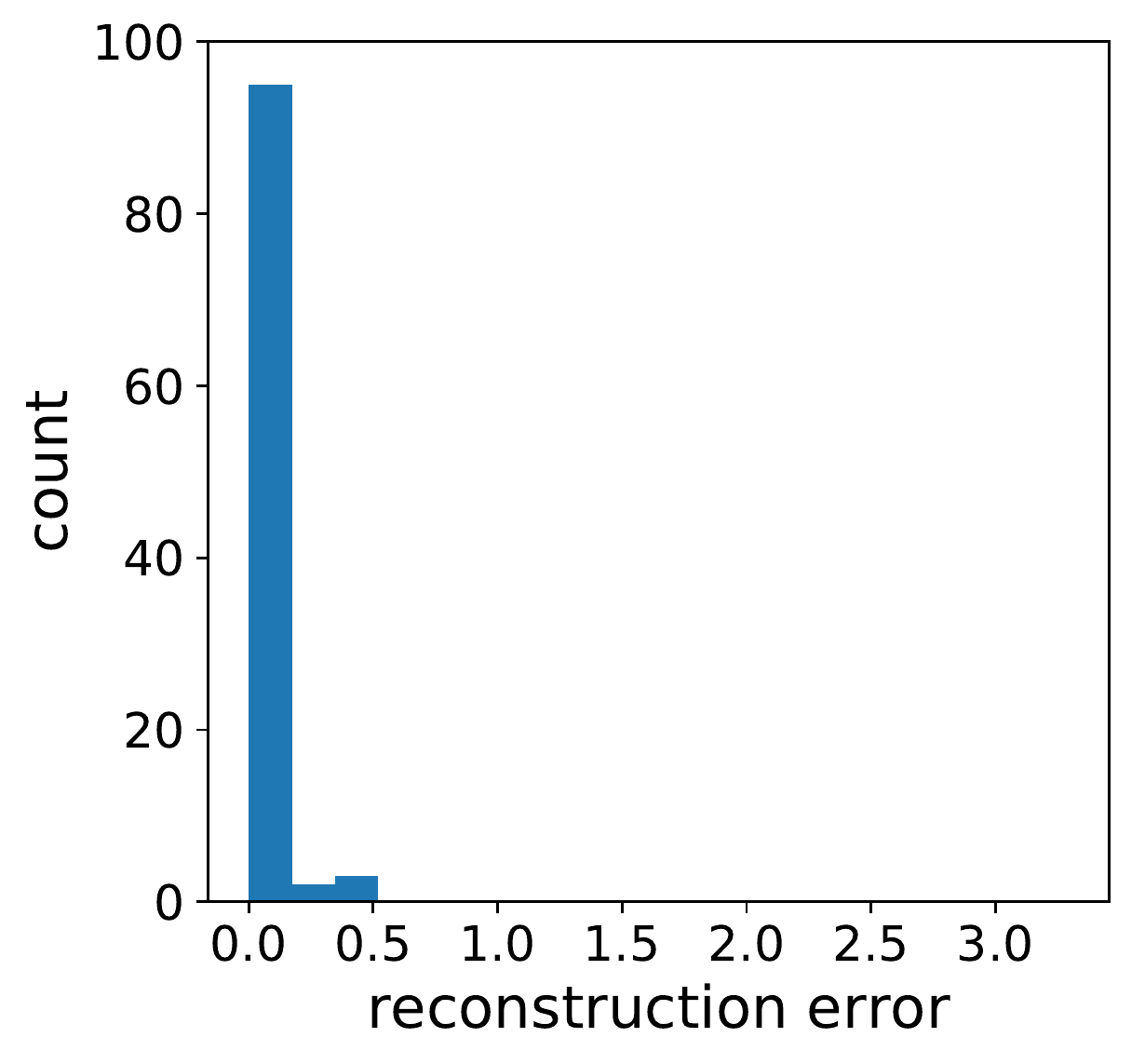}\\[-3pt]
   \rotatebox{90}{\hspace{0.8cm}{\small 1500}} &\includegraphics[width=25mm]{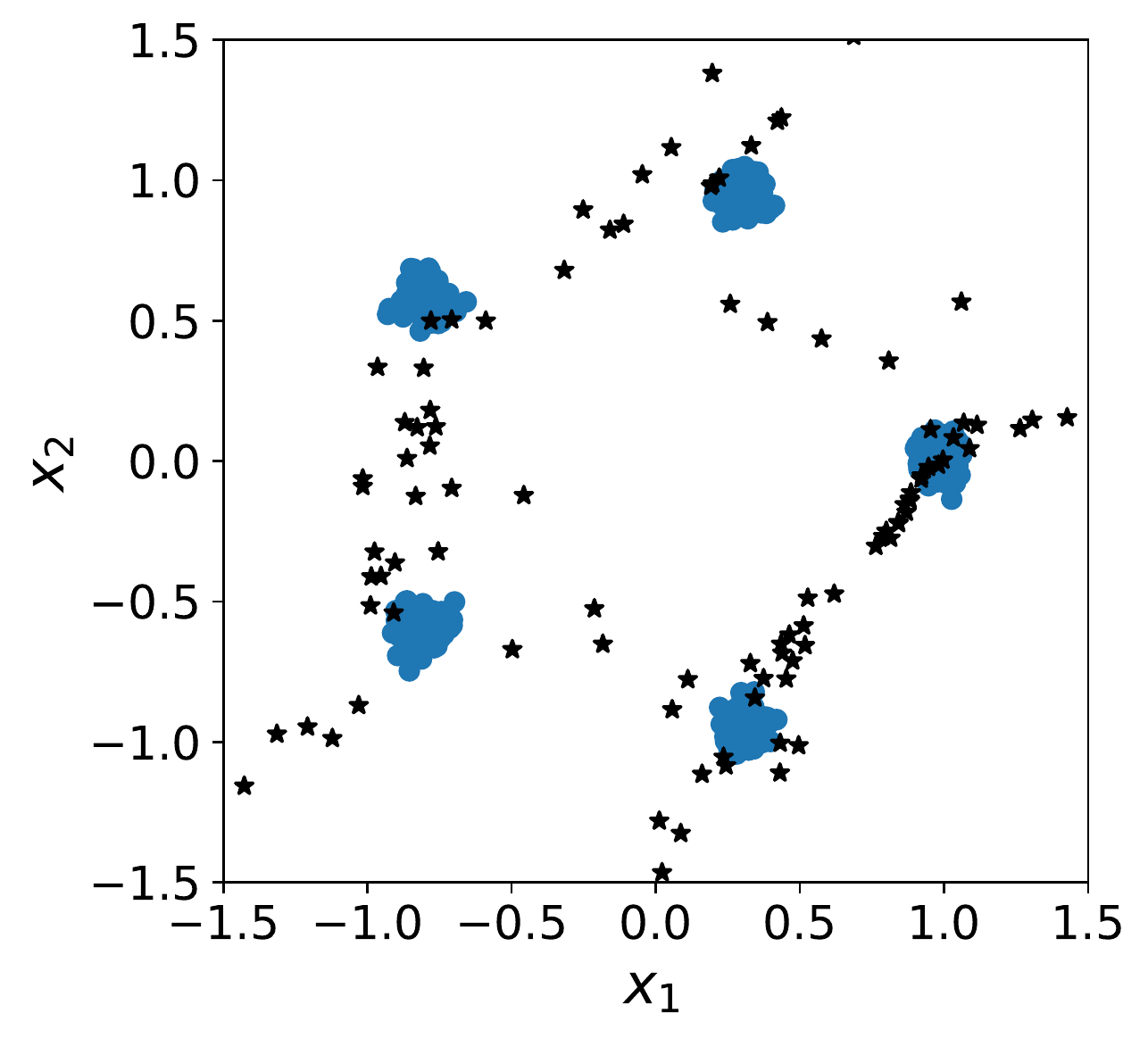}&\includegraphics[width=25mm]{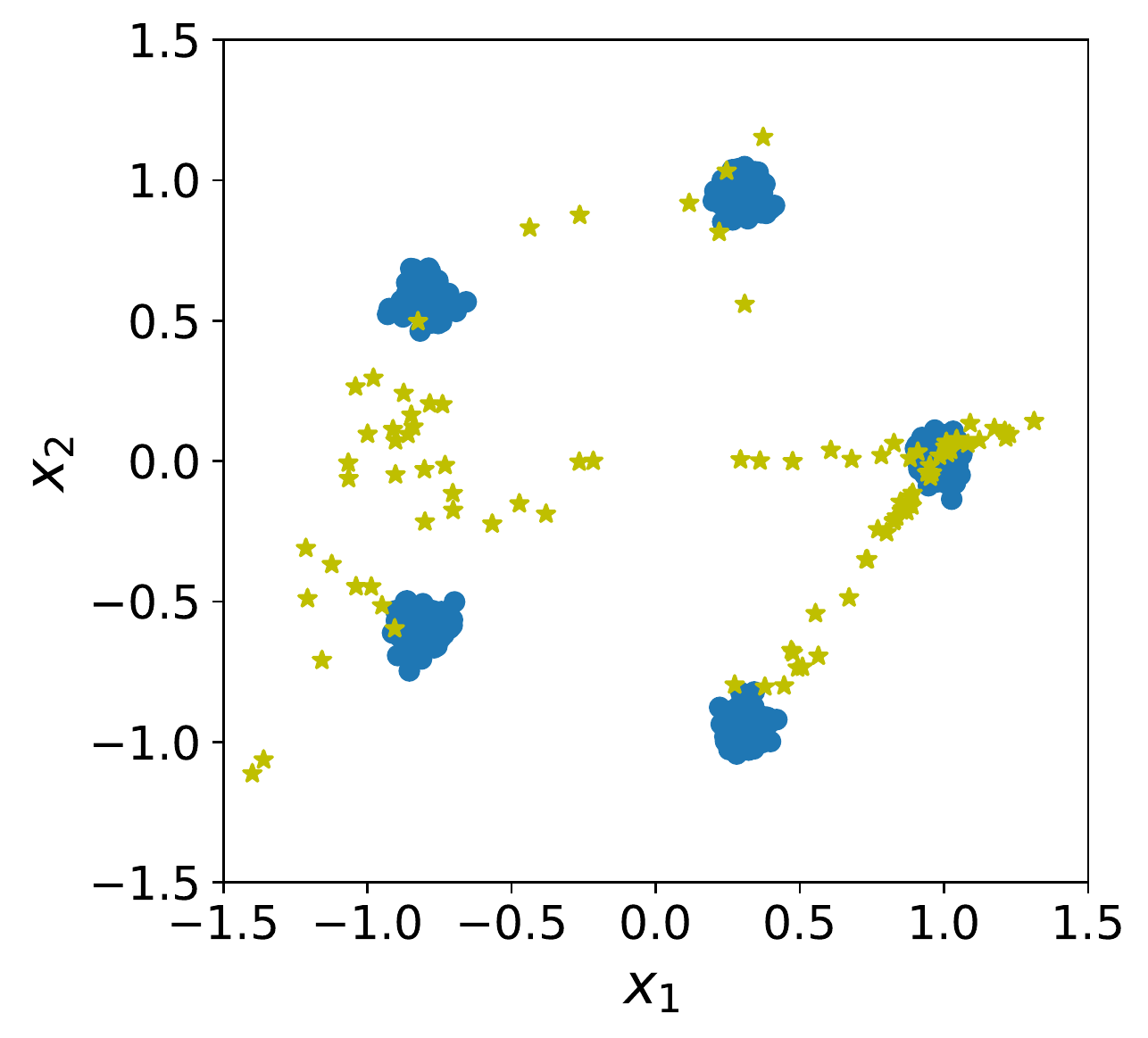}&\includegraphics[width=25mm]{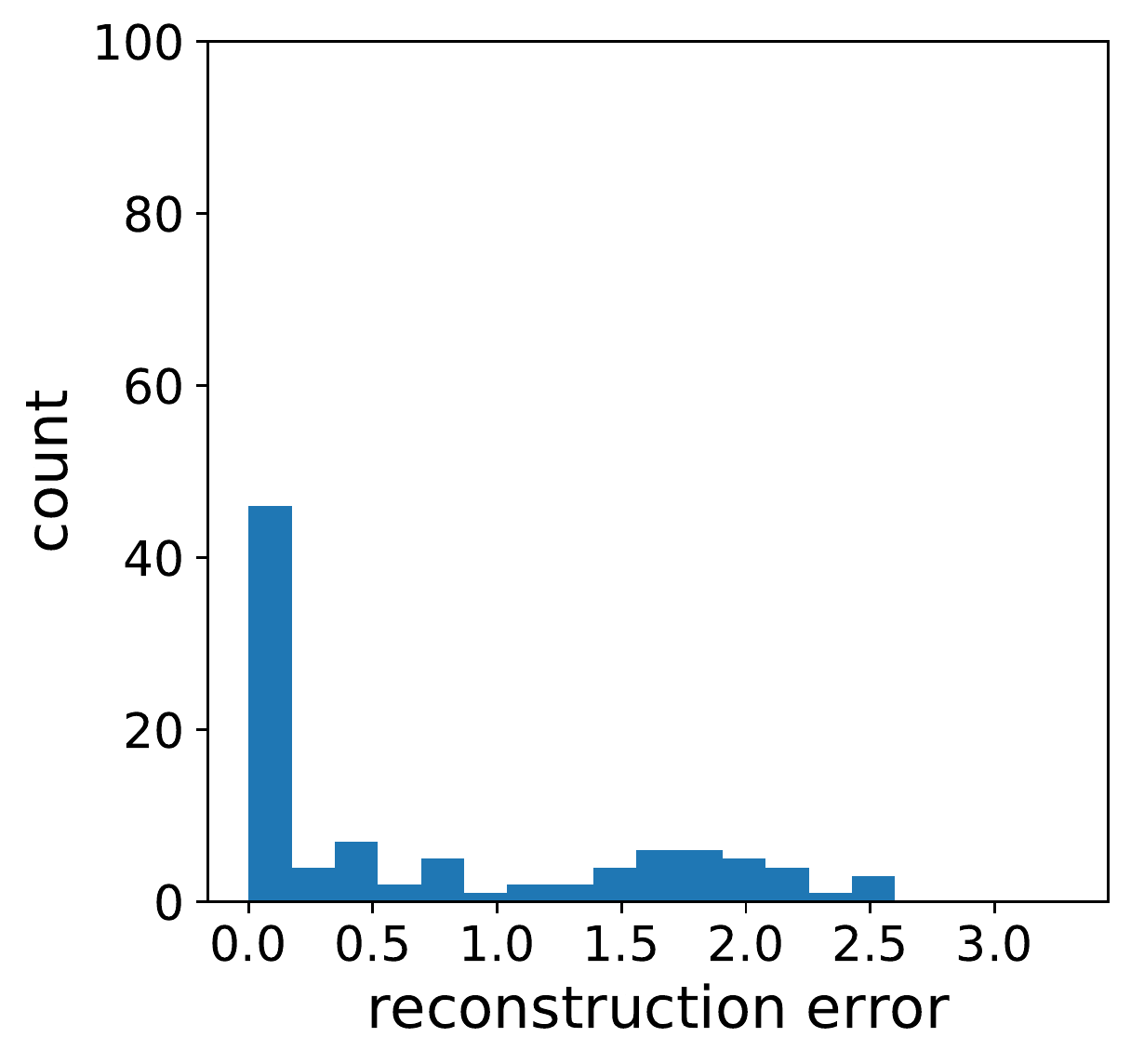}&\includegraphics[width=25mm]{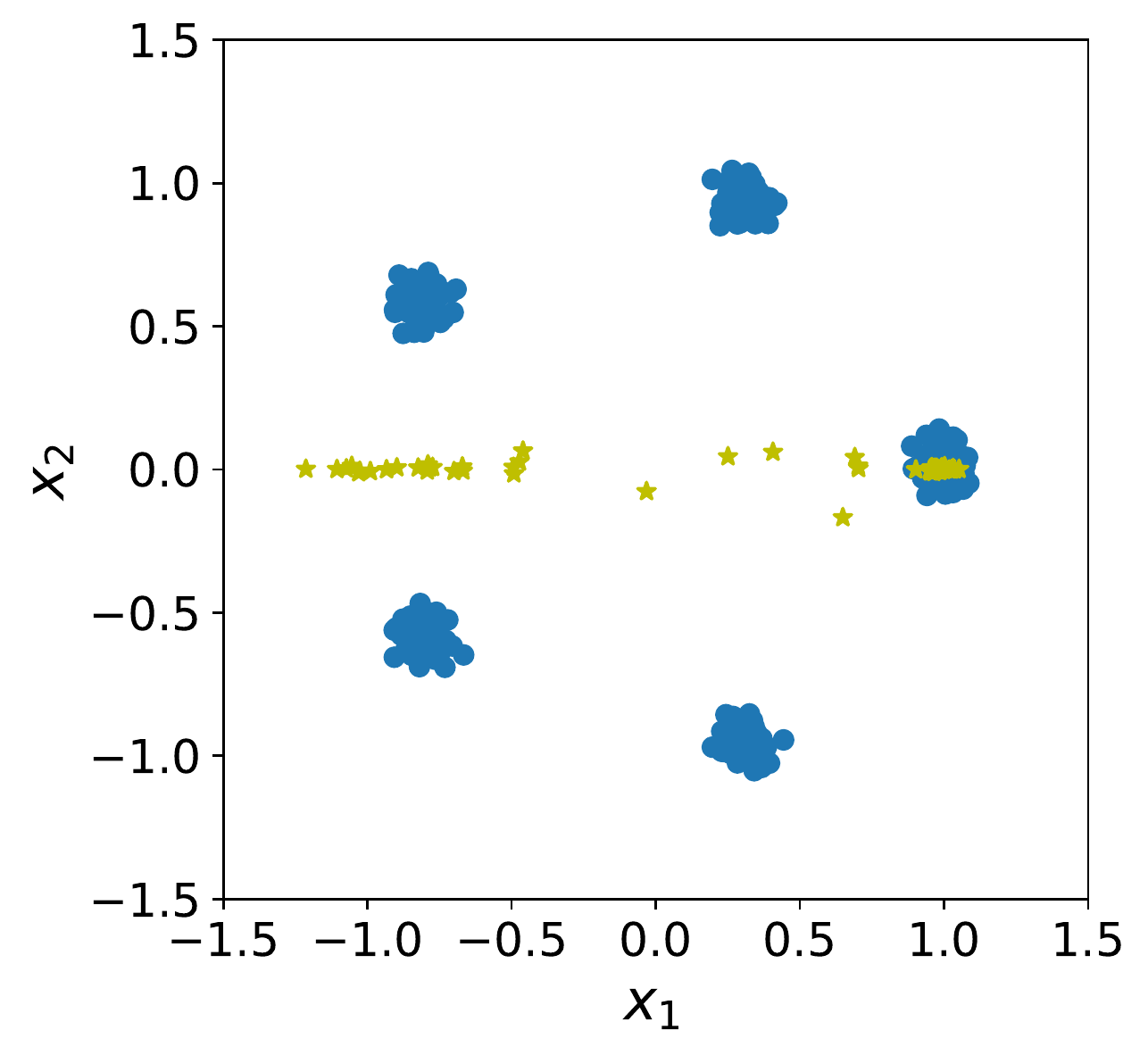}& \includegraphics[width=25mm]{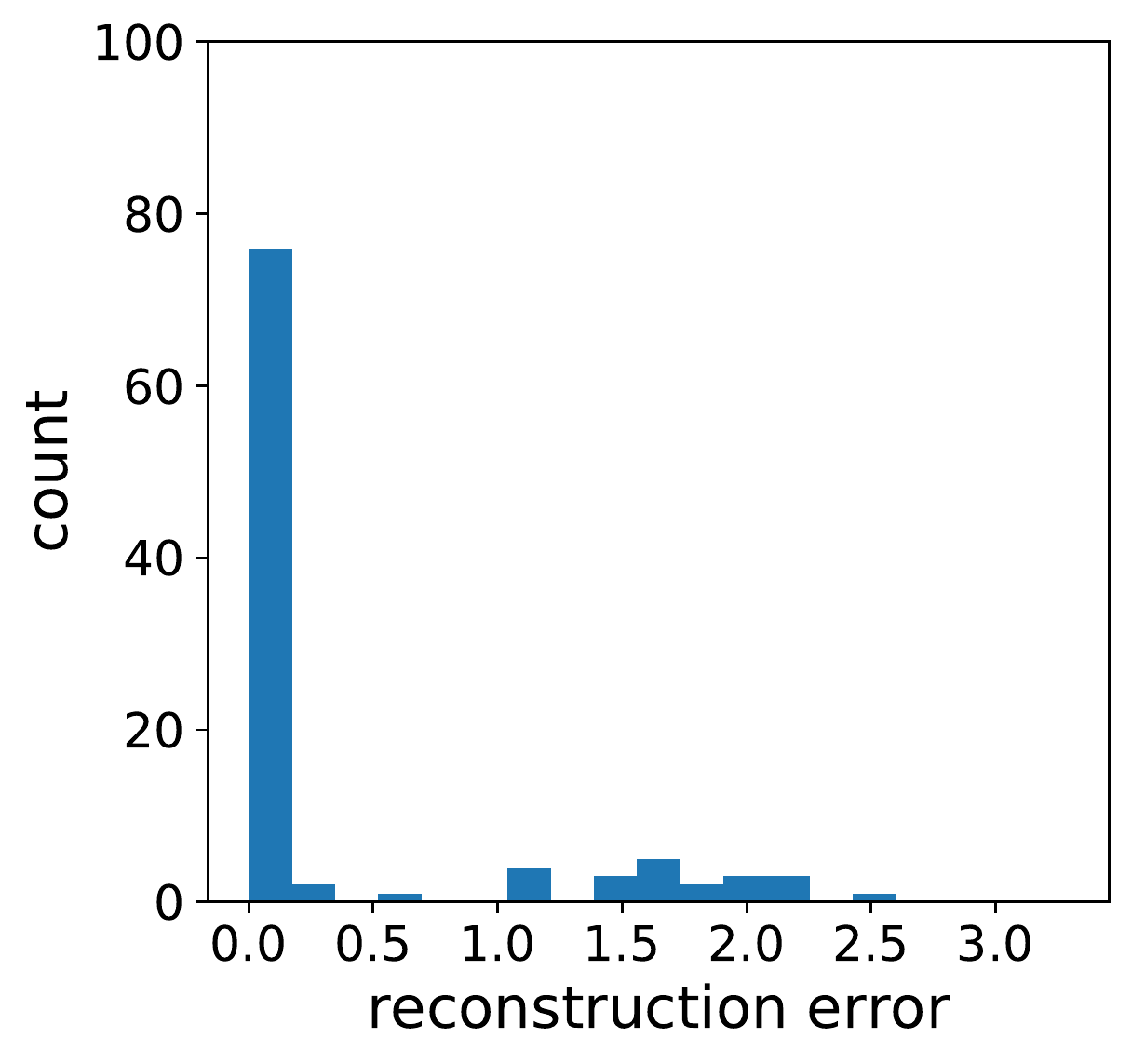}\\[-3pt]
   \rotatebox{90}{\hspace{0.8cm}{\small 2500}} &\includegraphics[width=25mm]{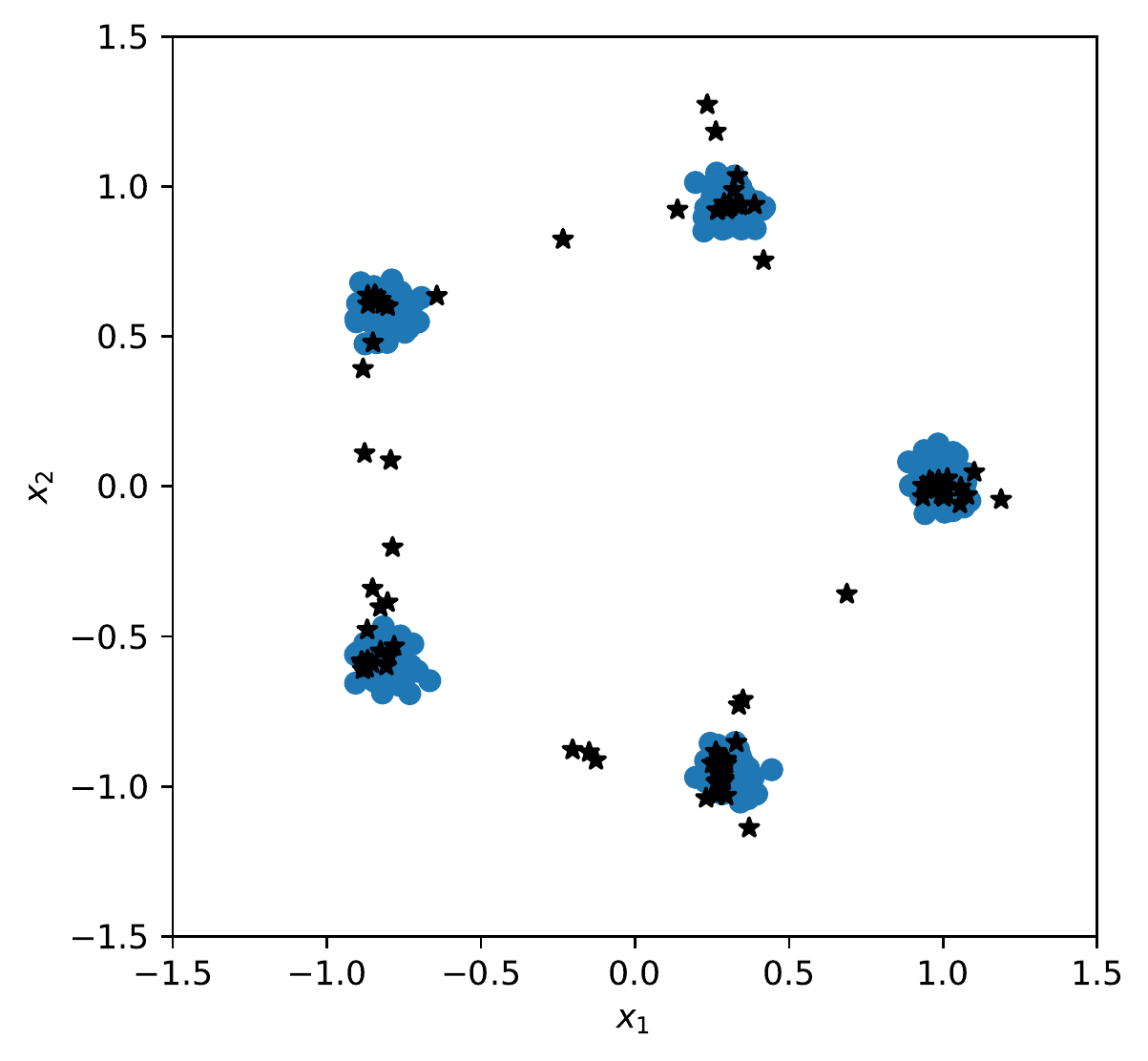}&\includegraphics[width=25mm]{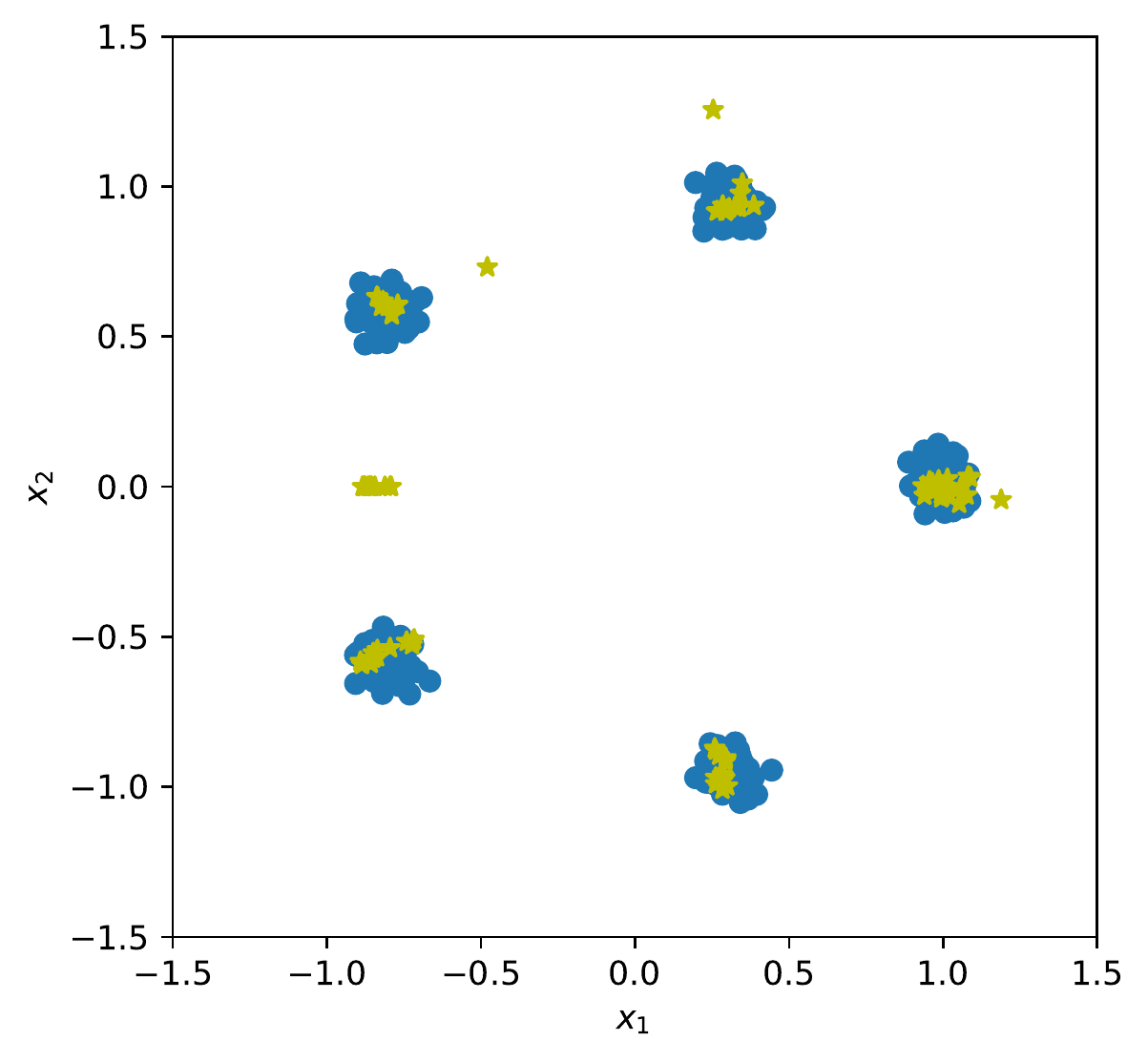}&\includegraphics[width=25mm]{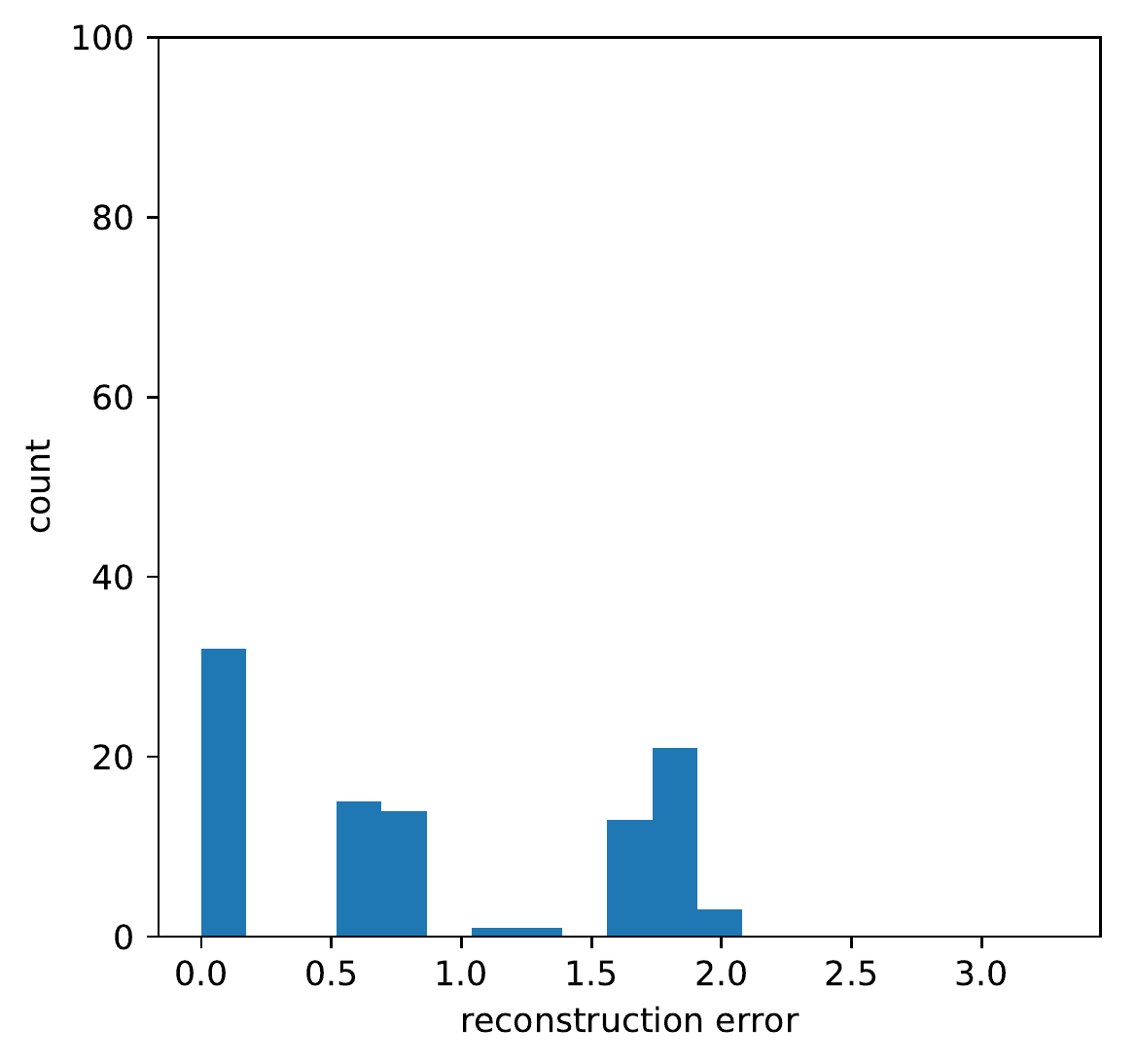}&\includegraphics[width=25mm]{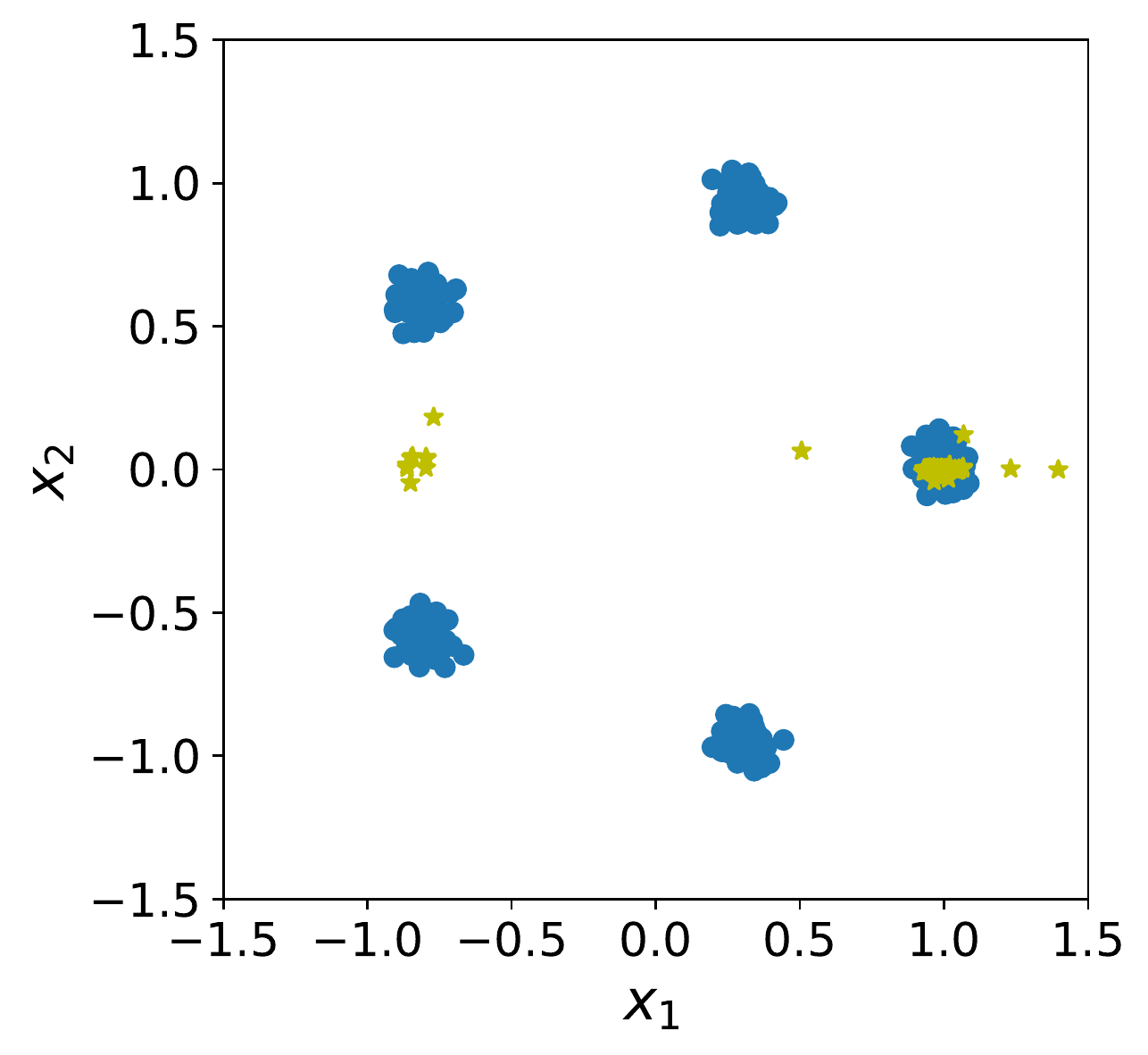}& \includegraphics[width=25mm]{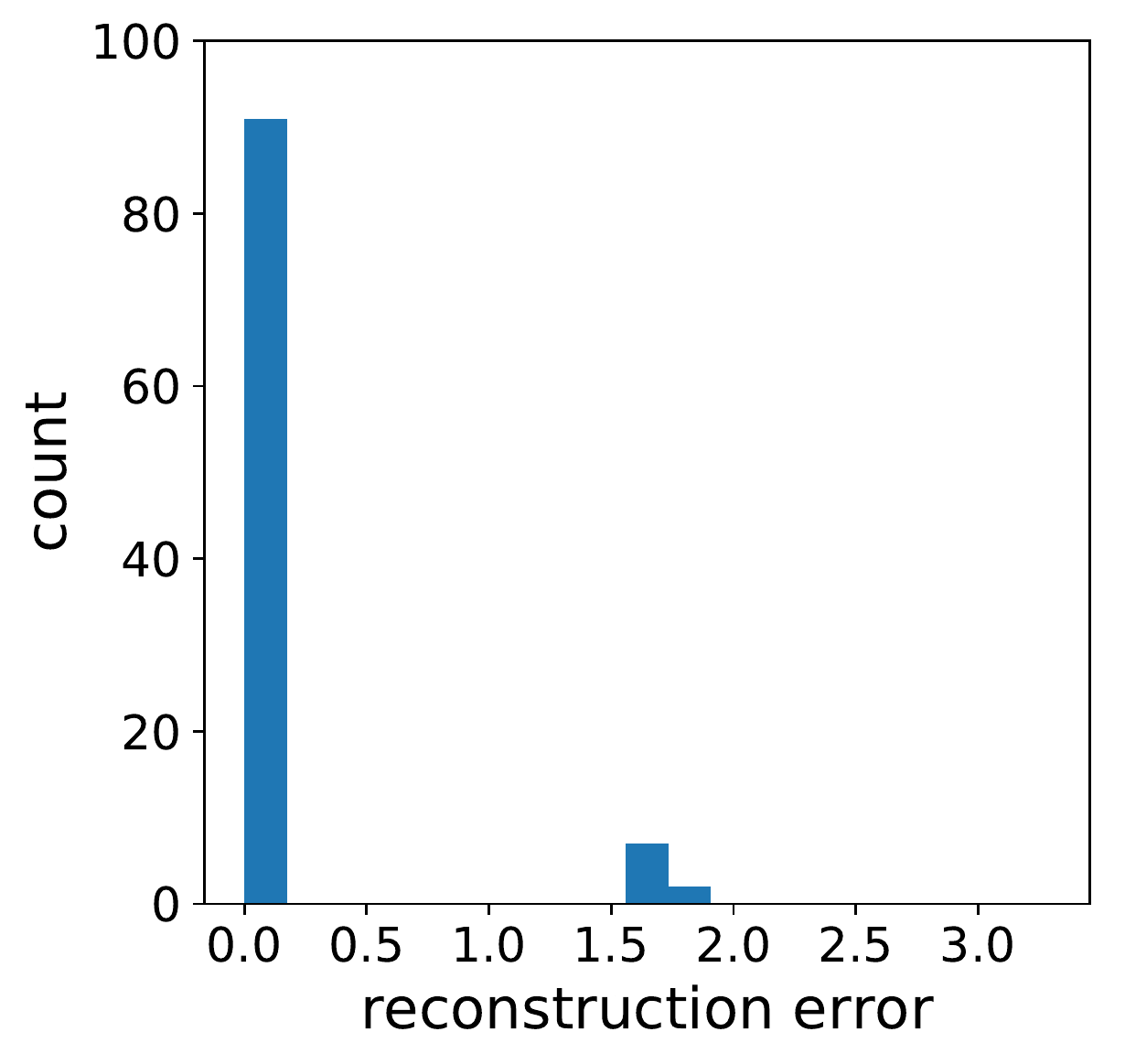}\\[-3pt]
   \rotatebox{90}{\hspace{0.8cm}{\small 15000}} &\includegraphics[width=25mm]{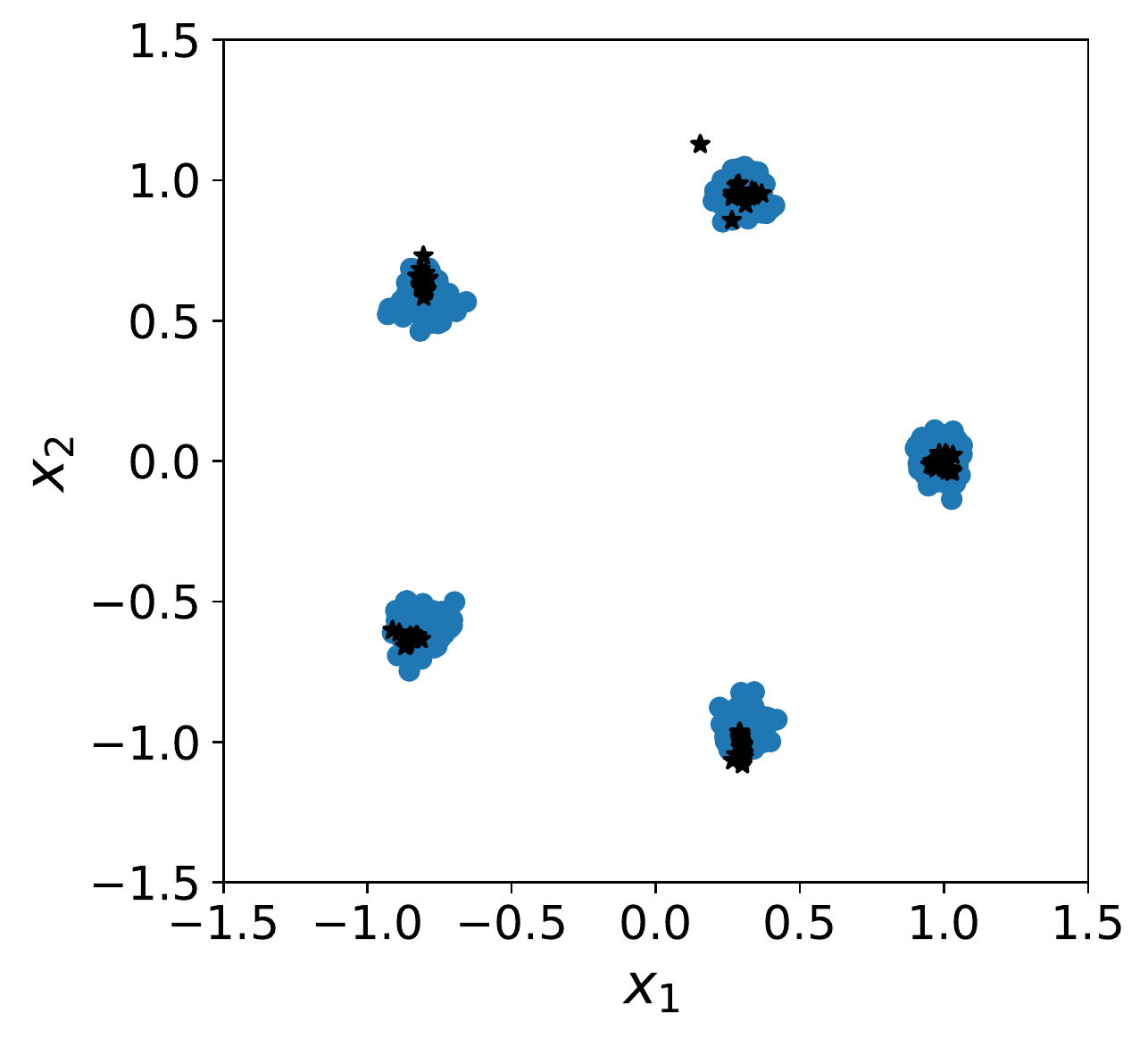}&\includegraphics[width=25mm]{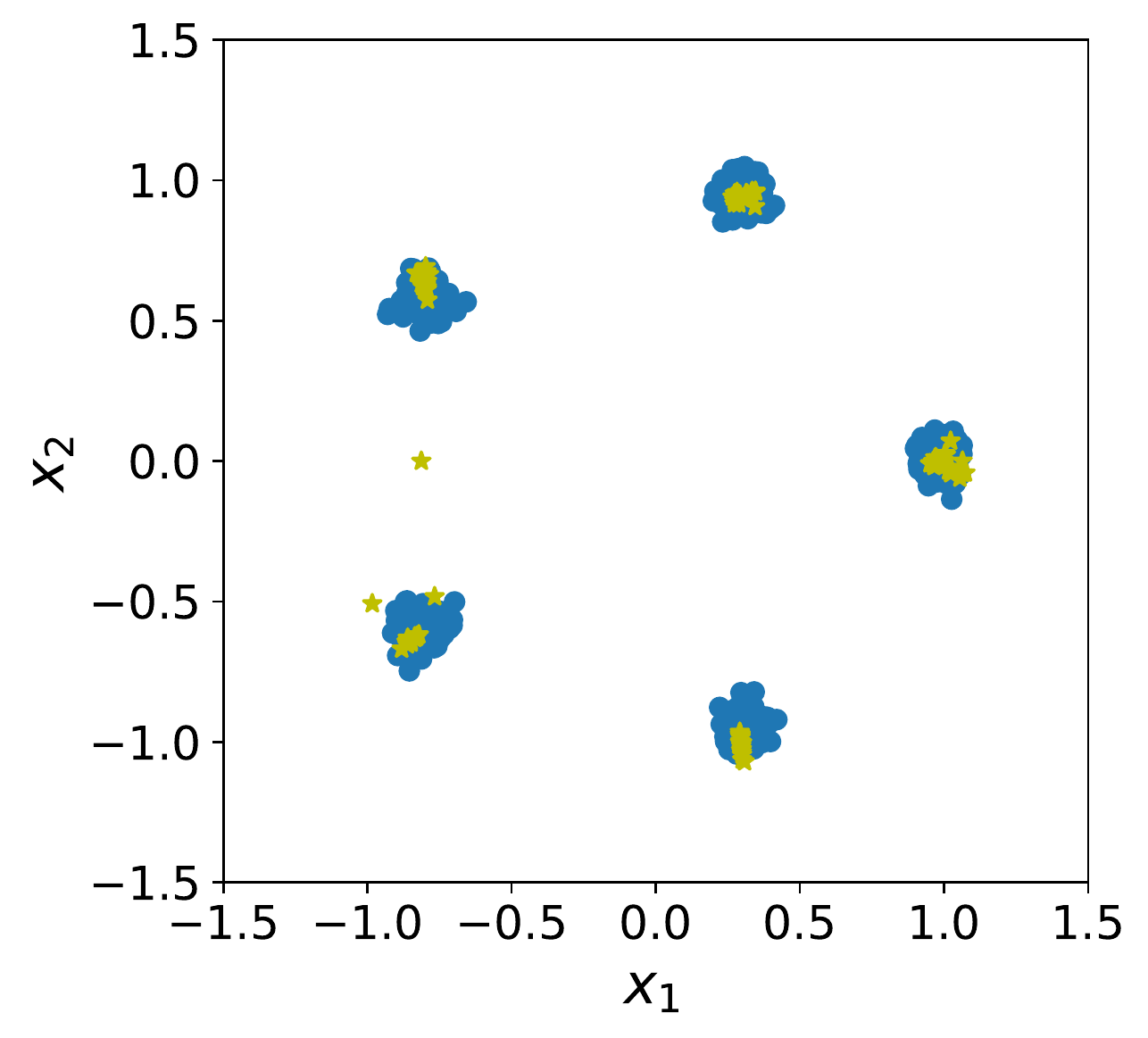}&\includegraphics[width=25mm]{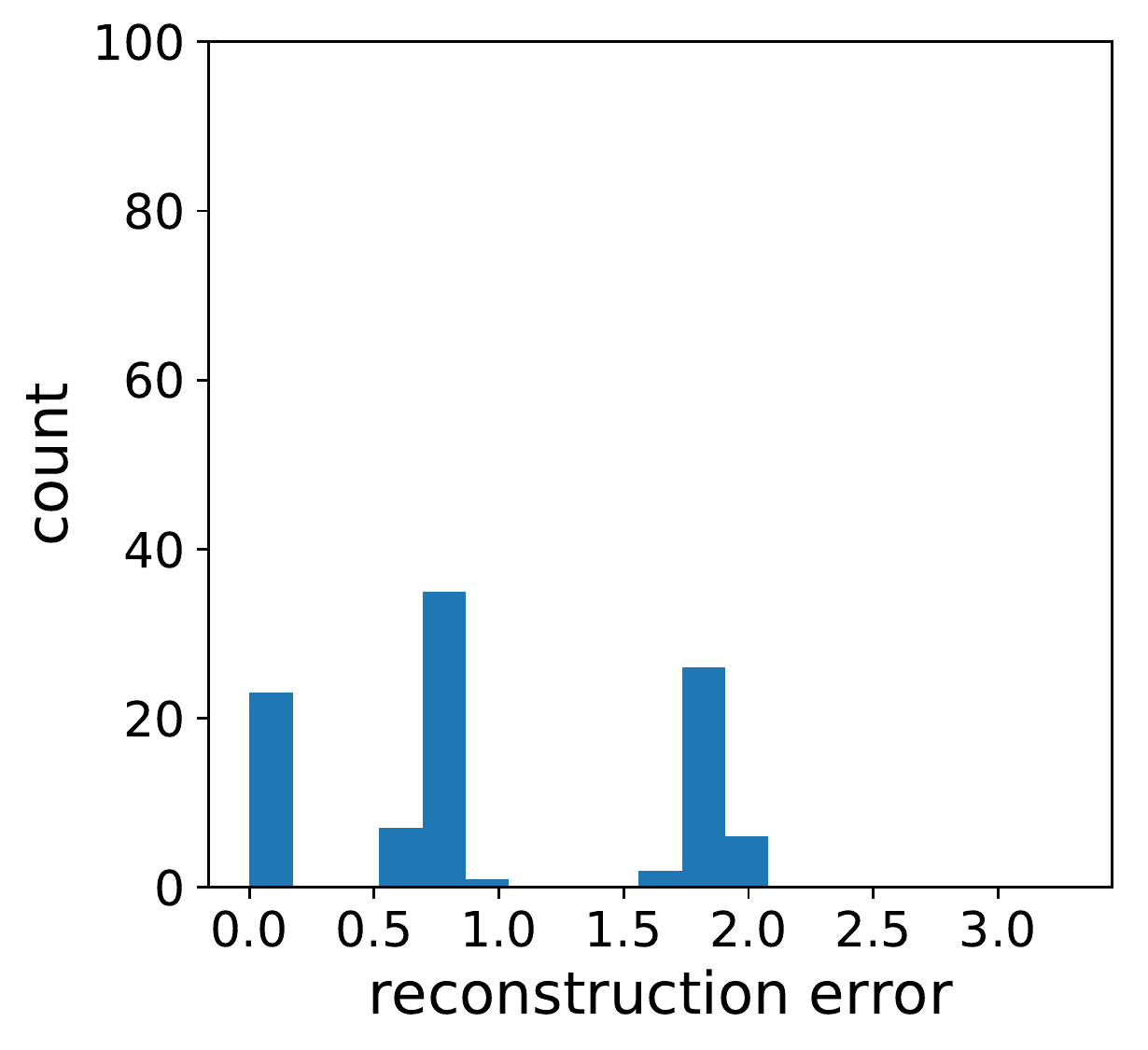}&\includegraphics[width=25mm]{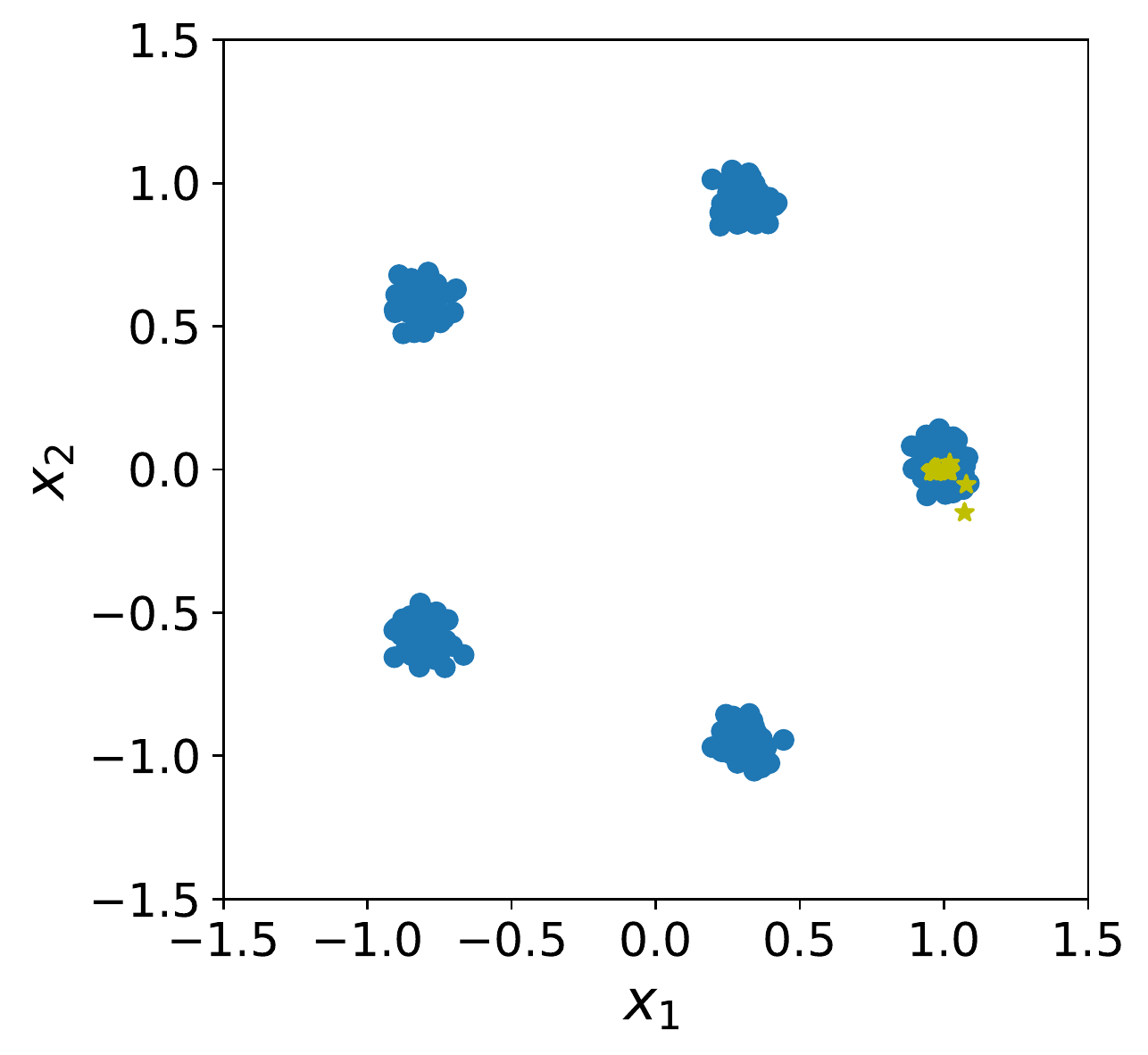}& \includegraphics[width=25mm]{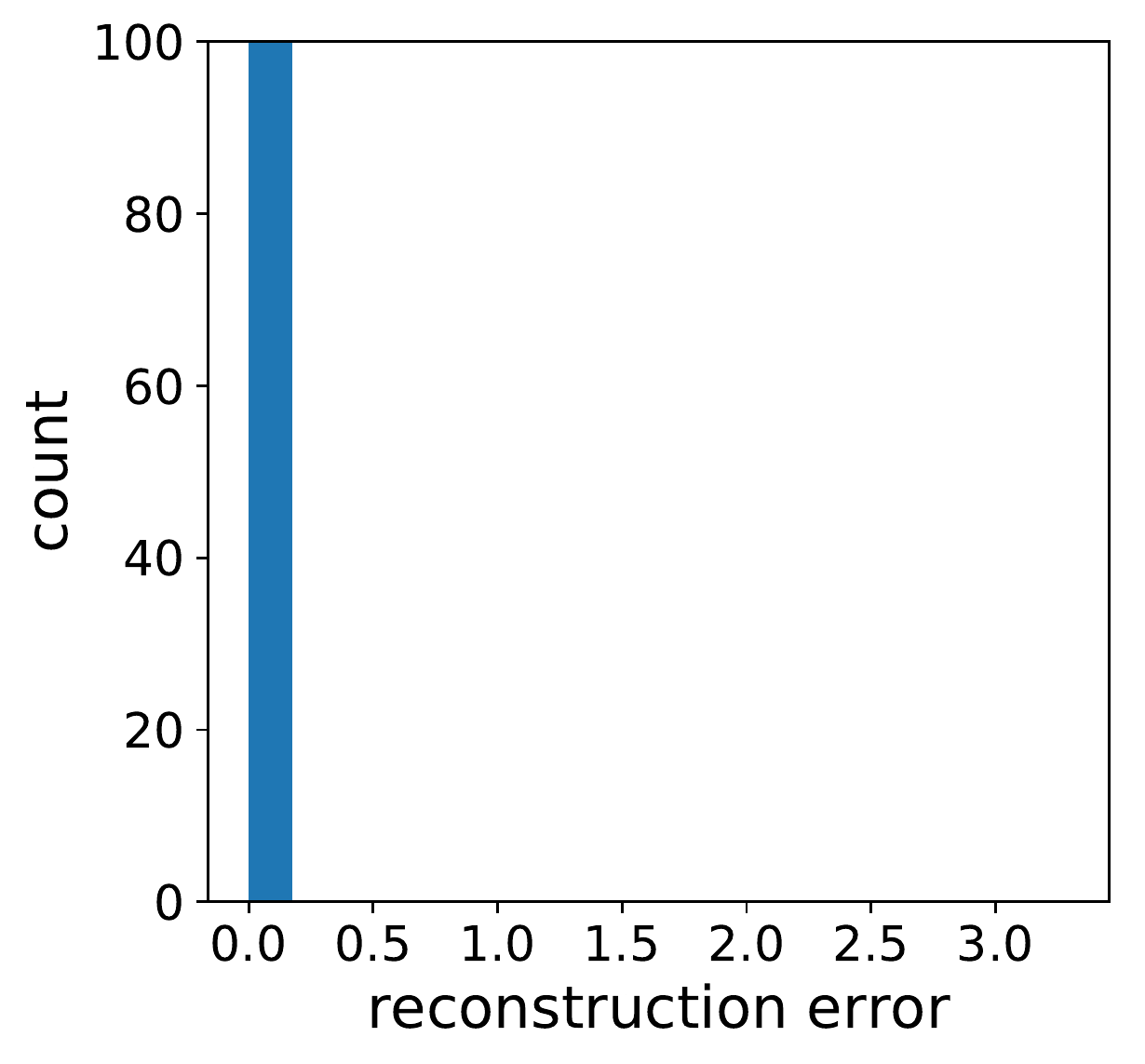}\\[-5pt]
  &\multicolumn{1}{c}{{\small (a)}} & \multicolumn{1}{c}{{\small (b)}} & \multicolumn{1}{c}{{\small (c)}} & \multicolumn{1}{c}{{\small (d)}} & \multicolumn{1}{c}{{\small (e)}}
\end{tabularx}
\vspace{-0.3cm}
\caption{Rows correspond to generators trained for a different number of epochs as indicated (left). The columns illustrate: (a) Samples generated with a vanilla GAN (black); (b) GD reconstructions from 100 random initializations; (c)  Reconstruction error bar plot for the result in column (b); (d) Reconstructions recovered with \algref{alg:ours}; (e) Reconstruction error bar plot for the results in column (d). 
}
\label{fig:toycomp}
\vspace{-0.3cm}
\end{figure}

\begin{figure}[t]
\centering
\setlength{\tabcolsep}{0pt}

\begin{tabular}{ccccc}
  \multicolumn{1}{c}{{\small $t = 100$}} & \multicolumn{1}{c}{{\small $t = 2000$}} & \multicolumn{1}{c}{{\small $t = 3000$}} & \multicolumn{1}{c}{{\small $t = 4000$}} & \multicolumn{1}{c}{{\small $t = 6000$}} \\[-3pt]
  \includegraphics[width=25mm]{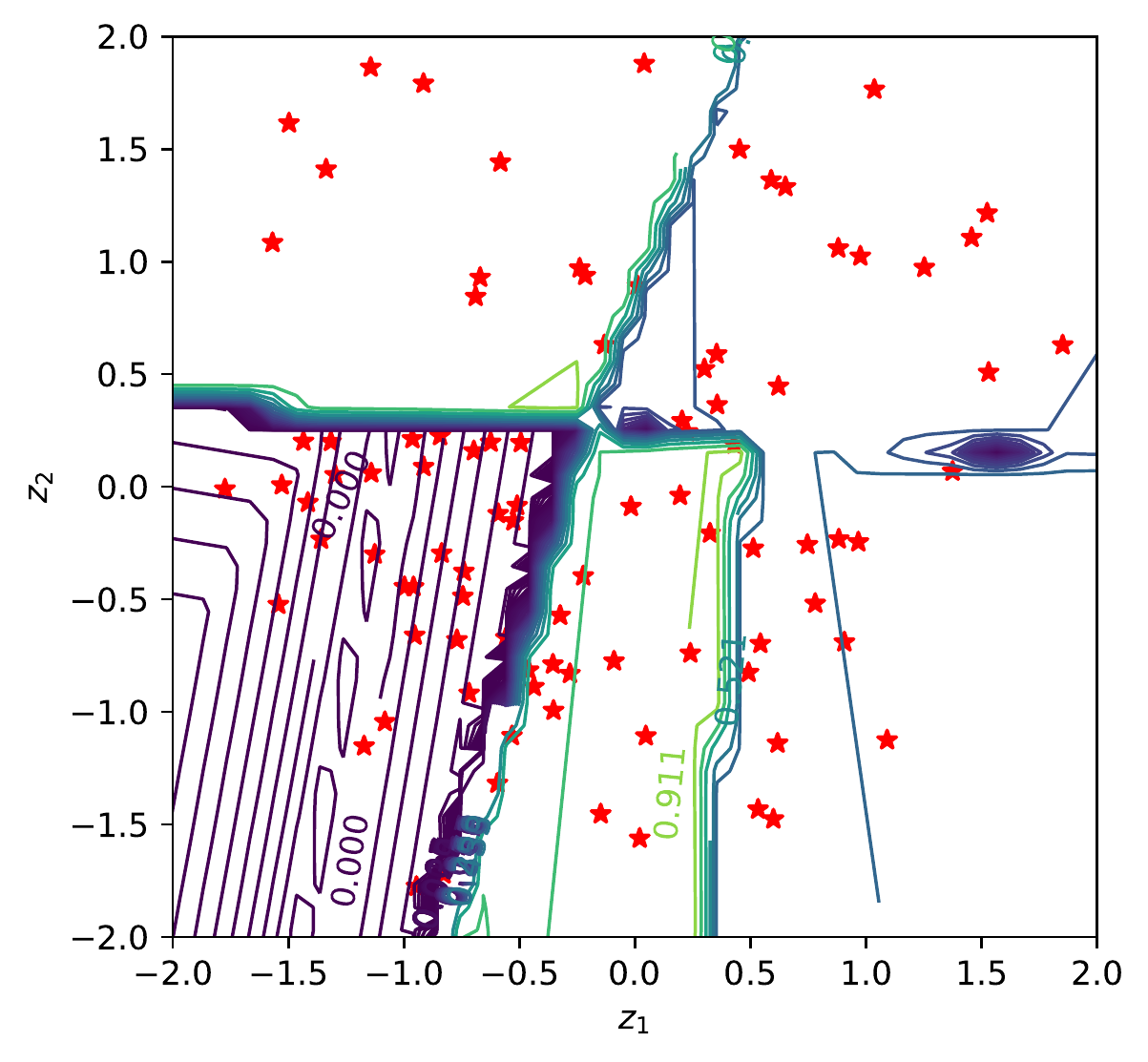}&\includegraphics[width=25mm]{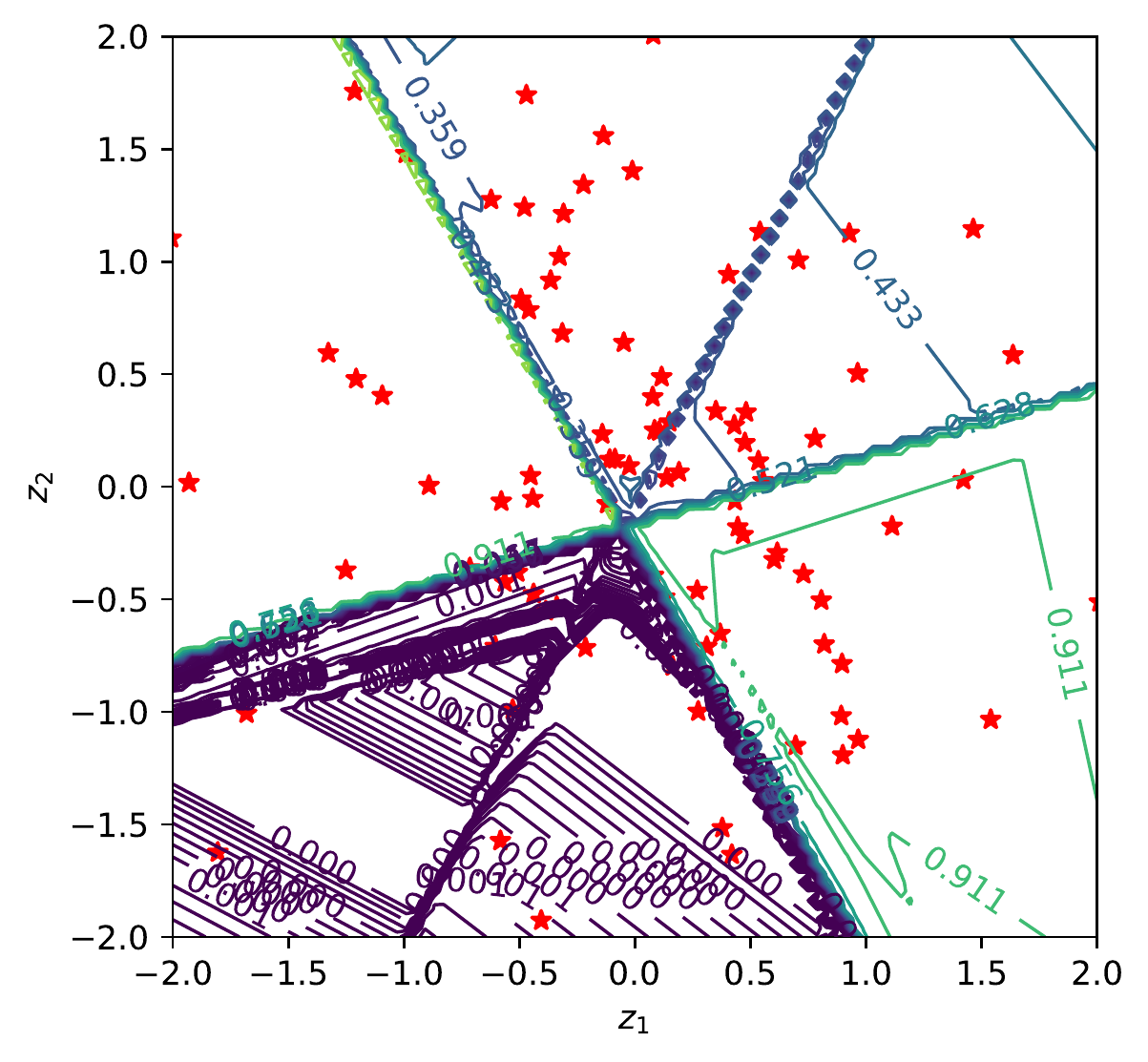}&\includegraphics[width=25mm]{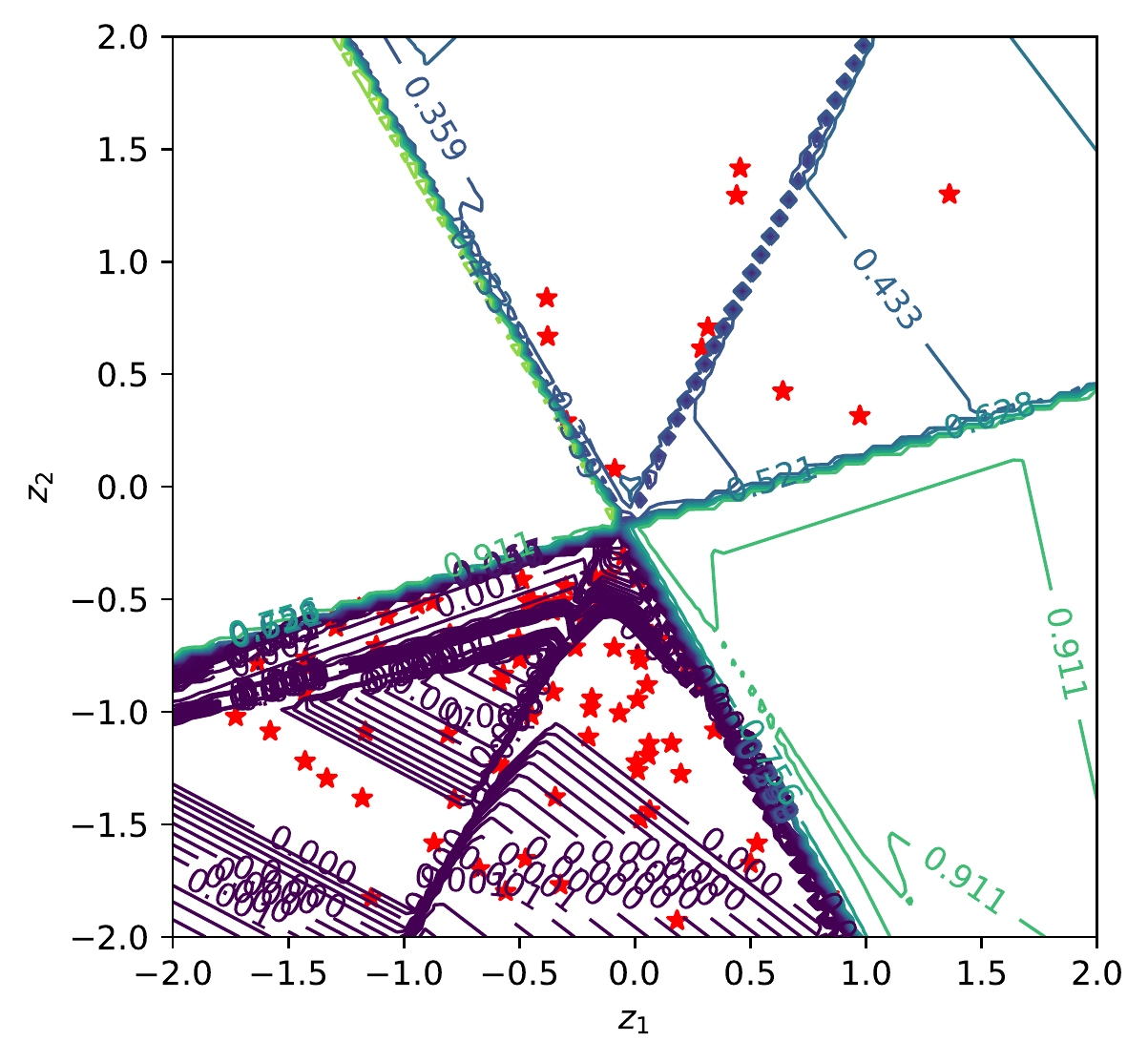}&\includegraphics[width=25mm]{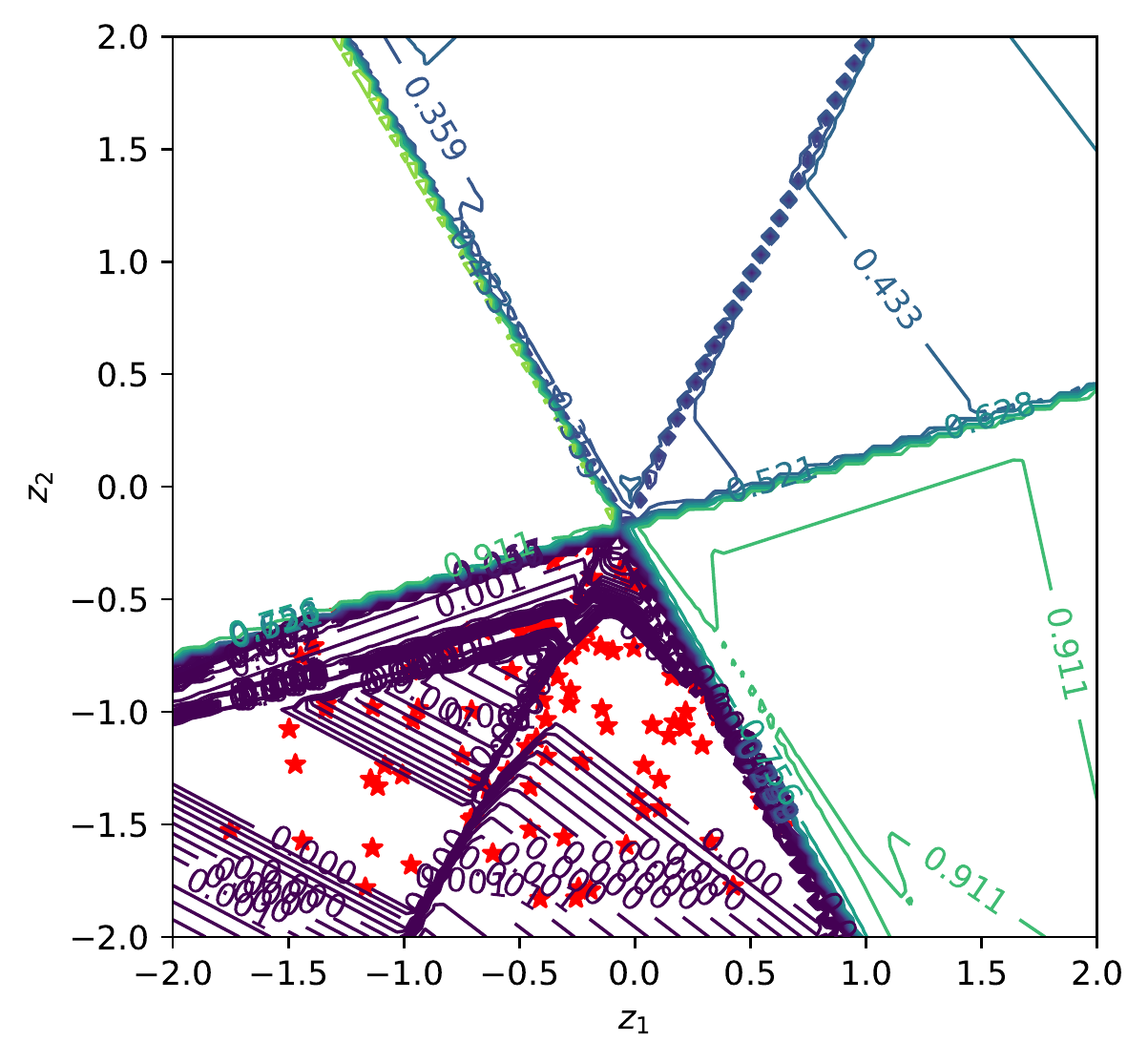}&\includegraphics[width=25mm]{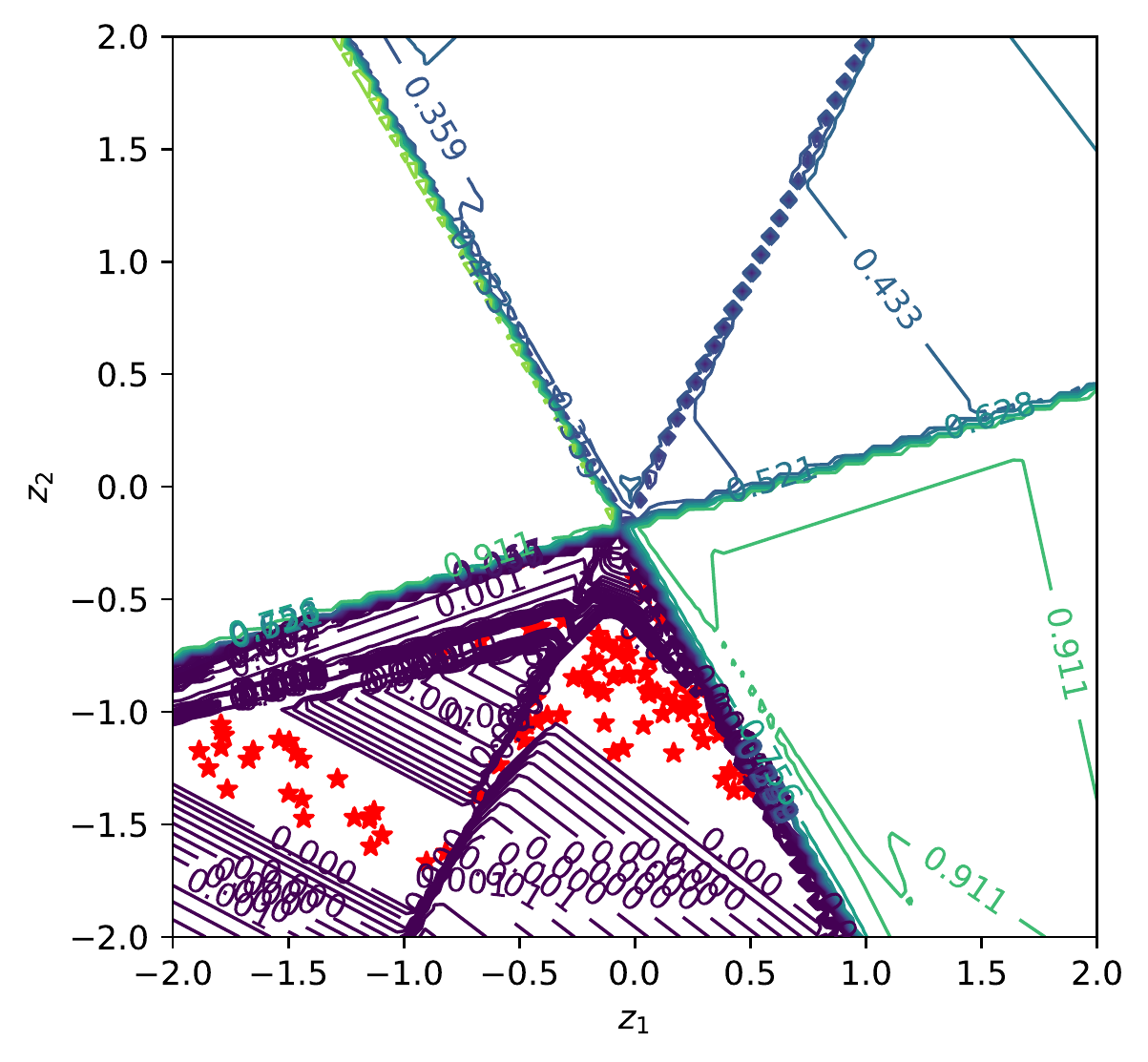} 
  \end{tabular}
\vspace{-0.2cm}
\caption{Samples $z$  in $\cZ$ space during the AIS procedure: after 100, 2k, 3k, 4k and 6k AIS loops.
 }
\label{fig:aisprocedure}
\vspace{-0.5cm}
\end{figure}

\vspace{-0.2cm}
\section{Experiments}
\label{sec:exp}
\vspace{-0.2cm}

\textbf{Baselines: }In the following, we evaluate the proposed approach on synthetic and imaging data. We compare \algref{alg:ours} with two GD baselines by employing two different initialization strategies. The first one is sampling a single $z$ randomly. The second  
picks that one sample $z$ from $5000$ initial points which best matches the objective given in \equref{eq:backprop}.  

\begin{figure}[t]
\centering
 \begin{tabularx}{\textwidth} {@{\hskip3pt}c@{\hskip3pt}c@{\hskip3pt}c@{\hskip3pt}c@{\hskip3pt}c@{\hskip3pt}c}
  \includegraphics[width=20mm]{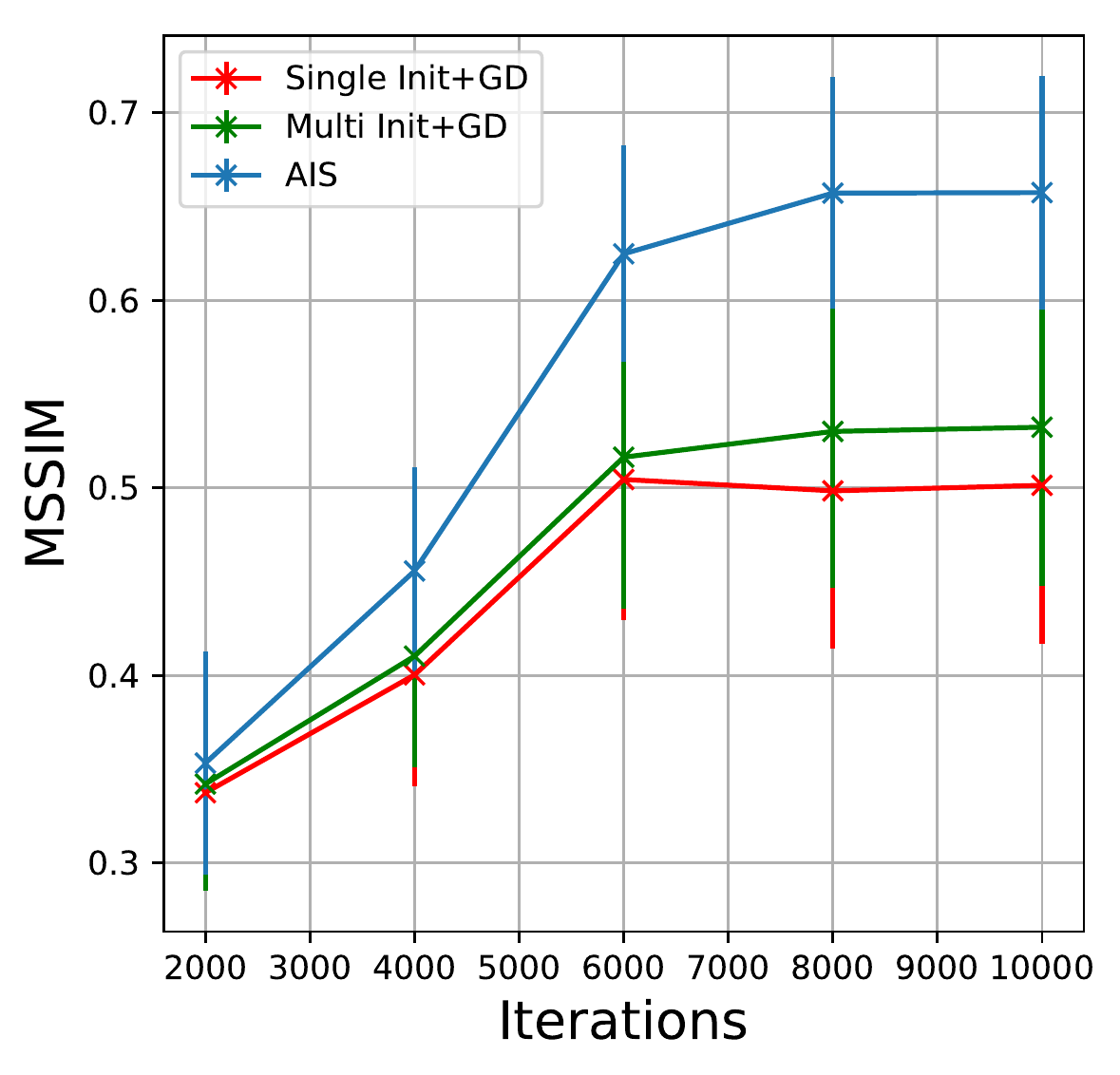}&\includegraphics[width=21mm]{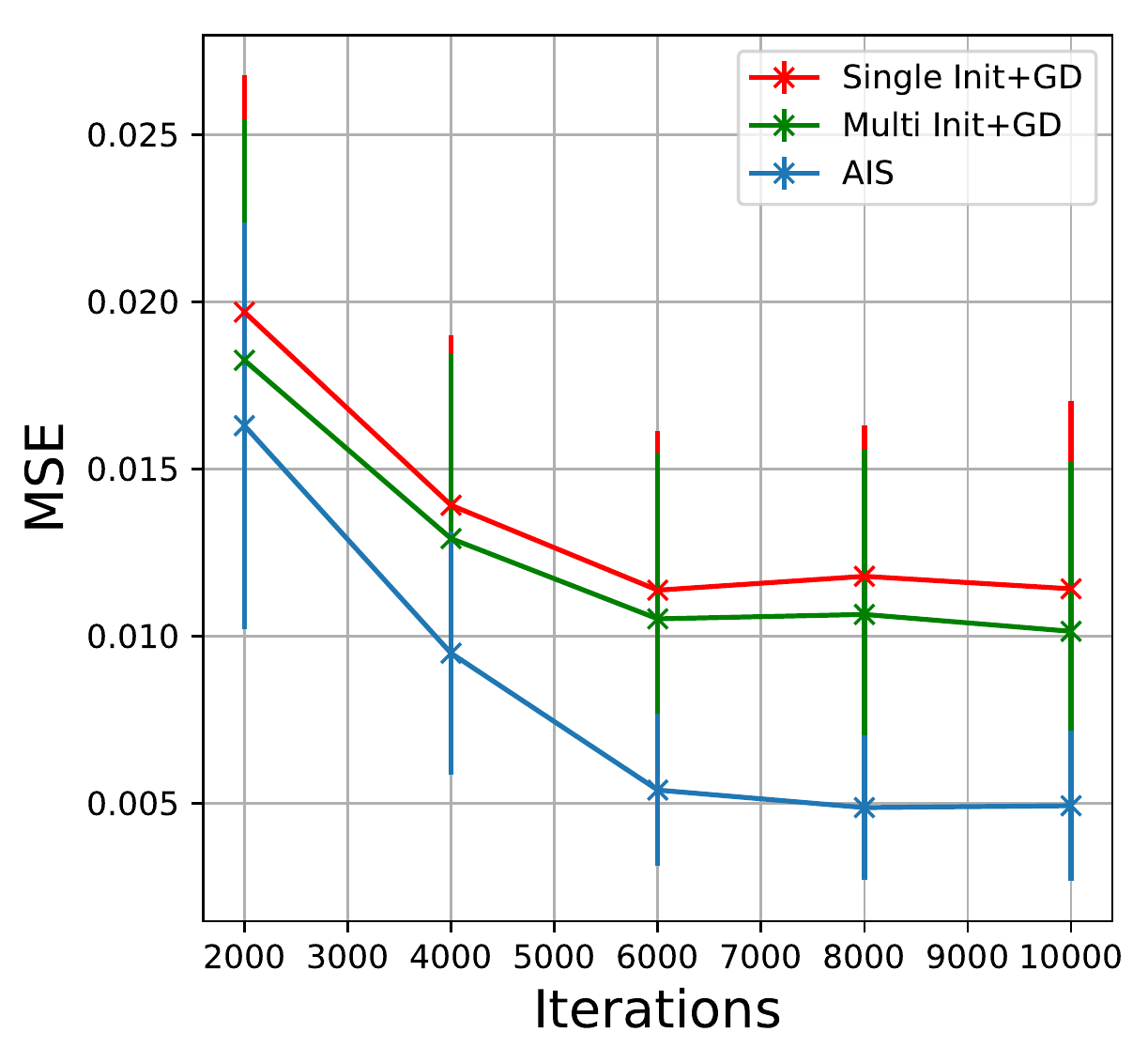}&\includegraphics[width=21mm]{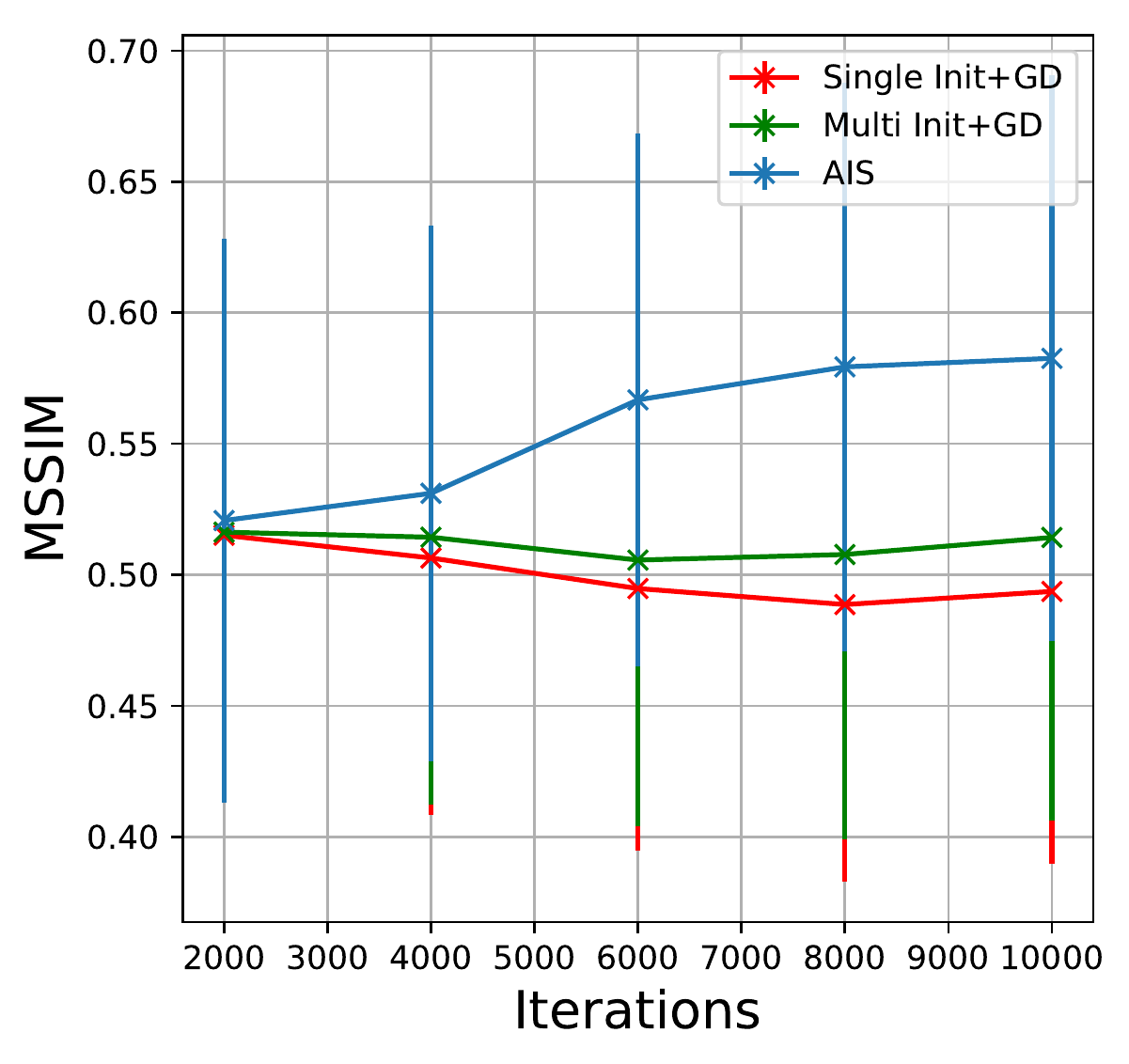}&\includegraphics[width=21mm]{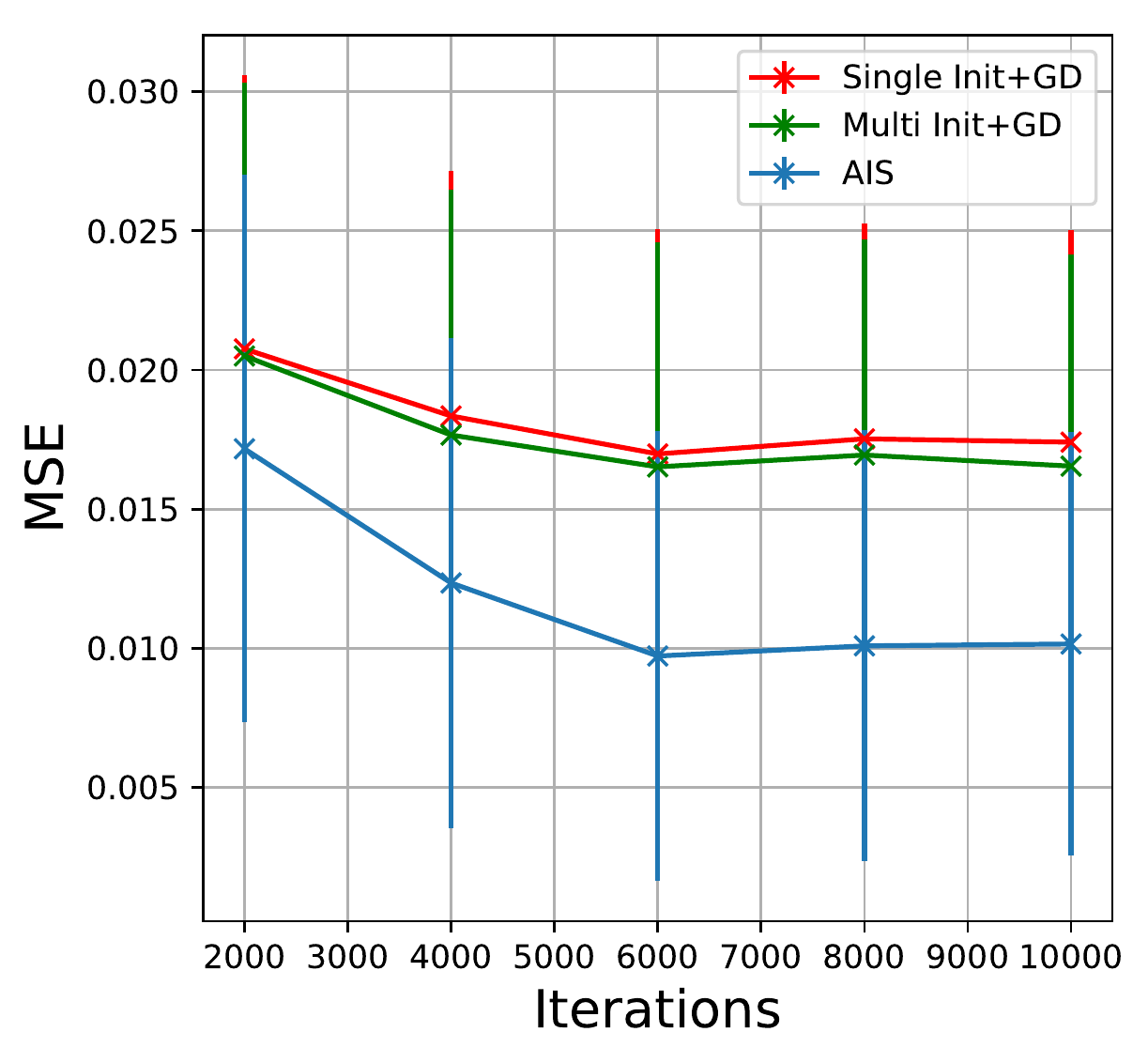}&\includegraphics[width=21mm]{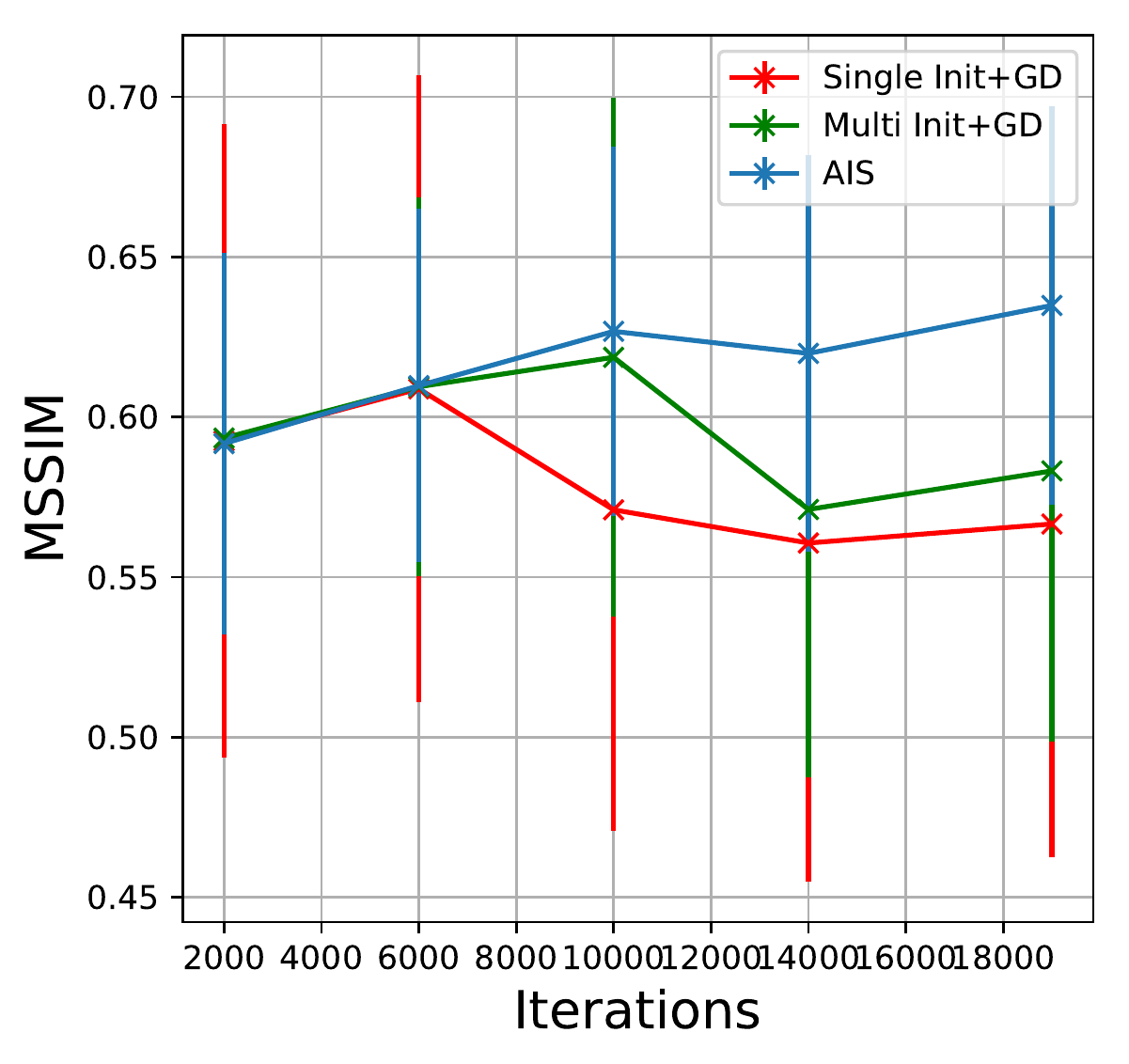}&\includegraphics[width=21mm]{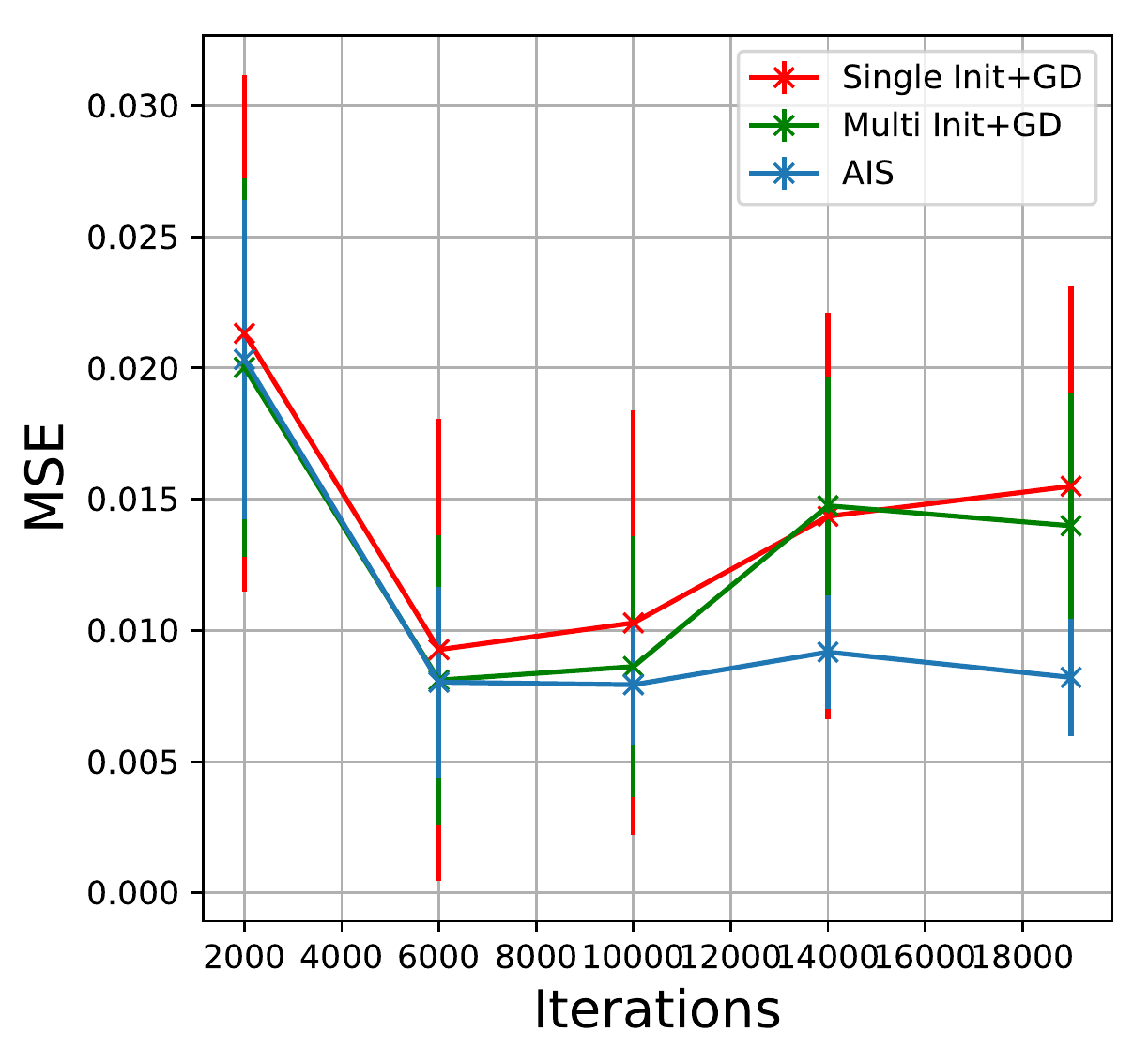}\\[-3pt]
  \multicolumn{1}{c}{{\small (a)}} & \multicolumn{1}{c}{{\small (b)}} & \multicolumn{1}{c}{{\small (c)}} & \multicolumn{1}{c}{{\small (d)}} & \multicolumn{1}{c}{{\small (e)}} & \multicolumn{1}{c}{{\small (f)}} 
\end{tabularx}
\vspace{-0.3cm}
\caption{Reconstructions errors over the number of progressive GAN training iterations. (a) MSSIM on CelebA; (b) MSE on CelebA; (c) MSSIM on LSUN; (d) MSE on LSUN; (e) MSSIM on CelebA-HQ; (f) MSE on CelebA-HQ.}
\label{fig:metric}
\vspace{-0.5cm}
\end{figure}

\begin{figure}[t]
\centering
\scalebox{0.9}{
\begin{tabular}{ccccc}
  \multicolumn{1}{c}{{\small Ground Truth}} & \multicolumn{1}{c}{{\small Masked Image}} & \multicolumn{1}{c}{{\small GD+single $z$}} & \multicolumn{1}{c}{{\small GD+multi $z$}} &
  \multicolumn{1}{c}{AIS}
\\
  \includegraphics[width=21mm]{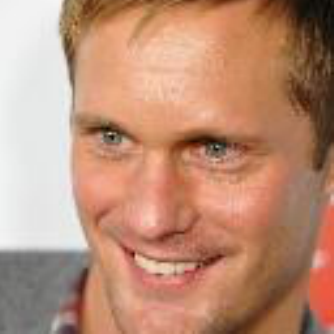}&\includegraphics[width=21mm]{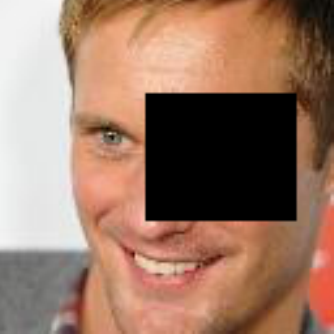}&\includegraphics[width=21mm]{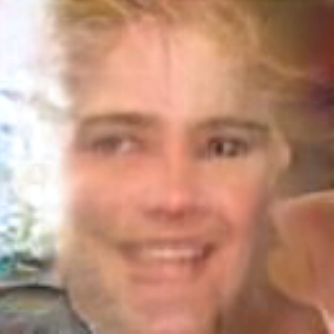}&\includegraphics[width=21mm]{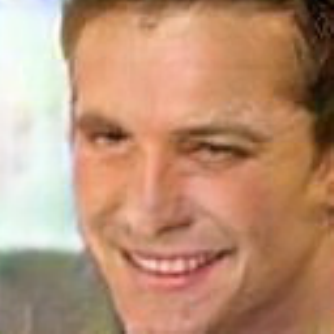}&\includegraphics[width=21mm]{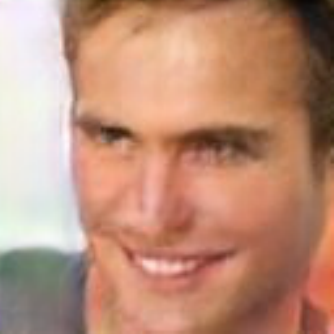}\\
   \includegraphics[width=21mm]{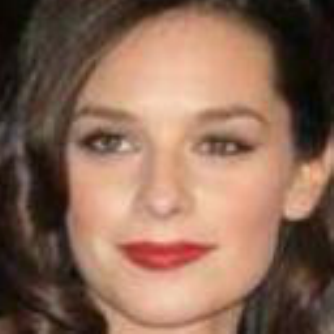}&\includegraphics[width=21mm]{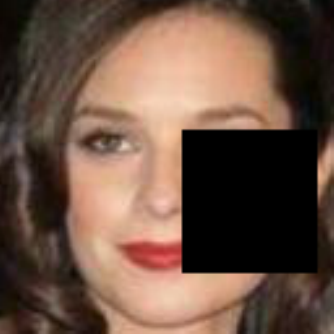}&\includegraphics[width=21mm]{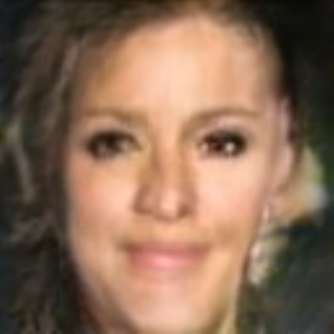}&\includegraphics[width=21mm]{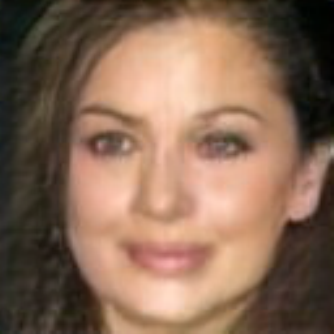}&\includegraphics[width=21mm]{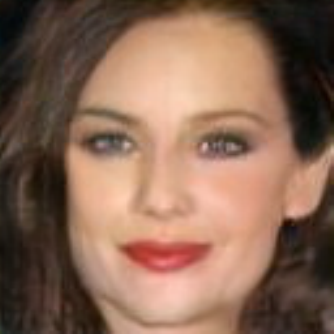}\\
   \includegraphics[width=21mm]{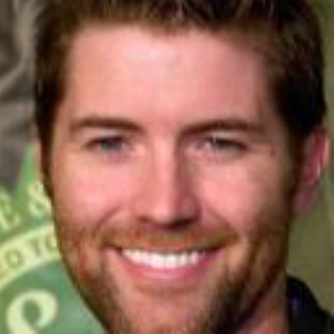}&\includegraphics[width=21mm]{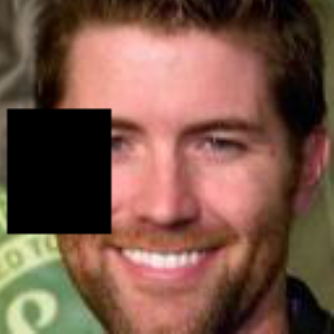}&\includegraphics[width=21mm]{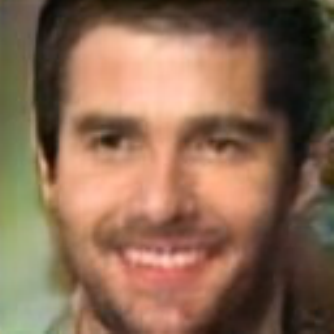}&\includegraphics[width=21mm]{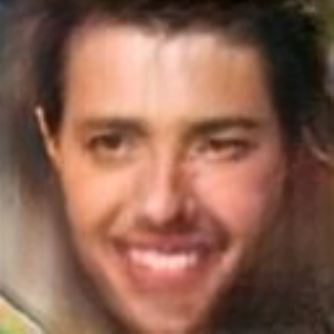}&\includegraphics[width=21mm]{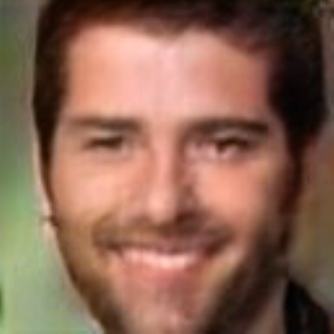}\\
\end{tabular}
}
\vspace{-0.3cm}
\caption{Reconstructions on $128\times 128$ CelebA images for a progressive GAN trained for 10k iterations.}
\label{fig:celebacomp}
\vspace{-0.3cm}
\end{figure}

\vspace{-0.2cm}
\subsection{Synthetic Data}
\vspace{-0.2cm}
To illustrate the advantage of our proposed method over the common baseline, we first demonstrate our results on 2-dimensional synthetic data. Specifically, the  2-dimensional data $x = (x_1, x_2)$ is drawn from a mixture of five equally weighted Gaussians each with a variance of $0.02$, the means of which are spaced equally on the unit circle. See the blue points in columns (a), (b), and (d) of \figref{fig:toycomp} for an illustration. 

In this experiment we aim to reconstruct ${x} = (x_1, x_2)$, given $x_o = x_2 = 0$. Considering the generator has learned the synthetic data very well, the optimal solution for the reconstruction is $\hat{{x}} = (1, 0)$, where the reconstruction error should be 0. However, as discussed  in reference to \figref{fig:toysgddiff} earlier, we observe that energy barriers in the $\cZ$-space complicate optimization. Specifically, if we initialize optimization with a sample far from the optimum, it is  hard to recover. 
While the  strategy to pick the best initializer from a set of $5,000$ points works reasonably well in the low-dimensional setting, it obviously breaks down quickly if the latent space dimension increases even moderately. 

In contrast, our proposed AIS co-generation method only requires one initialization to achieve the desired result after $6,000$ AIS loops, as shown in \figref{fig:toycomp} (15000 (d)). Specifically, reconstruction with generators trained for a different number of epochs (500, 1.5k, 2.5k and 15k) are shown in the rows. The samples obtained from the generator for the data (blue points in column (a)) are illustrated in column (a) using black color. Using the respective generator to solve the program given in \equref{eq:backprop} via GD yields results highlighted with yellow color in column (b). The empirical reconstruction error frequency for this baseline is given in column (c). The results and the reconstruction error frequency obtained with \algref{alg:ours} are shown in columns (d, e). We observe significantly better results and robustness to initialization. 

\begin{figure}[t]
\centering
\scalebox{0.9}{
\begin{tabular}{ccccc}
  \multicolumn{1}{c}{{\small Ground Truth}}&\multicolumn{1}{c}{{\small Masked Image}}&\multicolumn{1}{c}{{\small GD+single $z$}}& \multicolumn{1}{c}{{\small GD+multi $z$}}&\multicolumn{1}{c}{{\small AIS}}
\\
  \includegraphics[width=20mm]{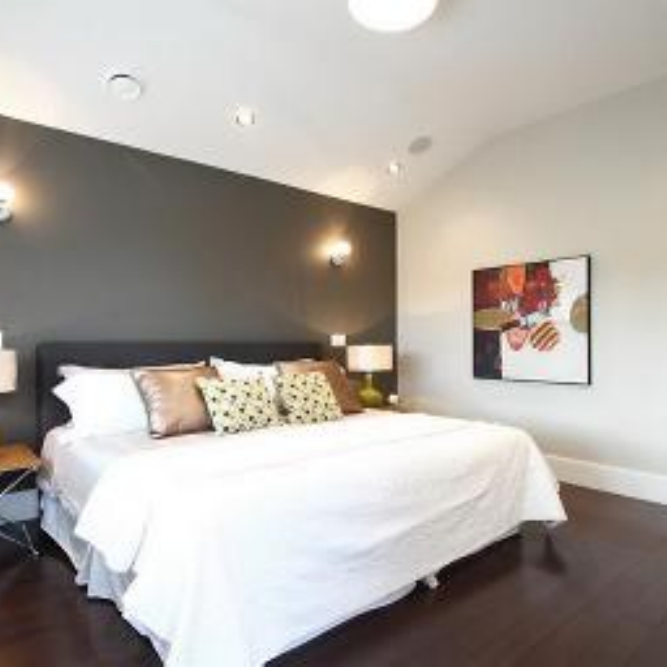}&\includegraphics[width=20mm]{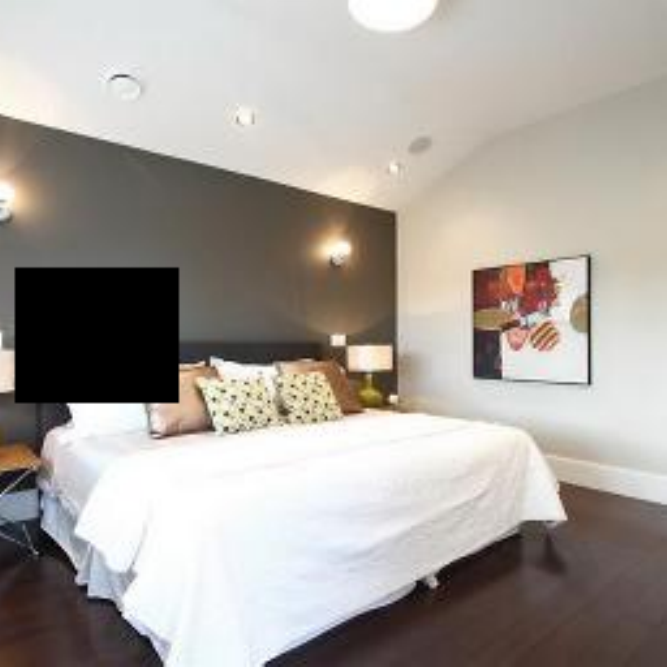}&\includegraphics[width=20mm]{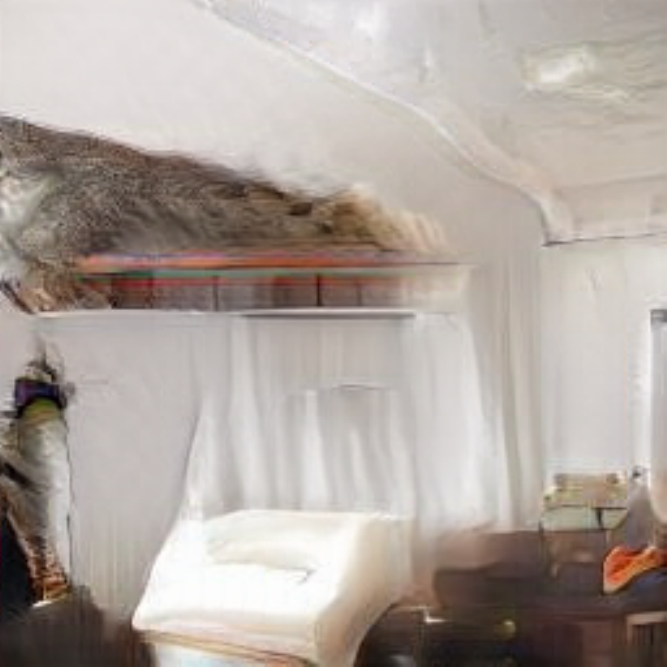}&\includegraphics[width=20mm]{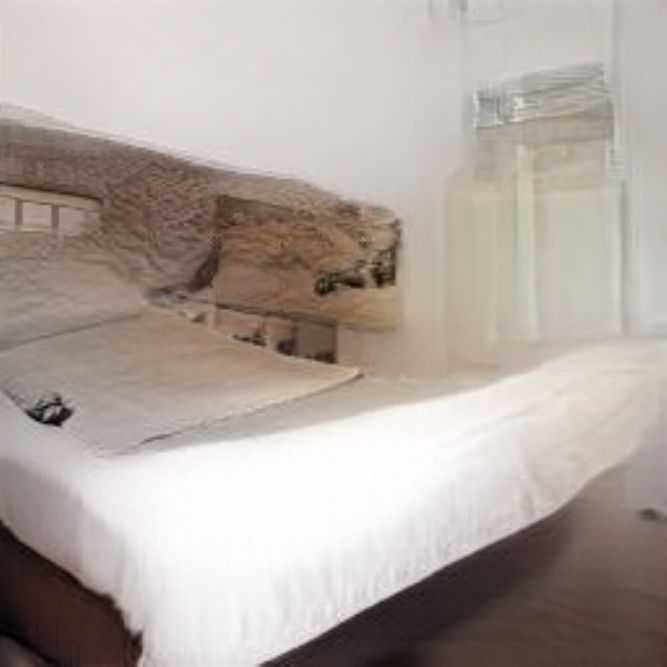}&\includegraphics[width=20mm]{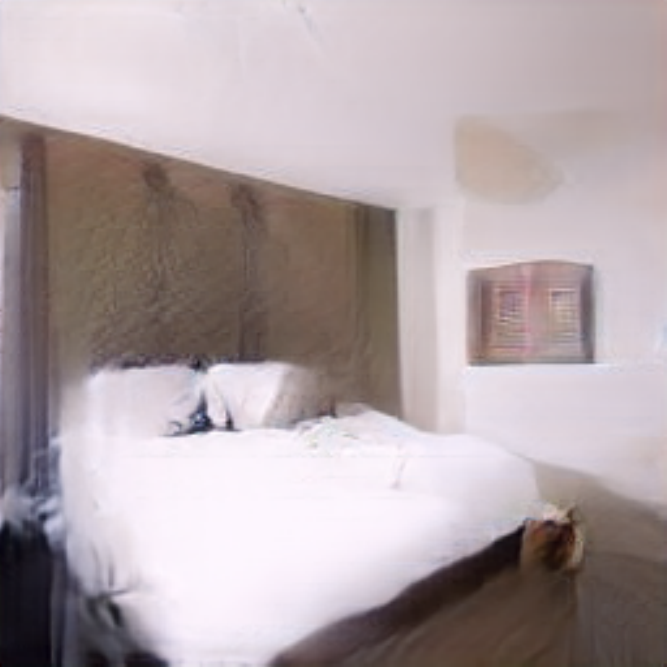}\\
   \includegraphics[width=20mm]{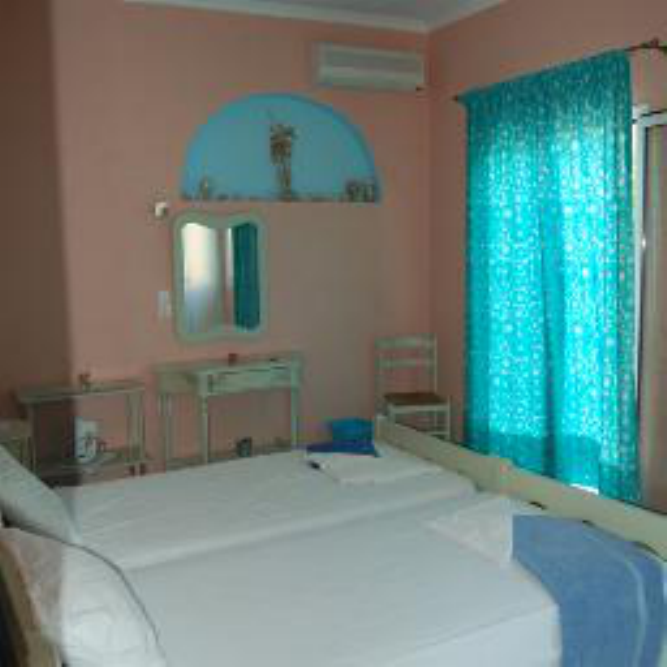}&\includegraphics[width=20mm]{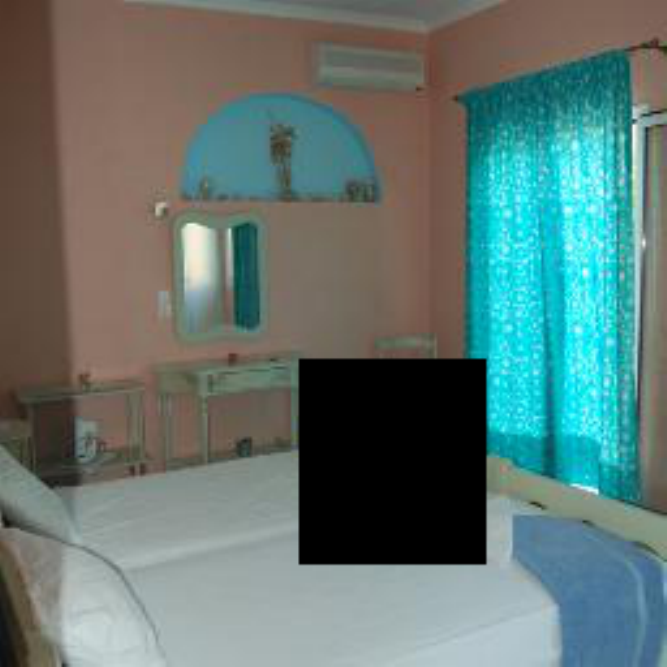}&\includegraphics[width=20mm]{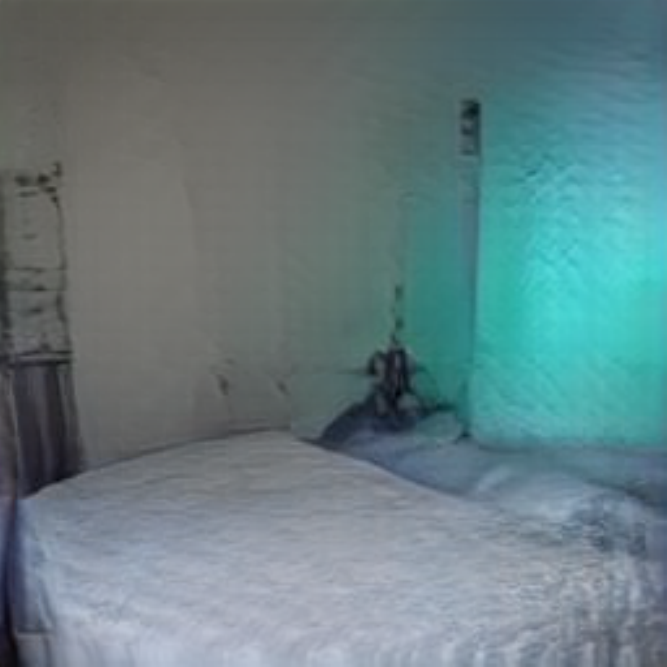}&\includegraphics[width=20mm]{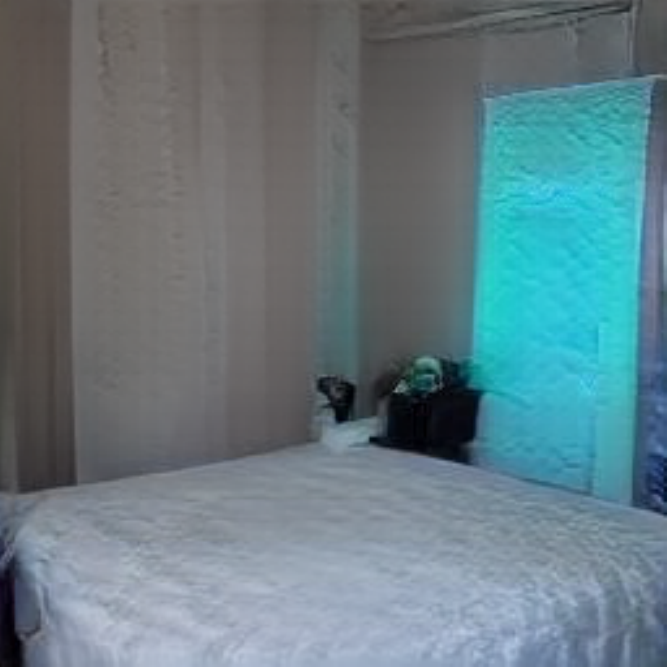}&\includegraphics[width=20mm]{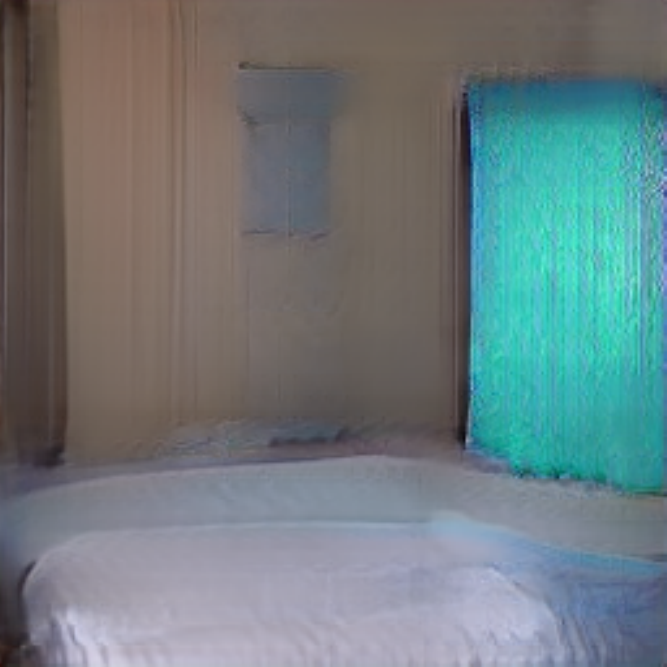}\\
   \includegraphics[width=20mm]{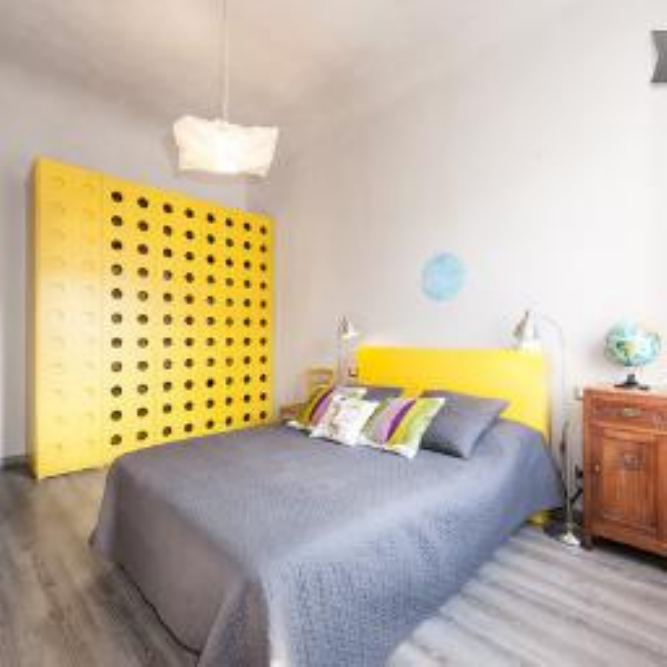}&\includegraphics[width=20mm]{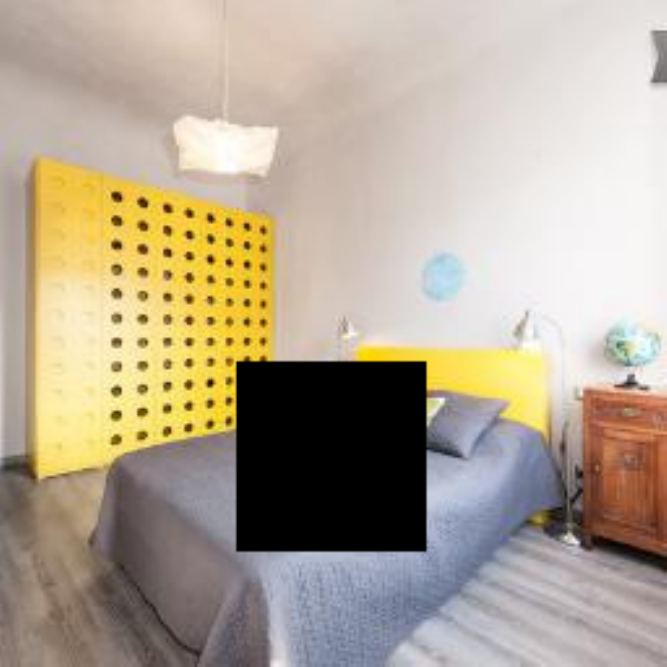}&\includegraphics[width=20mm]{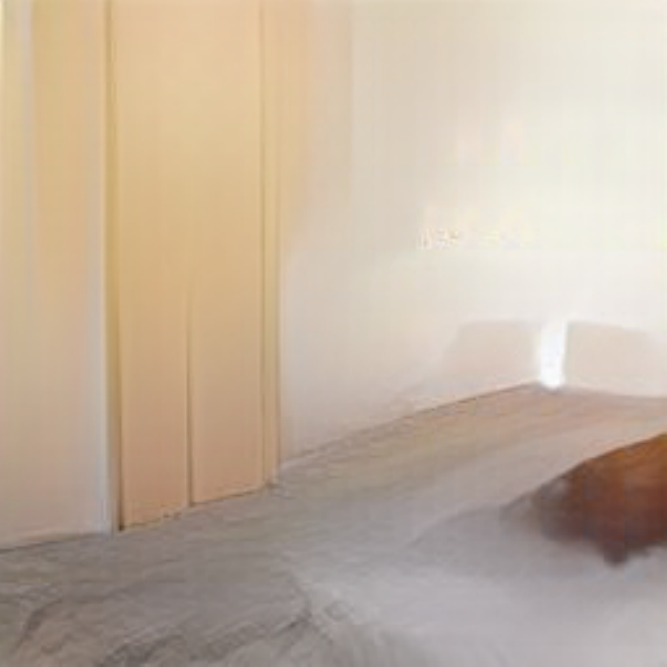}&\includegraphics[width=20mm]{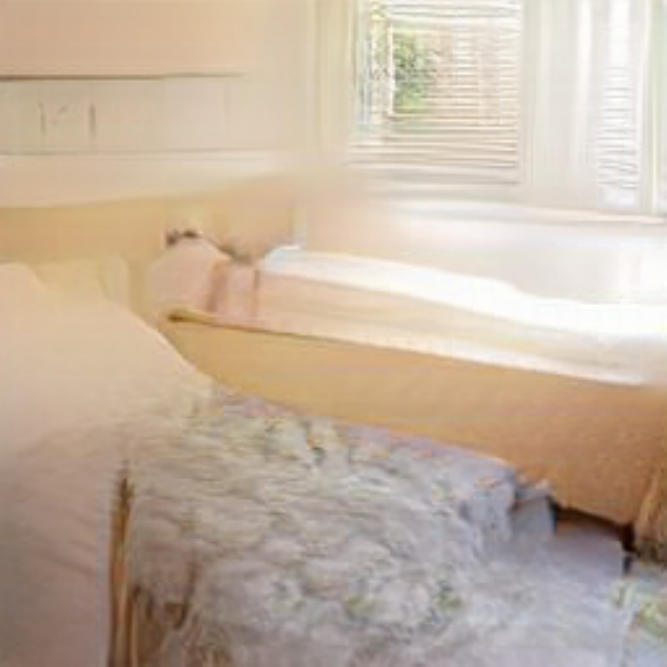}&\includegraphics[width=20mm]{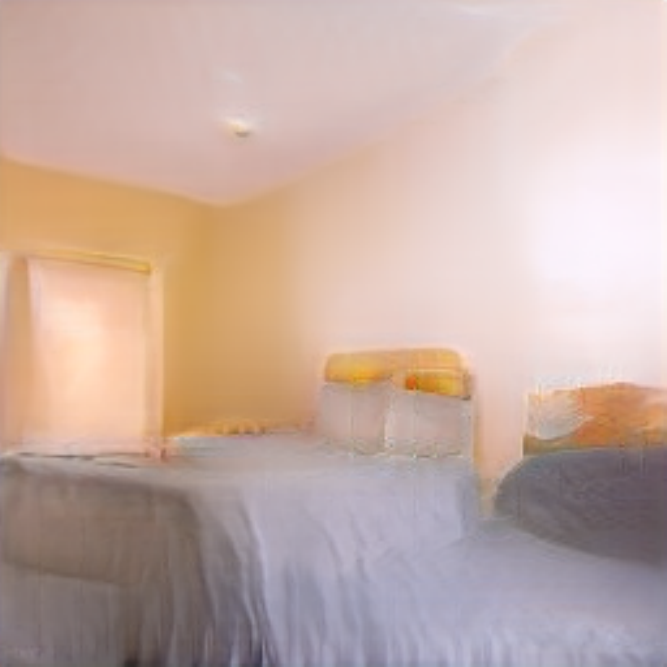}\\ 
\end{tabular}
}
\vspace{-0.3cm}
\caption{Reconstructions on $256 \times 256$ LSUN images using a pre-trained progressive GAN trained for 10k iter.}
\label{fig:lsuncomp}
\end{figure}

\begin{figure}[t]
\centering
\setlength{\tabcolsep}{2pt}
 \begin{tabular}{cccccccc}
  \multicolumn{0}{c}{{\small LR}} & \multicolumn{0}{c}{{\small GD+single $z$}} & \multicolumn{0}{c}{{\small GD+multi $z$}} & \multicolumn{0}{c}{{\small AIS}} & \multicolumn{0}{c}{{\small LR}} & \multicolumn{0}{c}{{\small GD+single $z$}} & \multicolumn{0}{c}{{\small GD+multi $z$}} & \multicolumn{0}{c}{{\small AIS}}
\\
  \includegraphics[width=15mm]{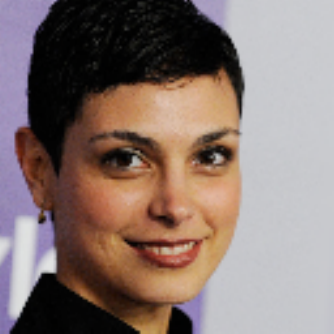}&\includegraphics[width=15mm]{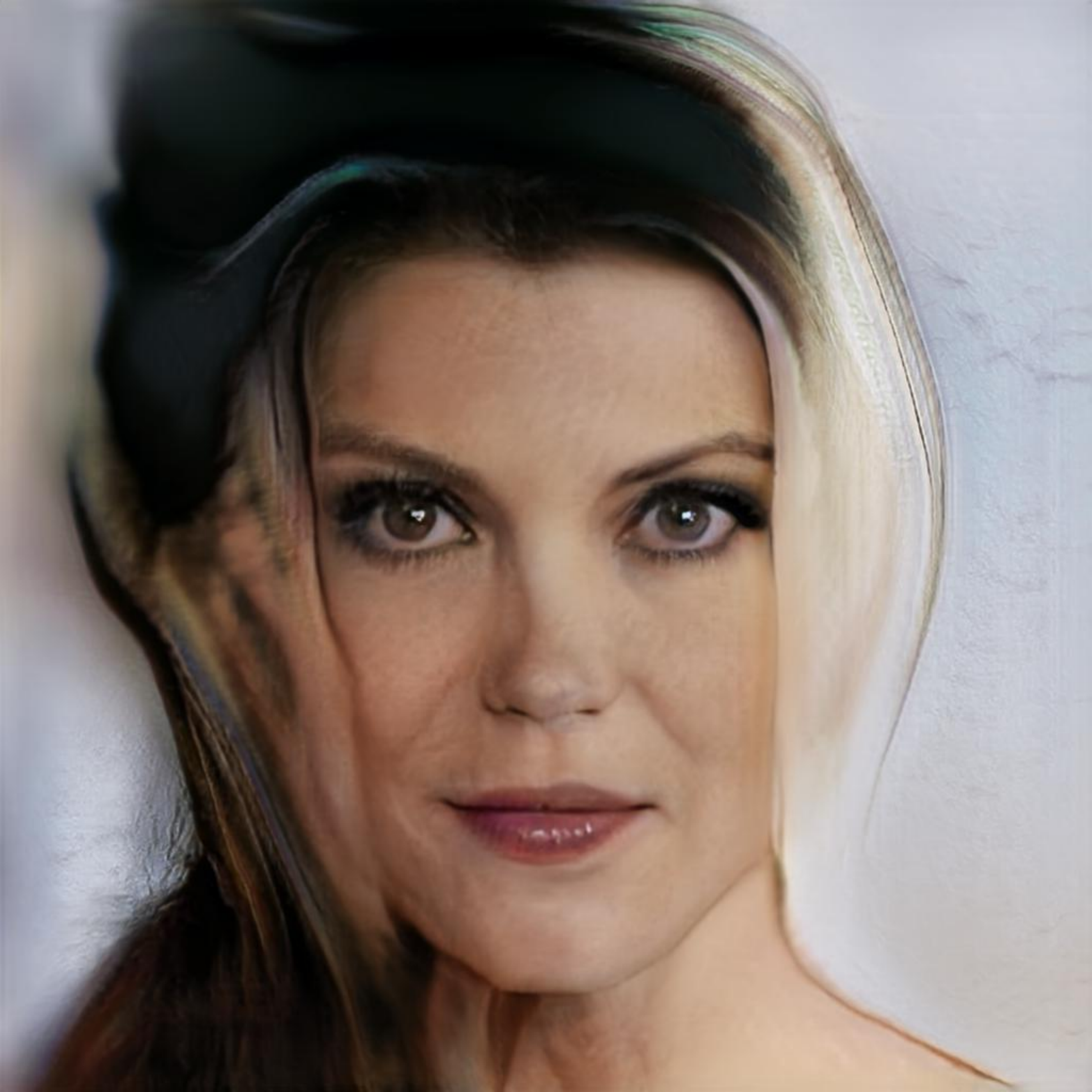}&\includegraphics[width=15mm]{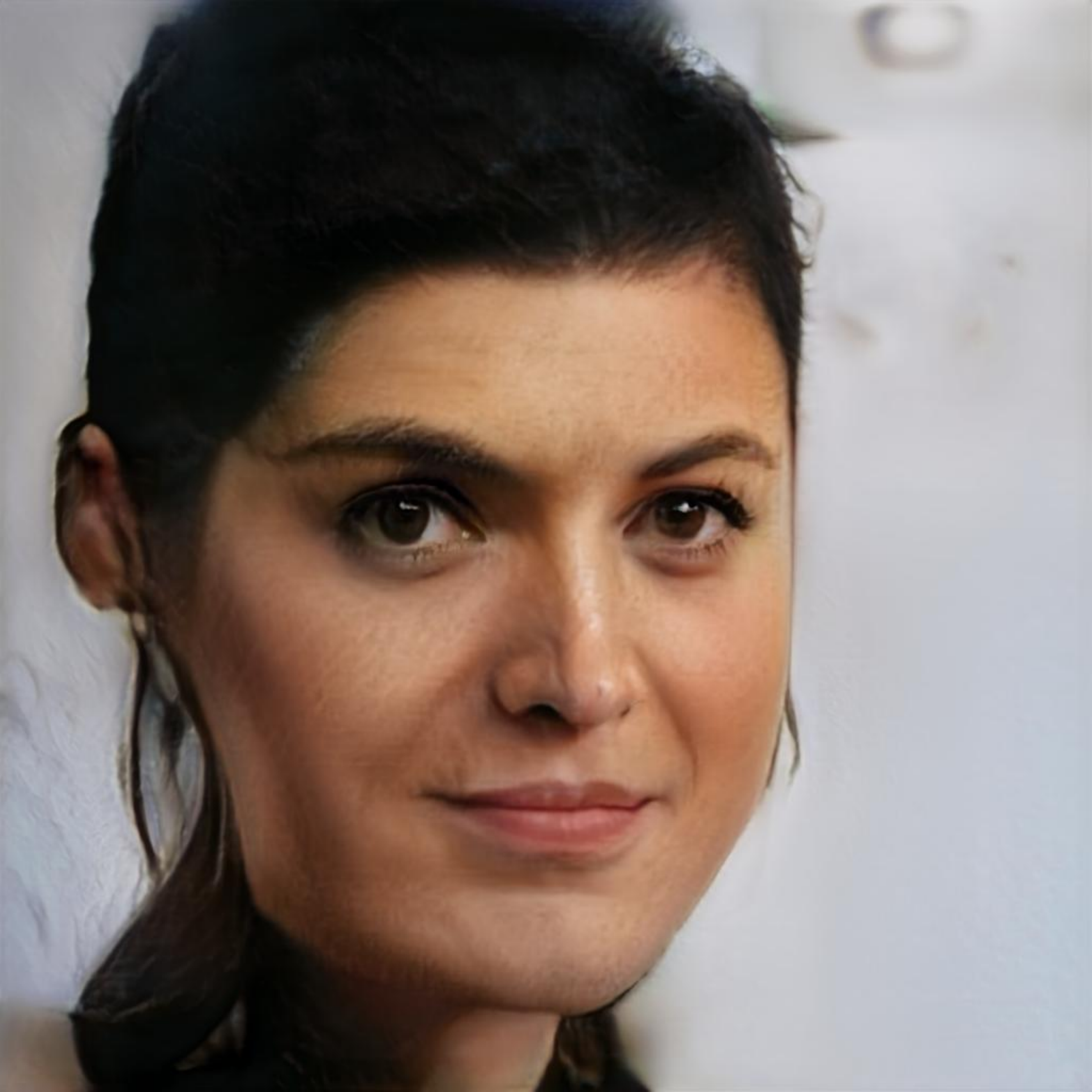}&\includegraphics[width=15mm]{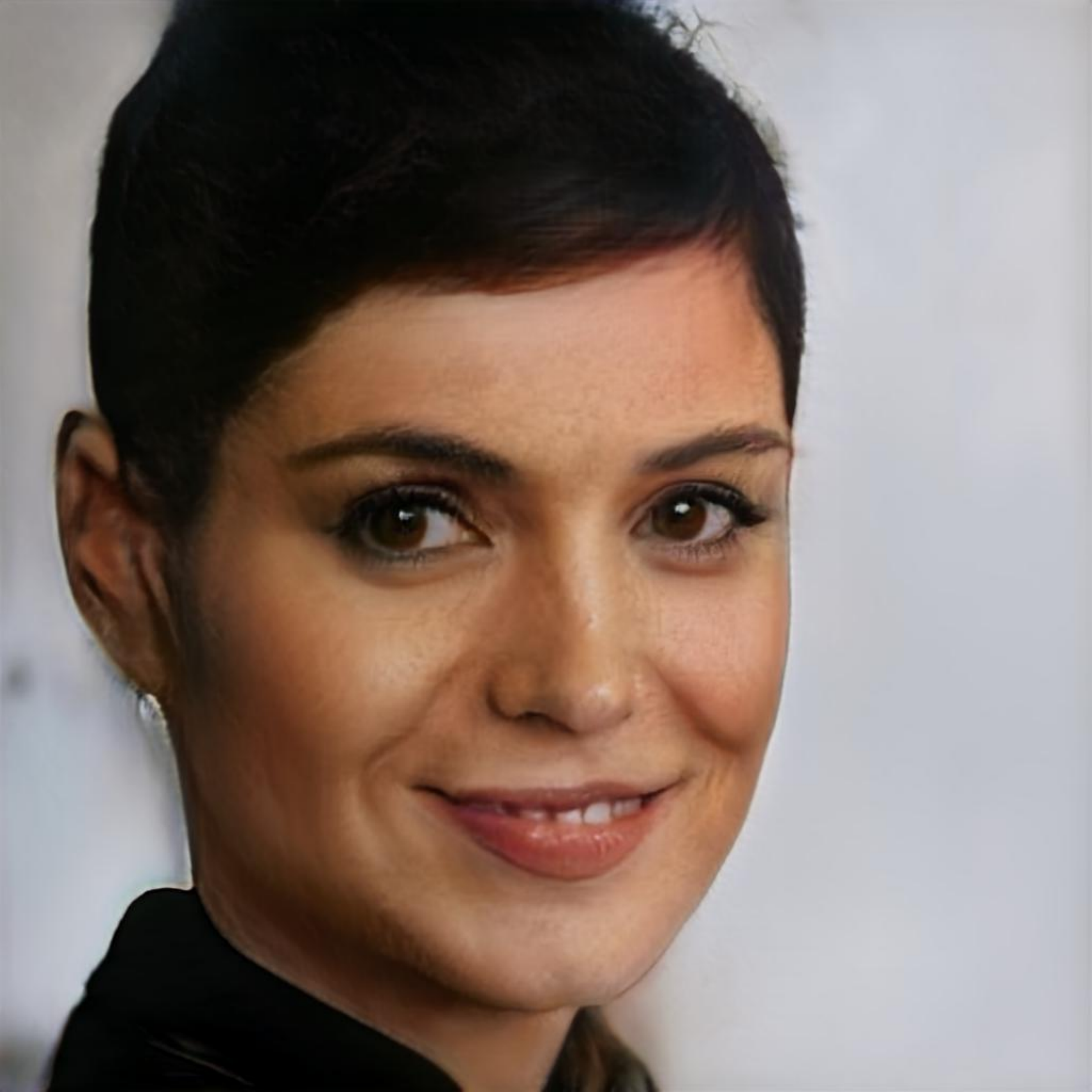}&\includegraphics[width=15mm]{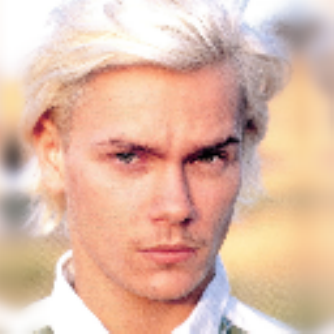}&\includegraphics[width=15mm]{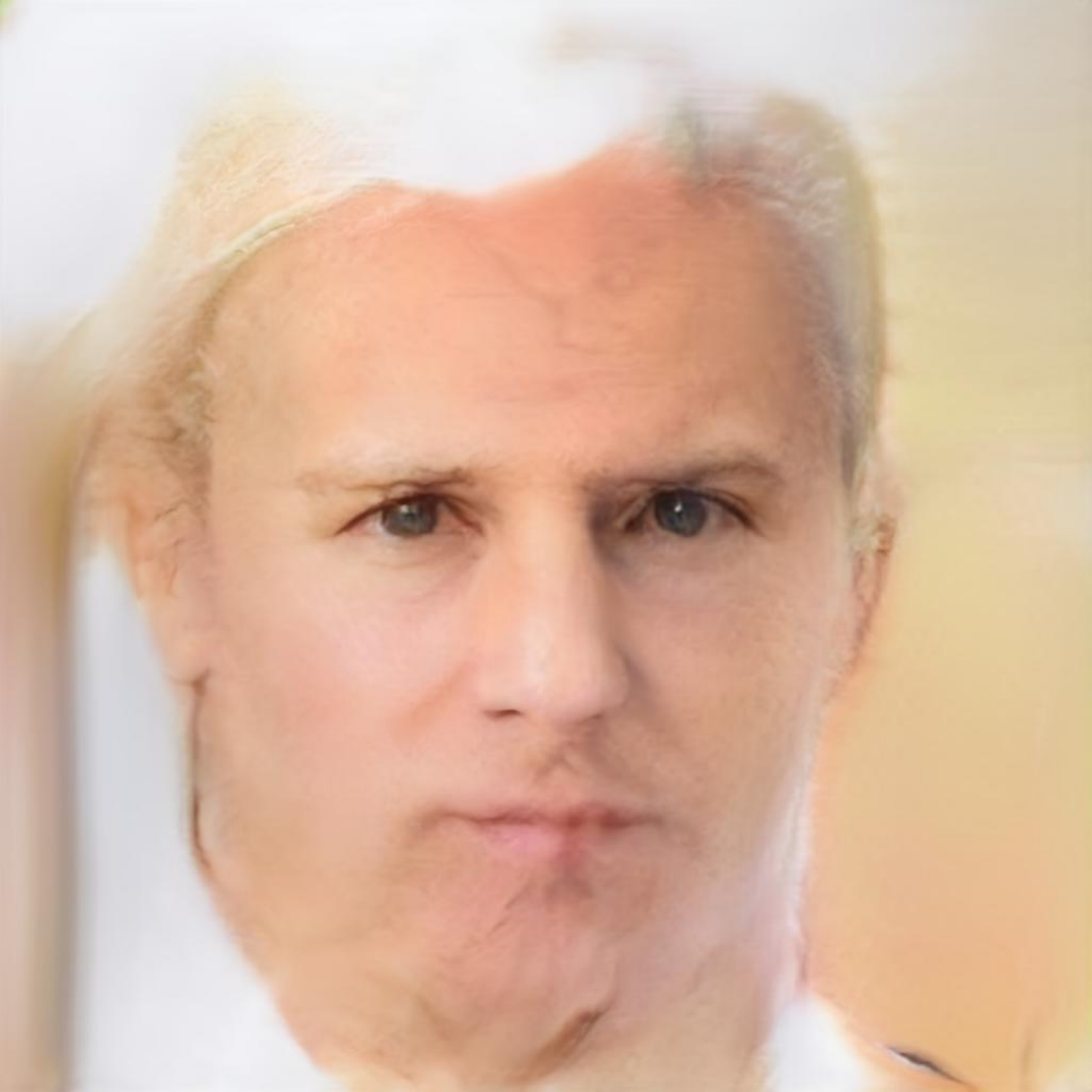}&\includegraphics[width=15mm]{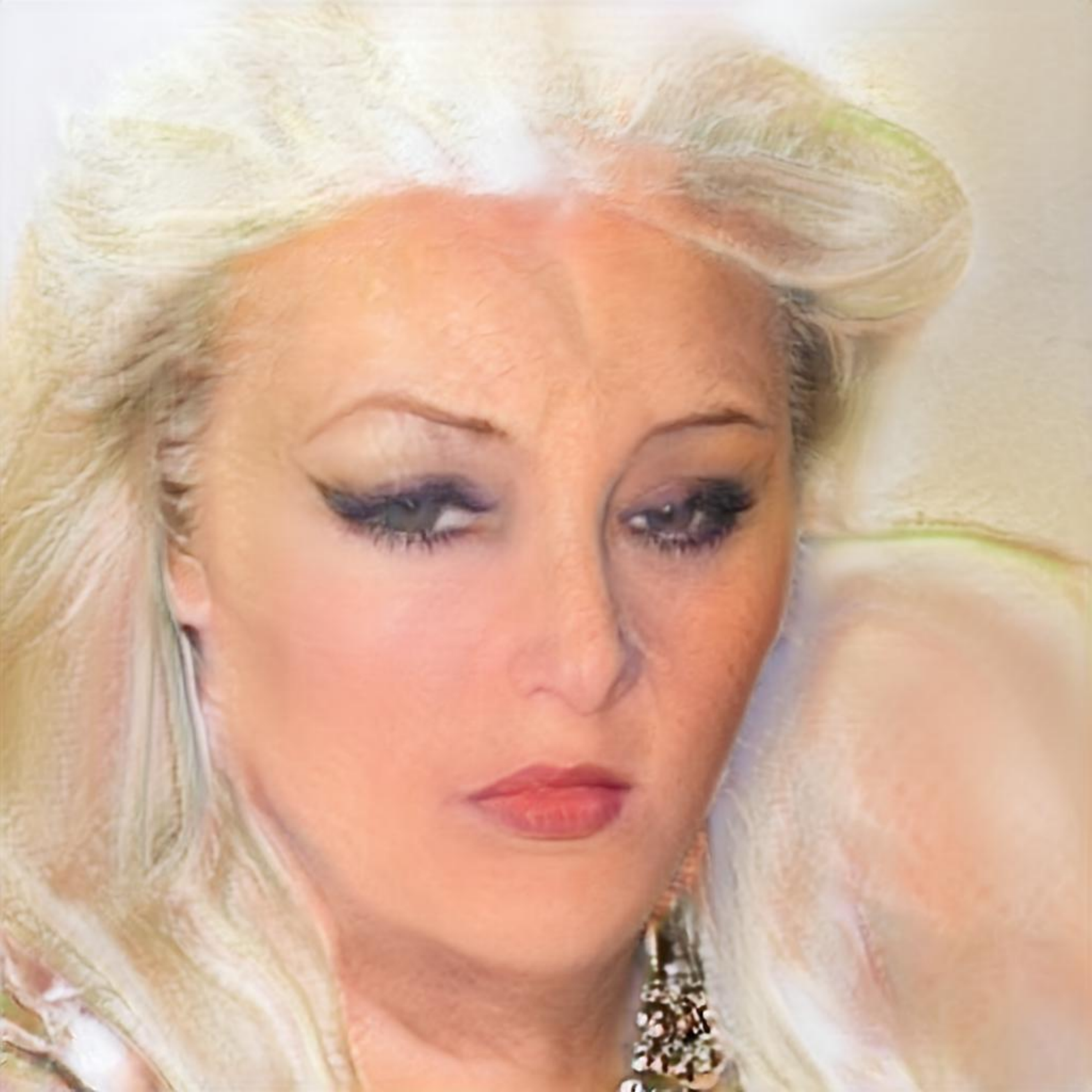}&\includegraphics[width=15mm]{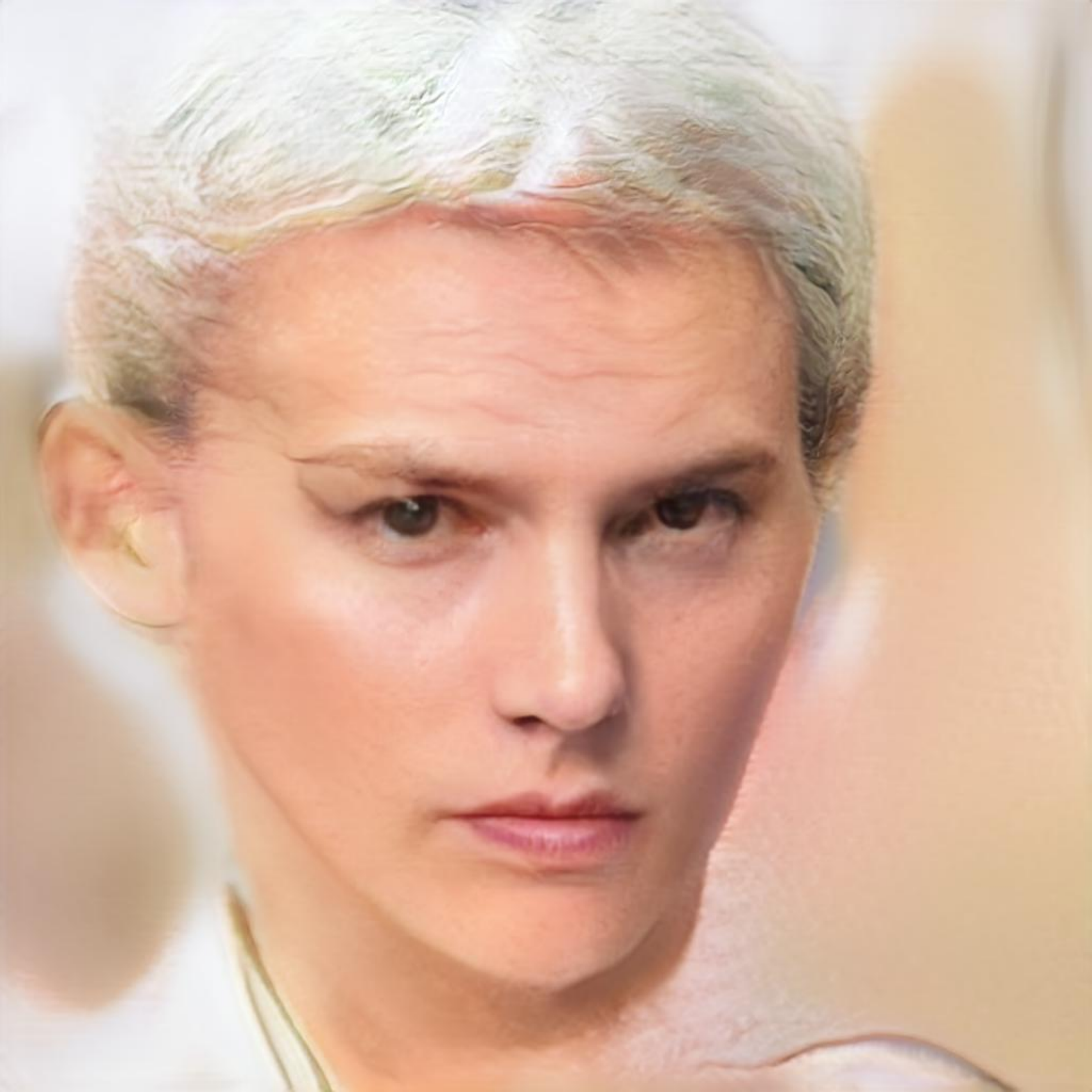}\\
  
   \includegraphics[width=15mm]{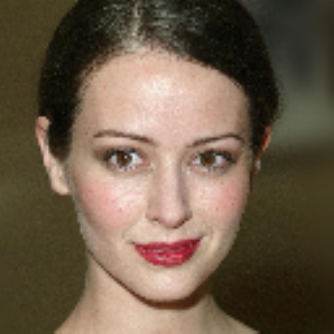}&\includegraphics[width=15mm]{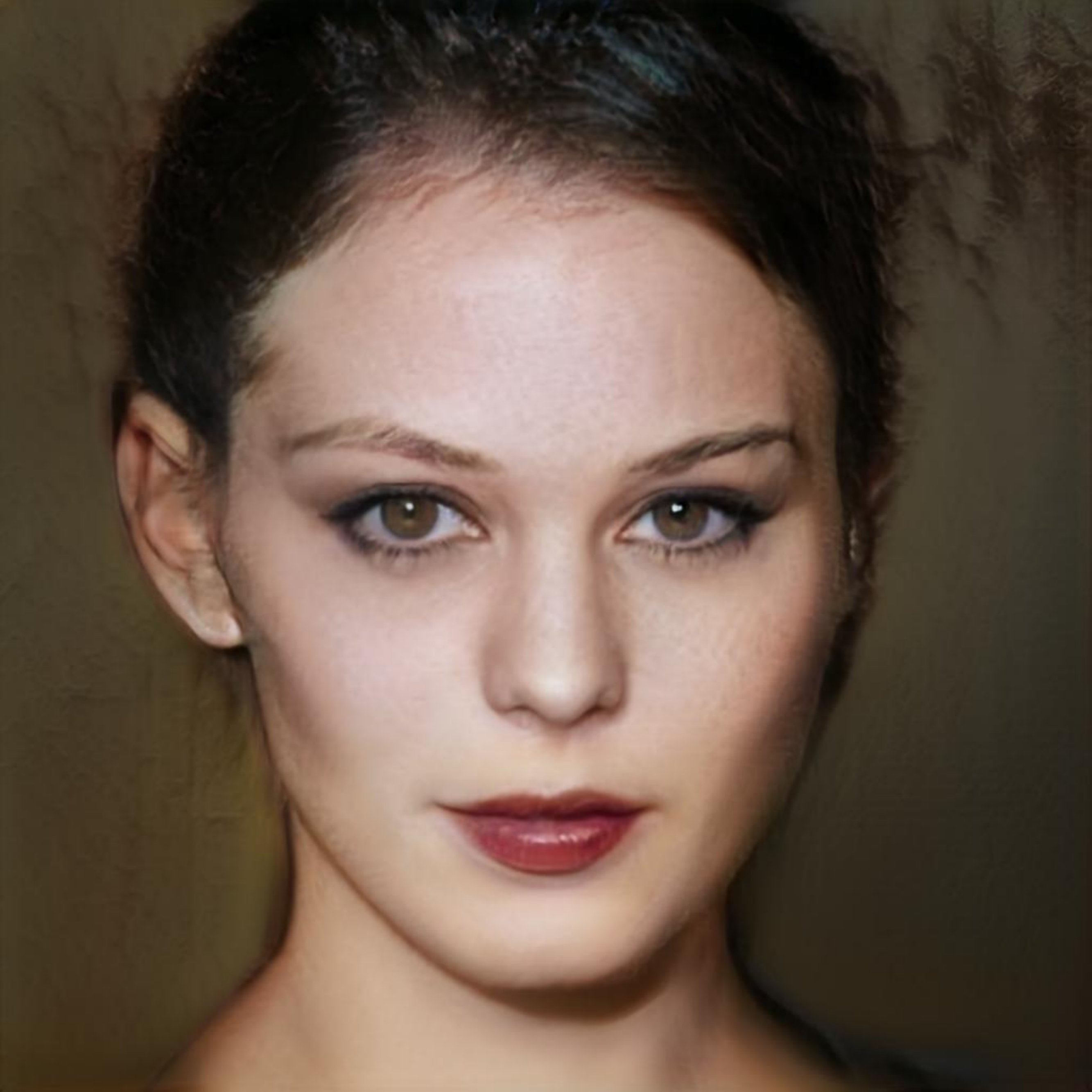}&\includegraphics[width=15mm]{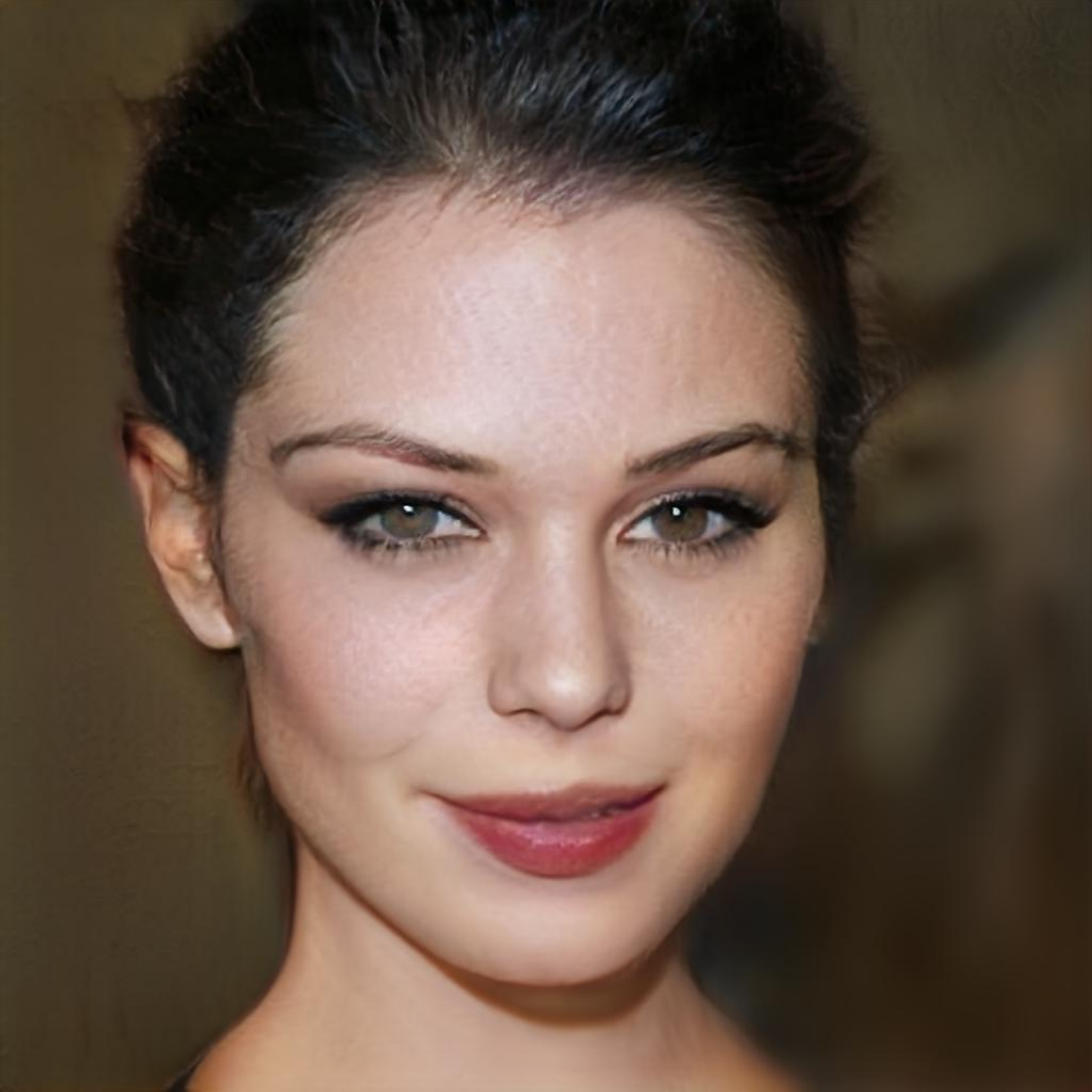}&\includegraphics[width=15mm]{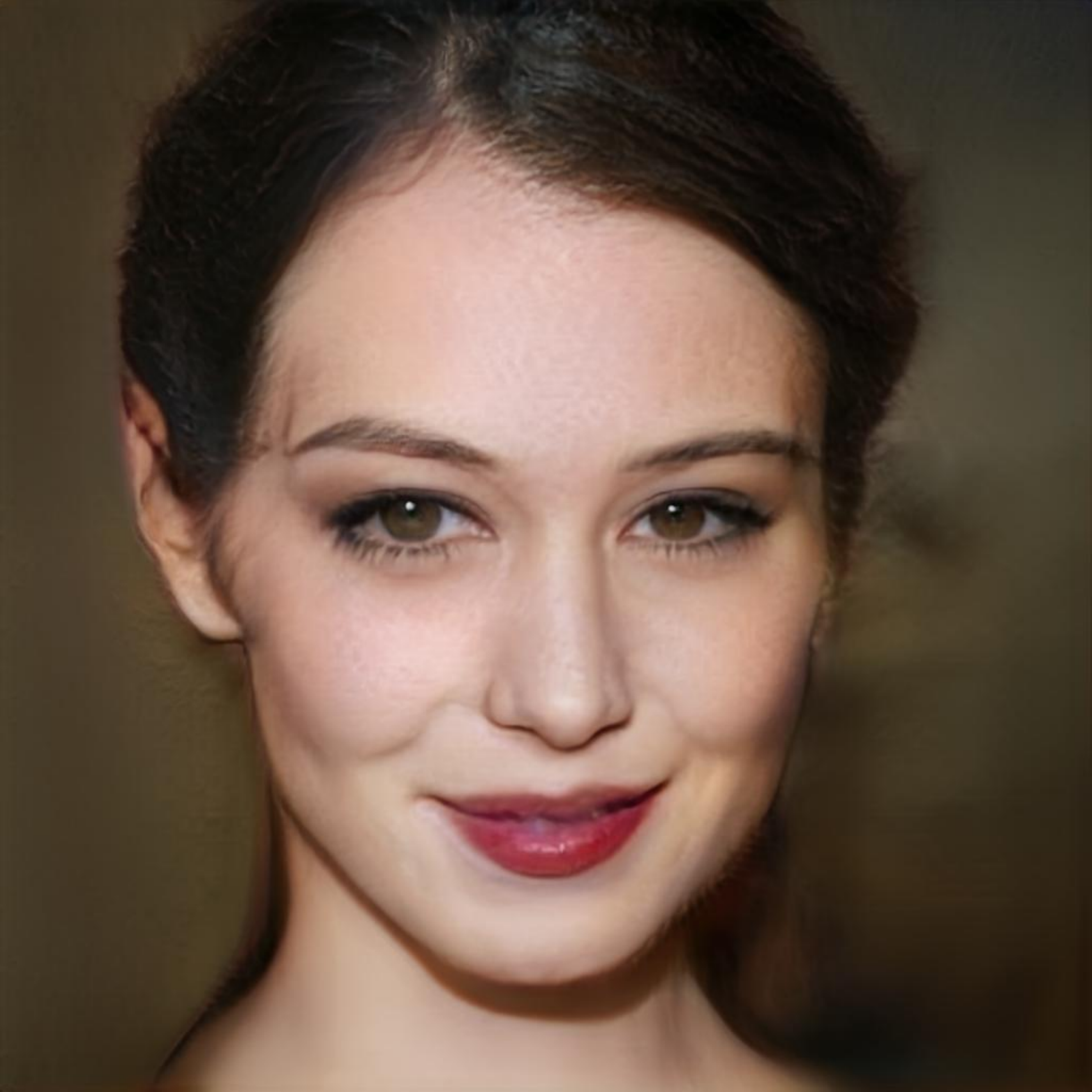}&\includegraphics[width=15mm]{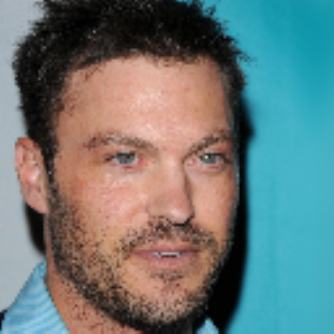}&\includegraphics[width=15mm]{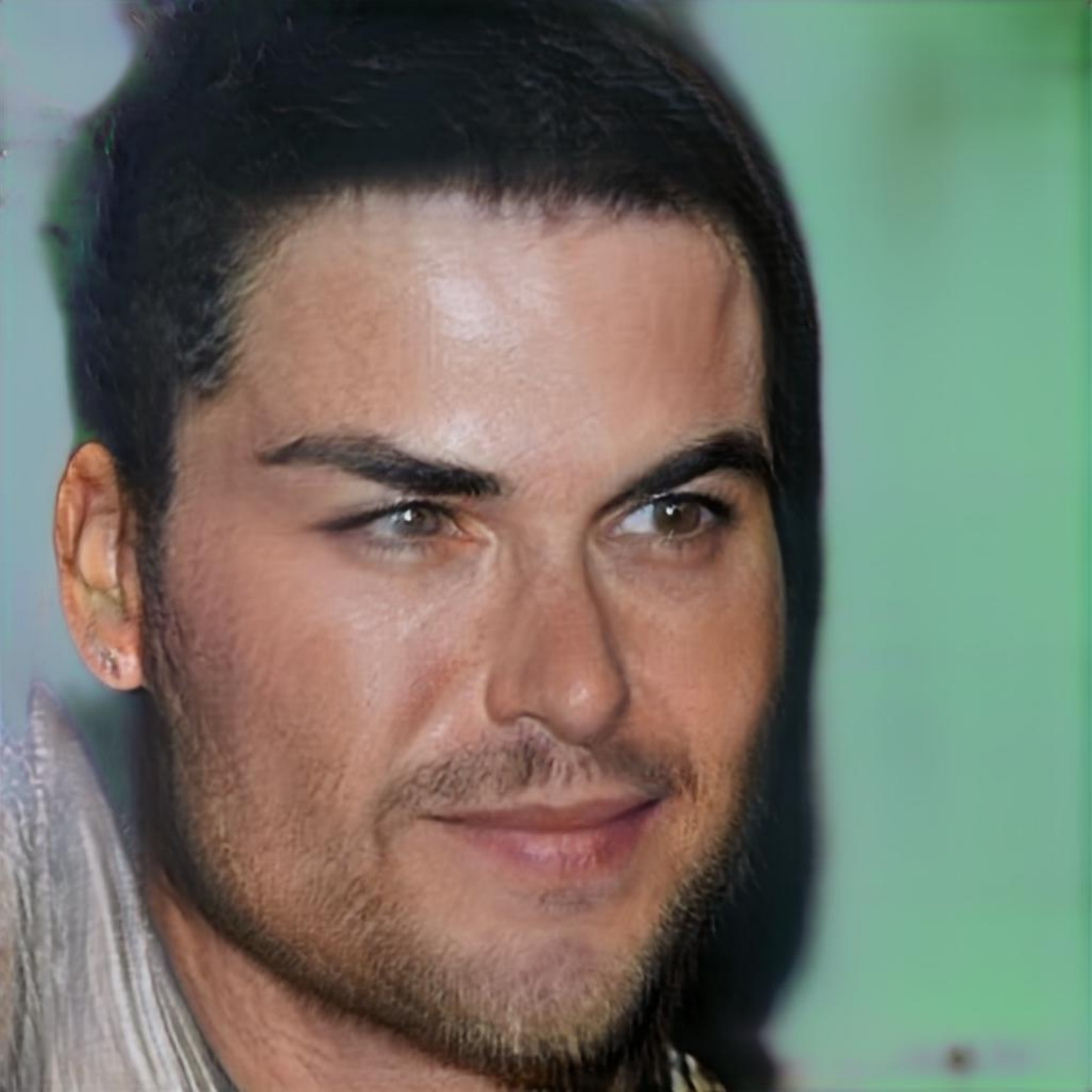}&\includegraphics[width=15mm]{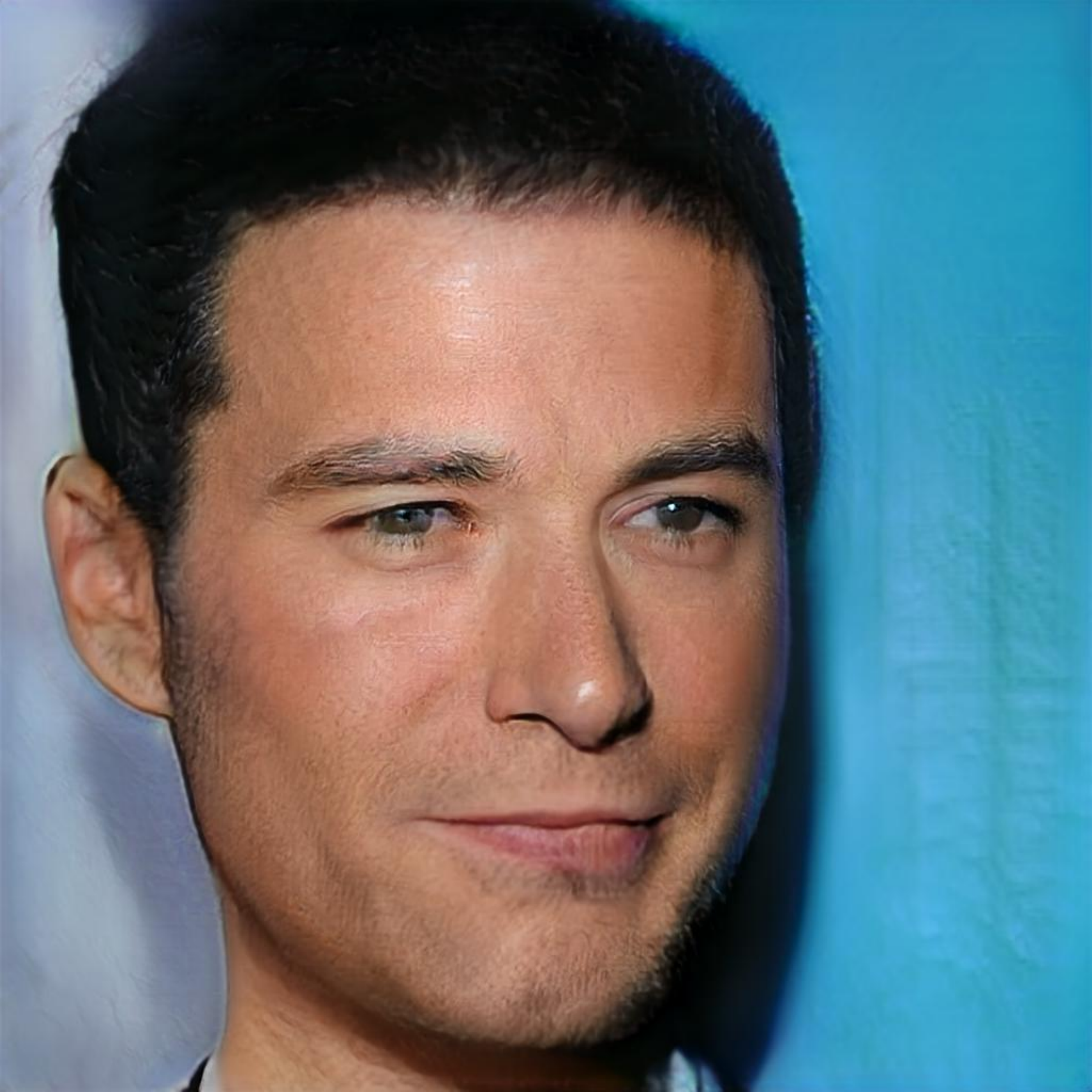}&\includegraphics[width=15mm]{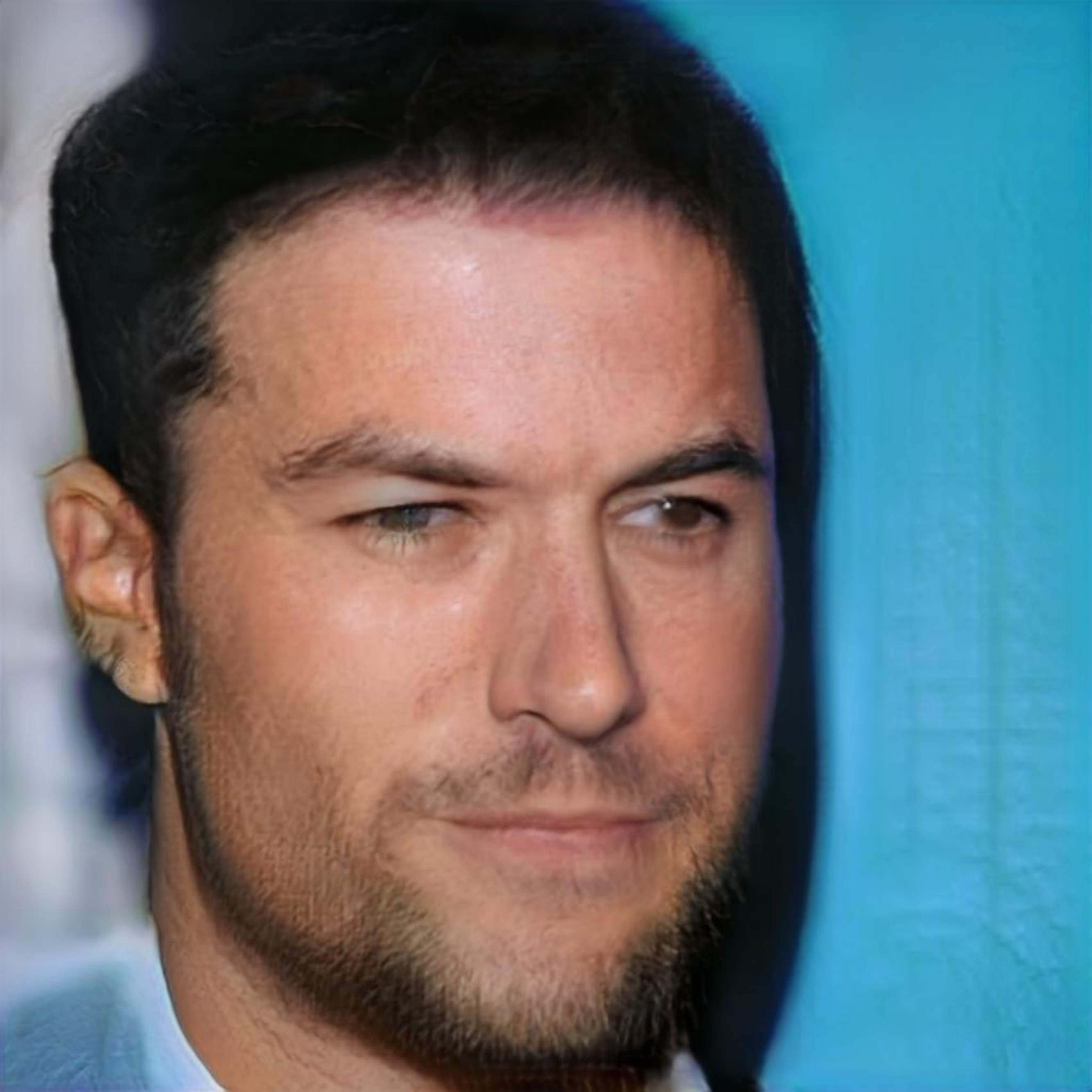}\\
\end{tabular}
\vspace{-0.3cm}
\caption{SISR: $128\times 128$ to $1024\times 1024$ for CelebA-HQ images using a progressive GAN (19k iter.).}
\label{fig:celebahqcomp}
\vspace{-0.4cm}
\end{figure}

In \figref{fig:aisprocedure} we show for 100 samples that \algref{alg:ours}  moves them across the energy barriers during the annealing procedure, illustrating the benefits of AIS based HMC over GD.

\vspace{-0.2cm}
\subsection{Imaging Data}
\vspace{-0.2cm}
To validate our method on real data, we evaluate  on three datasets, using MSE and MSSIM metrics. For all three experiments, we  use the progressive GAN architecture~\cite{karras2017progressive} and evaluate baselines and AIS on progressive GAN training data.

\textbf{CelebA:} 
For CelebA, the size of the input and the output are 512  and  $128\times 128$ respectively. We generate corrupted images by randomly masking blocks of width and height ranging from 30 to 60. Then we use \algref{alg:ours} for reconstruction with 500 HMC loops. 

In \figref{fig:metric} (a,b), we observe that \algref{alg:ours} outperforms  both baselines for all GAN training iterations on both MSSIM and MSE metrics. The difference increases for better trained generators. In \figref{fig:celebacomp}, we show some results generated by both baselines and \algref{alg:ours}. Compared to baselines, \algref{alg:ours} results are more similar to the ground truth and more robust to different mask locations.  Note that \algref{alg:ours} only uses one initialization, which demonstrates its robustness  to  initialization. 

\textbf{LSUN: }
The output size is 256$\times$256. We mask images with blocks of width and height between 50 to 80. The complex distribution and intricate details of LSUN challenge the reconstruction. Here, we sample 5 initializations in our \algref{alg:ours} (line 2). We use 500 HMC loops for each initialization independently. For each image, we pick the best score among five and show the average in \figref{fig:metric} (c,d). We observe that \algref{alg:ours} with 5 initializations easily outperforms GD with 5,000 initializations. We also show reconstructions in \figref{fig:lsuncomp}.

\textbf{CelebA-HQ}
Besides recovering masked images, we also demo co-generation on single image super-resolution (SISR). In this task, the ground truth is a high-resolution image $x$ ($1024\times 1024$) and the exposure information $x_e$ is a low-resolution image ($128 \times 128$). Here, we use the Progressive CelebA-HQ GAN as the generator. After obtaining the generated high-resolution image, we downsample it to $128\times 128$ via pooling and aim to reduce the squared error between it and the final result. We use 3 $z$ samples for the SISR task. We show MSSIM and MSE between the ground truth ($1024 \times 1024$) and the final output on \figref{fig:metric} (e, f). \figref{fig:celebahqcomp} compares the outputs of baselines to those of \algref{alg:ours}.

\vspace{-0.3cm}
\section{Conclusion}
\label{sec:conc}
\vspace{-0.3cm}
We propose a co-generation approach, \ie, we complete partially given input data,  using annealed importance sampling (AIS) based on the Hamiltonian Monte Carlo (HMC) method. Different from the classical optimization based methods, specifically GD, which get easily trapped in local optima when solving this task, the proposed approach is much more robust. Importantly, the method can traverse large energy barriers that occur when training generative adversarial nets. Its robustness is due to AIS gradually annealing a  probability distribution and HMC avoiding localized walks. 

\noindent\textbf{Acknowledgments:}
This work is supported in part by NSF under Grant No.\ 1718221 and MRI \#1725729, UIUC, Samsung, 3M, Cisco Systems Inc.\ (Gift Award CG 1377144) and Adobe. We thank NVIDIA for providing GPUs used for this work and Cisco for access to the Arcetri cluster.

\clearpage
{\small
\bibliography{neurips_2019.bbl}
\bibliographystyle{abbrvnat}
}
\clearpage

\section{Appendix: Additional Synthetic Data Analysis}

\begin{figure}[ht]
\centering
\begin{tabularx}{\textwidth} {@{\hskip3pt}c@{\hskip3pt}c@{\hskip3pt}c@{\hskip3pt}c@{\hskip3pt}c@{\hskip3pt}c}
      \rotatebox{90}{\hspace{0.5cm}{\small 15000}} &\includegraphics[width=25mm]{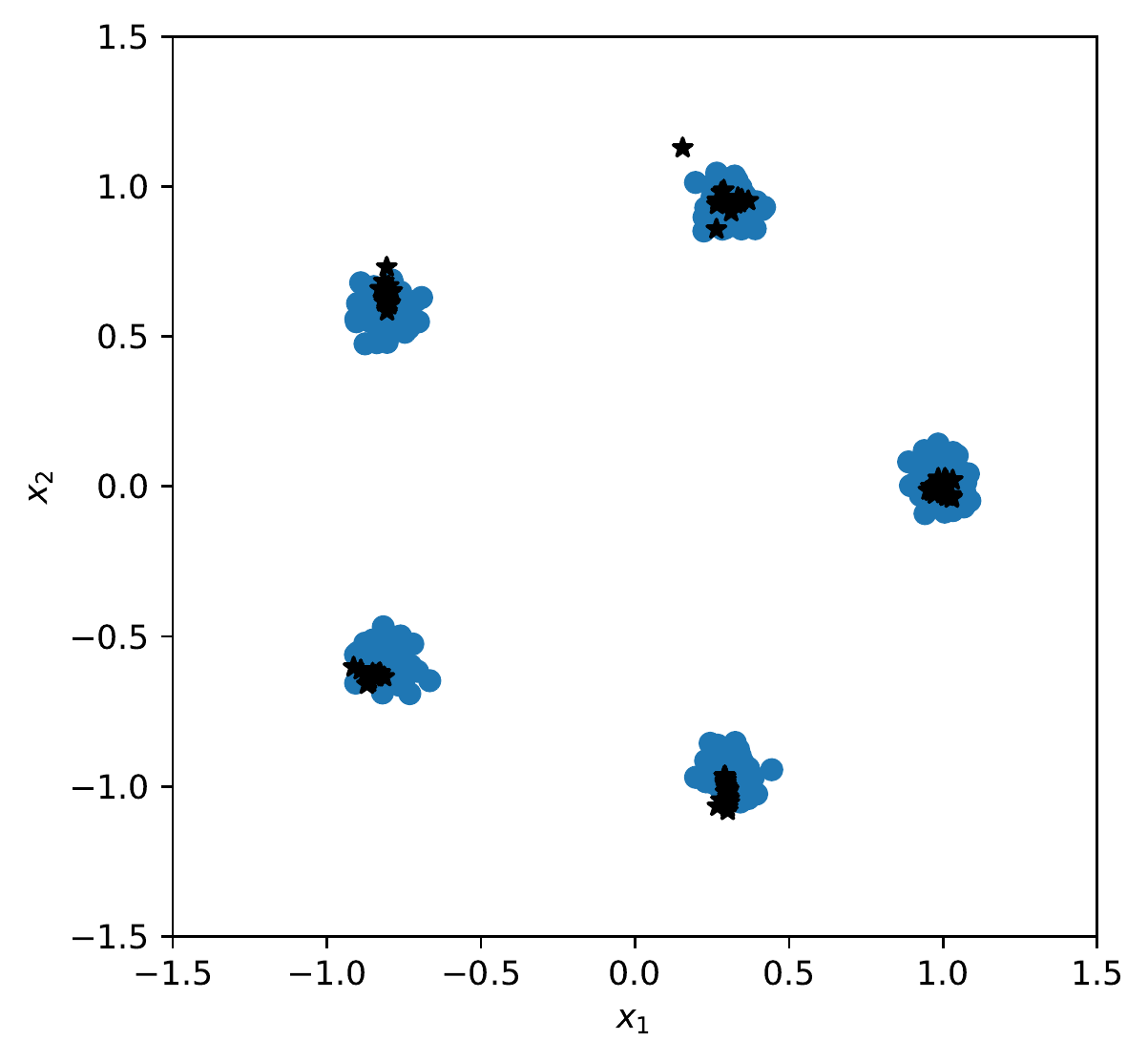}&\includegraphics[width=25mm]{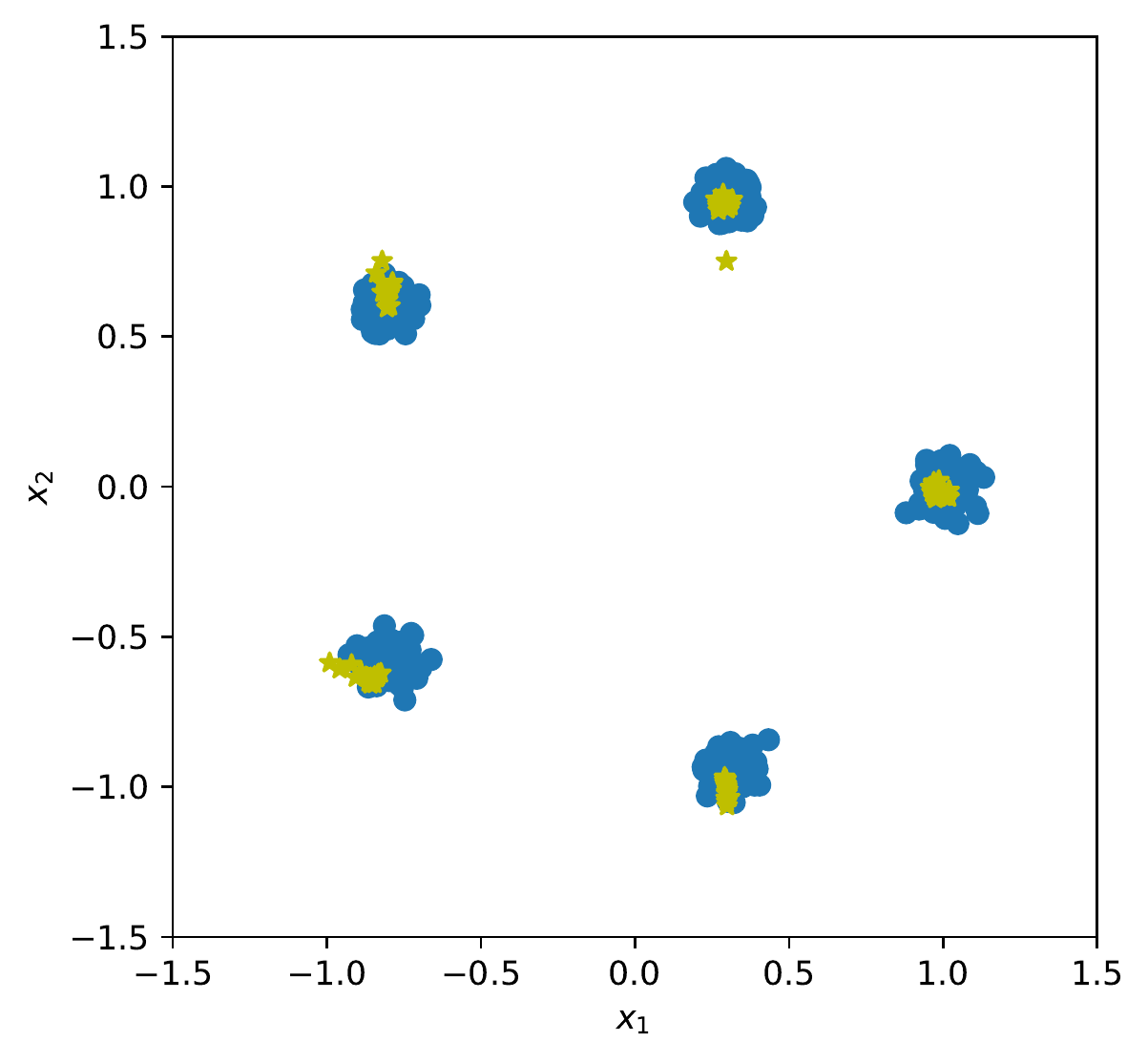}&\includegraphics[width=25mm]{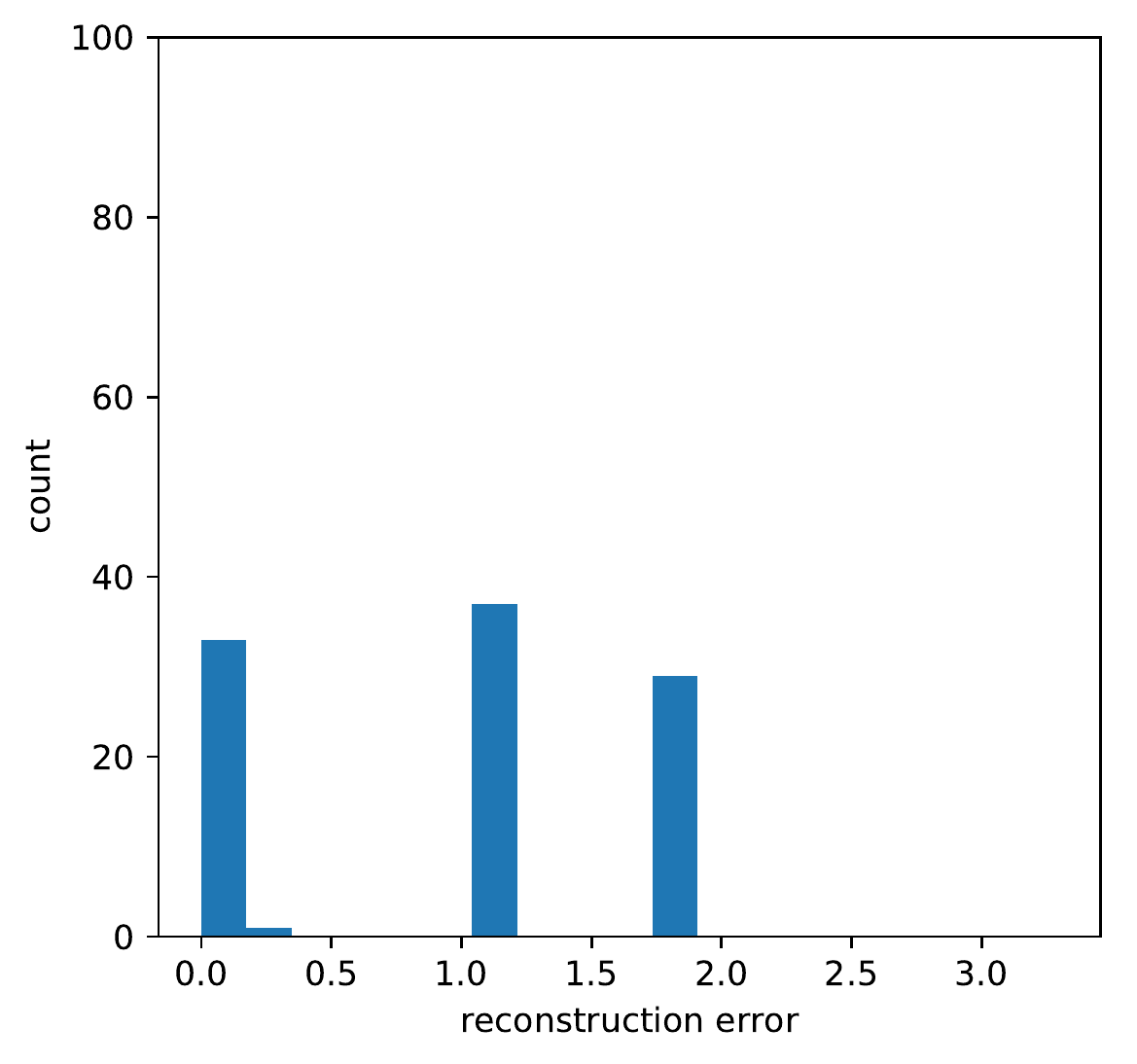}&\includegraphics[width=25mm]{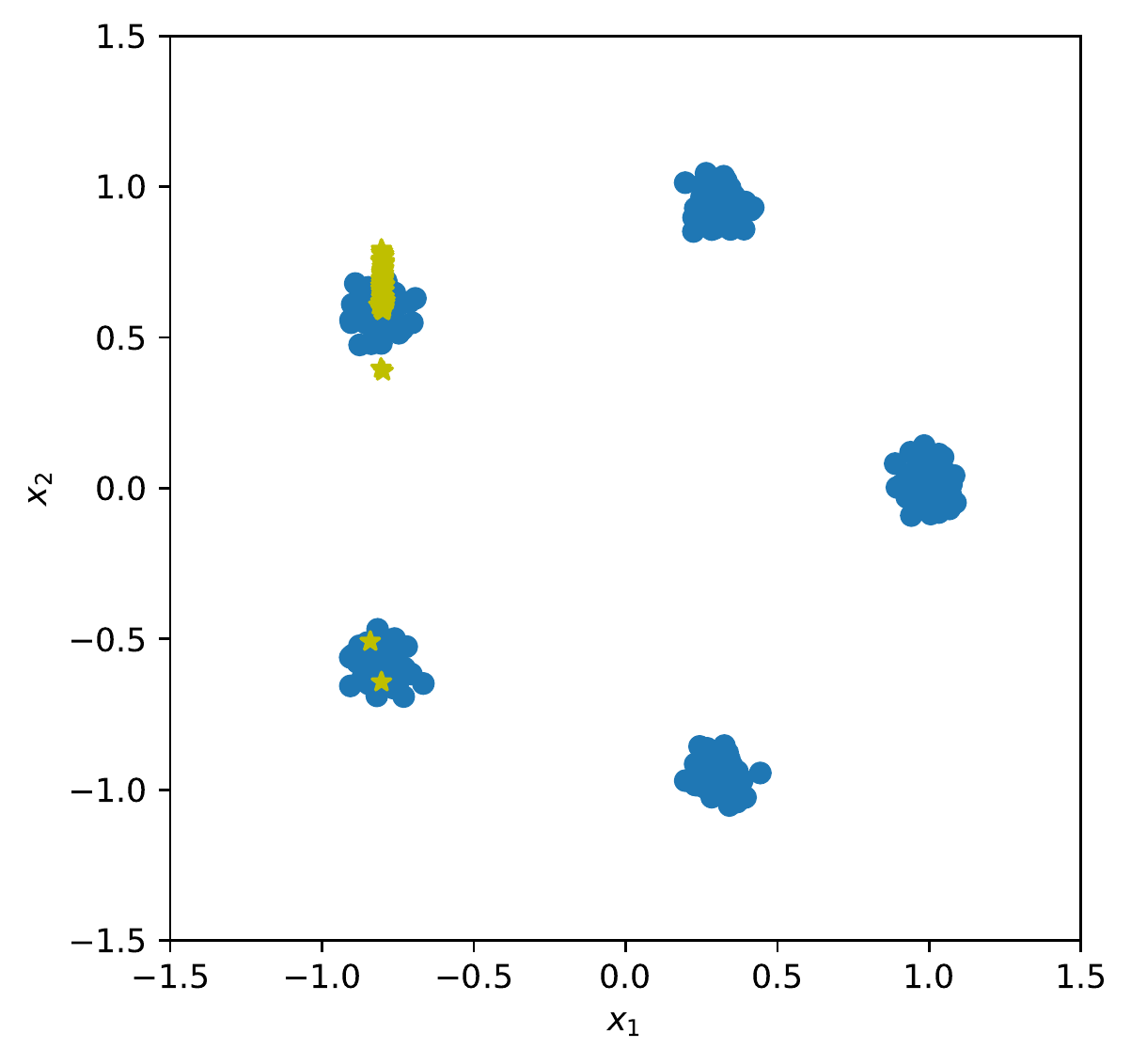}& \includegraphics[width=25mm]{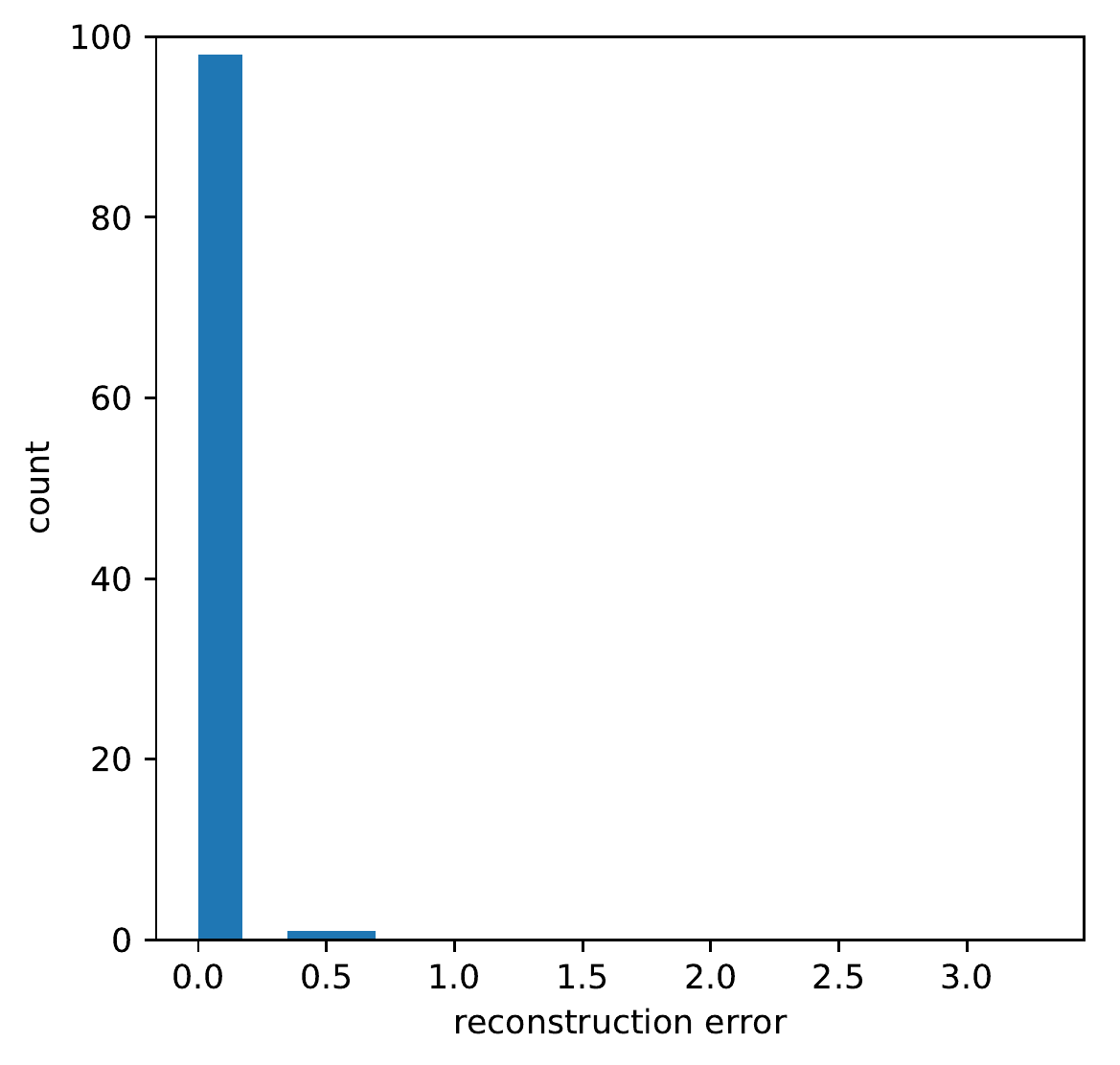}\\[-7pt]
  &\multicolumn{1}{c}{{\scriptsize (a)}} & \multicolumn{1}{c}{{\scriptsize (b)}} & \multicolumn{1}{c}{{\scriptsize (c)}} & \multicolumn{1}{c}{{\scriptsize (d)}} & \multicolumn{1}{c}{{\scriptsize (e)}} 
\end{tabularx}

\caption{(a) Samples generated with a vanilla GAN (black); (b) GD reconstructions from 100 random initializations; (c) Reconstruction error bar plot for the result in column (b); (d) Reconstructions recovered with Alg.~1; (e) Reconstruction error bar plot for the results in column (d). 
}

\label{fig:ambigres}
\end{figure}

\begin{figure}[ht]
\centering
 \begin{tabularx}{\textwidth} {@{\hskip3pt}c@{\hskip3pt}c@{\hskip3pt}c@{\hskip3pt}c}
  \multicolumn{1}{c}{\small1000th iteration} & \multicolumn{1}{c}{\small 5000th iteration} & \multicolumn{1}{c}{\small10000th iteration} & \multicolumn{1}{c}{\small50000th iteration} 
\\
  \includegraphics[width=33mm]{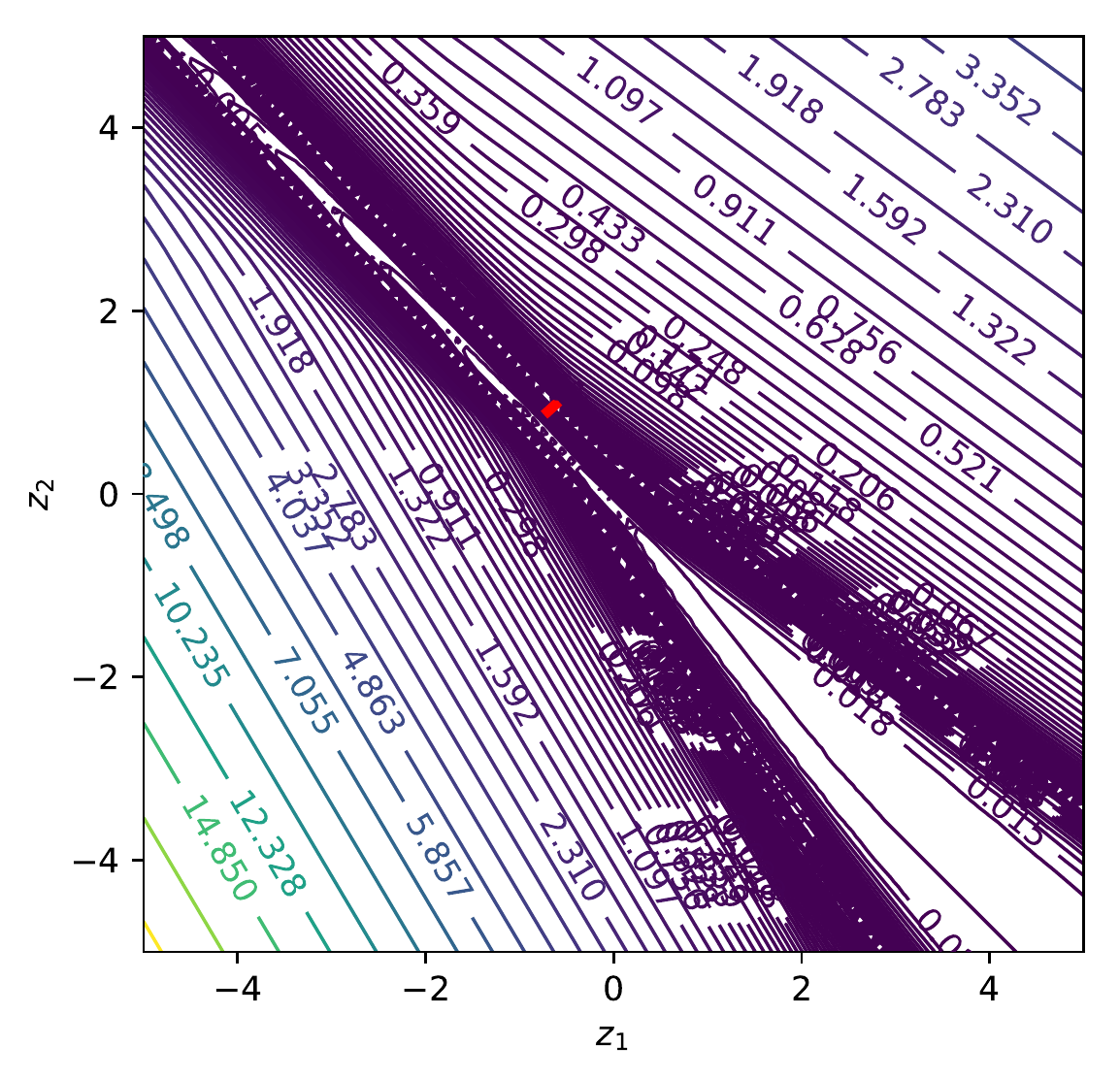}&\includegraphics[width=33mm]{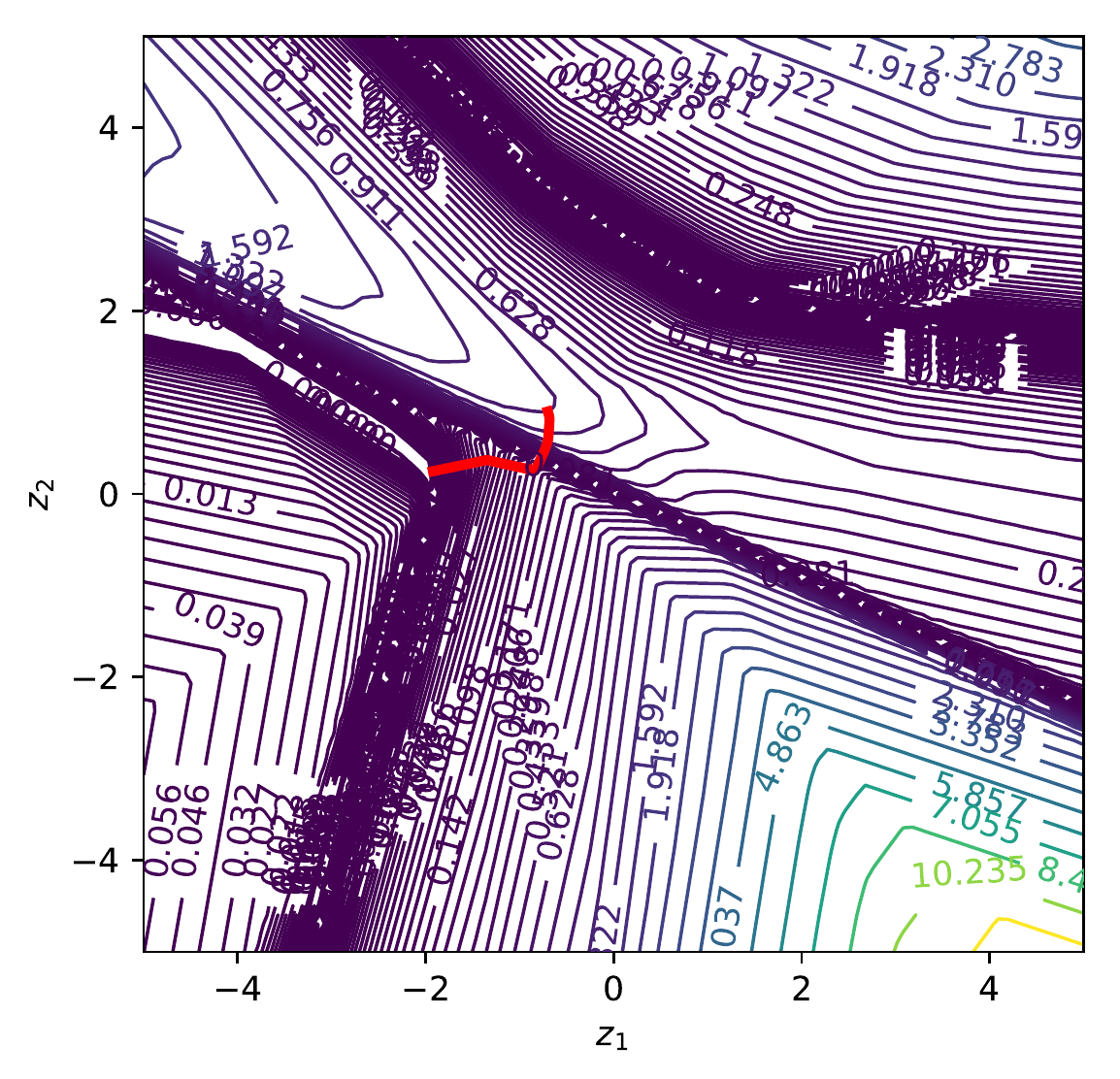}&\includegraphics[width=33mm]{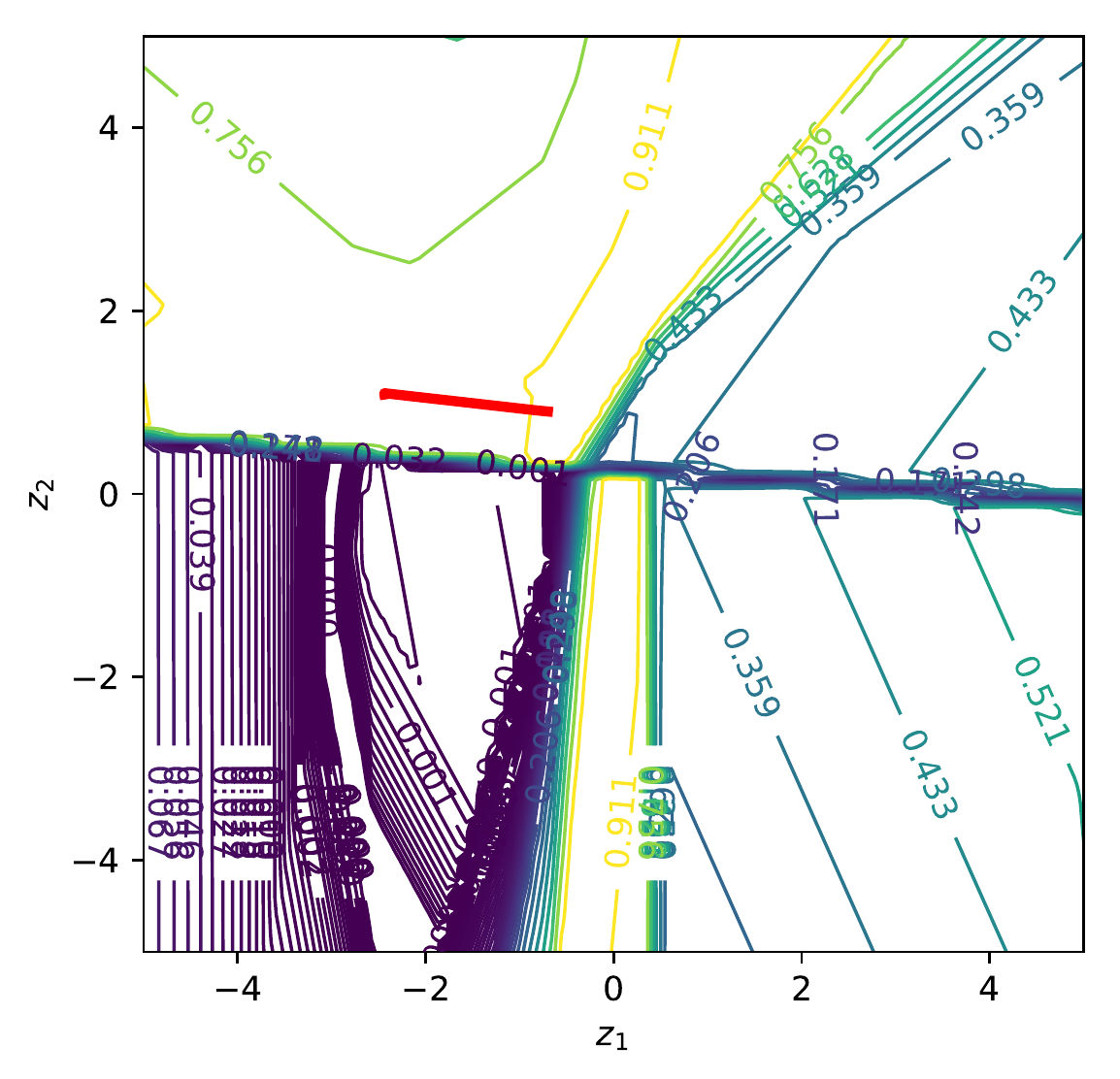}&\includegraphics[width=33mm]{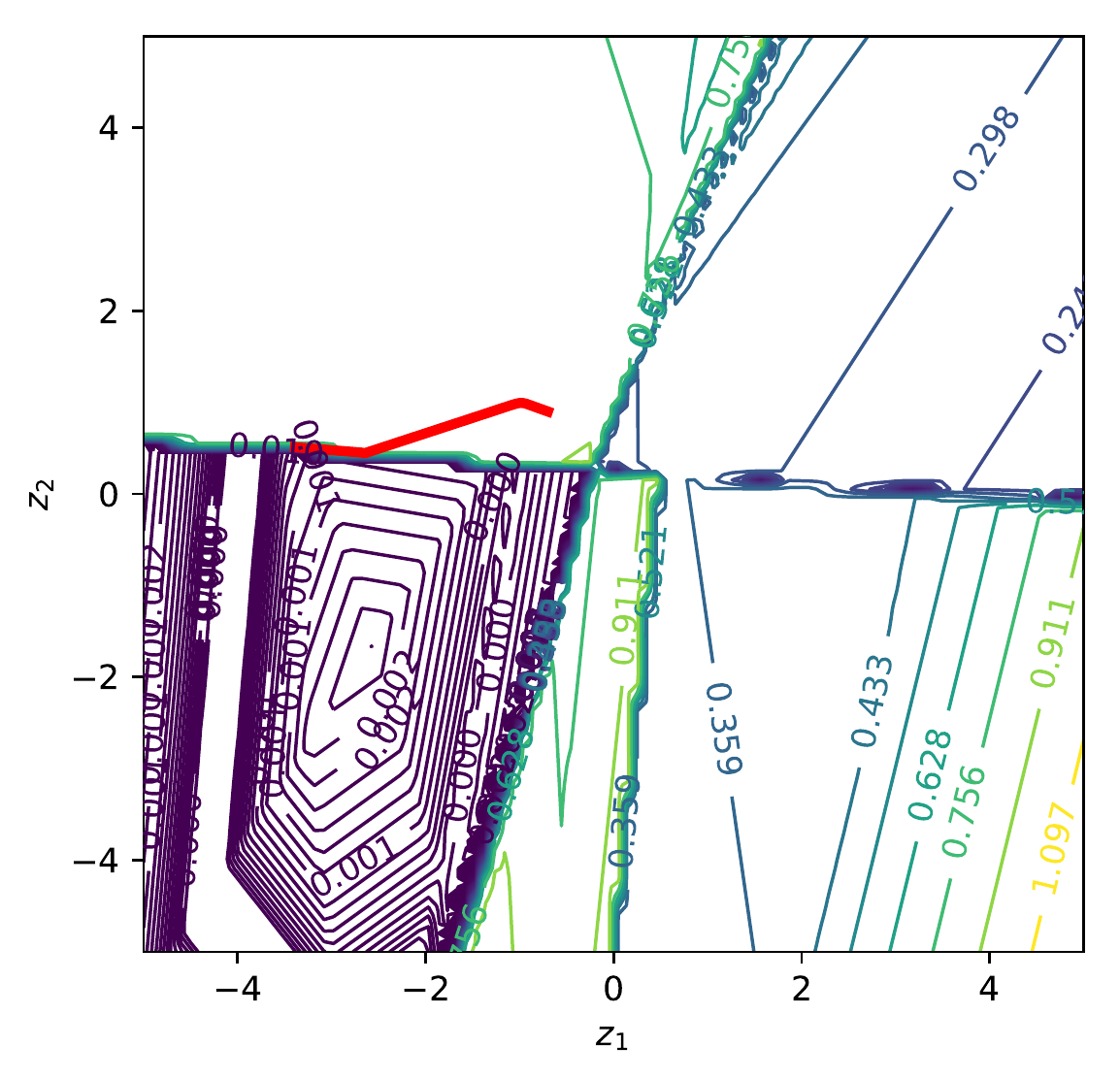} \\
   \includegraphics[width=33mm]{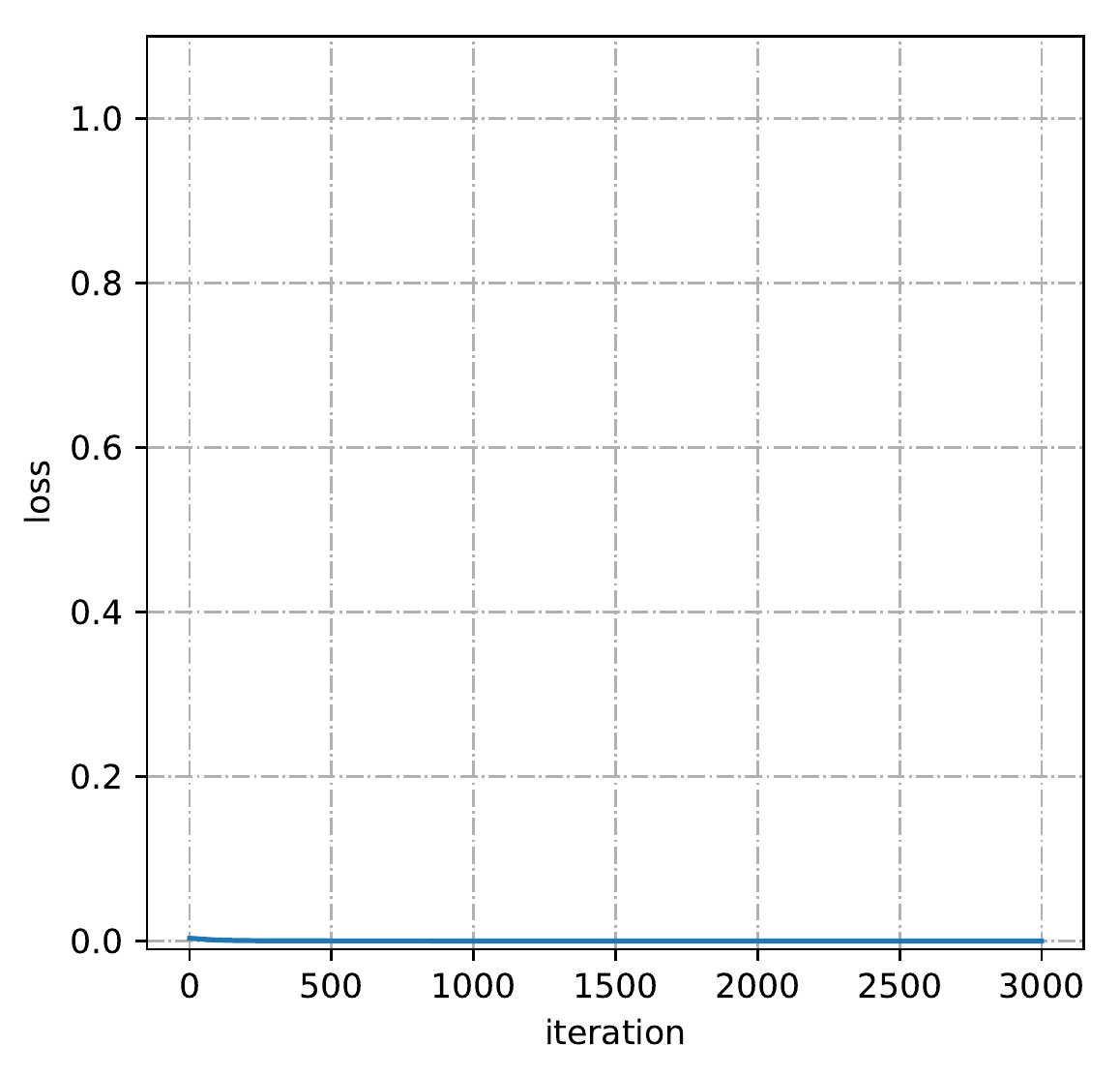}&\includegraphics[width=33mm]{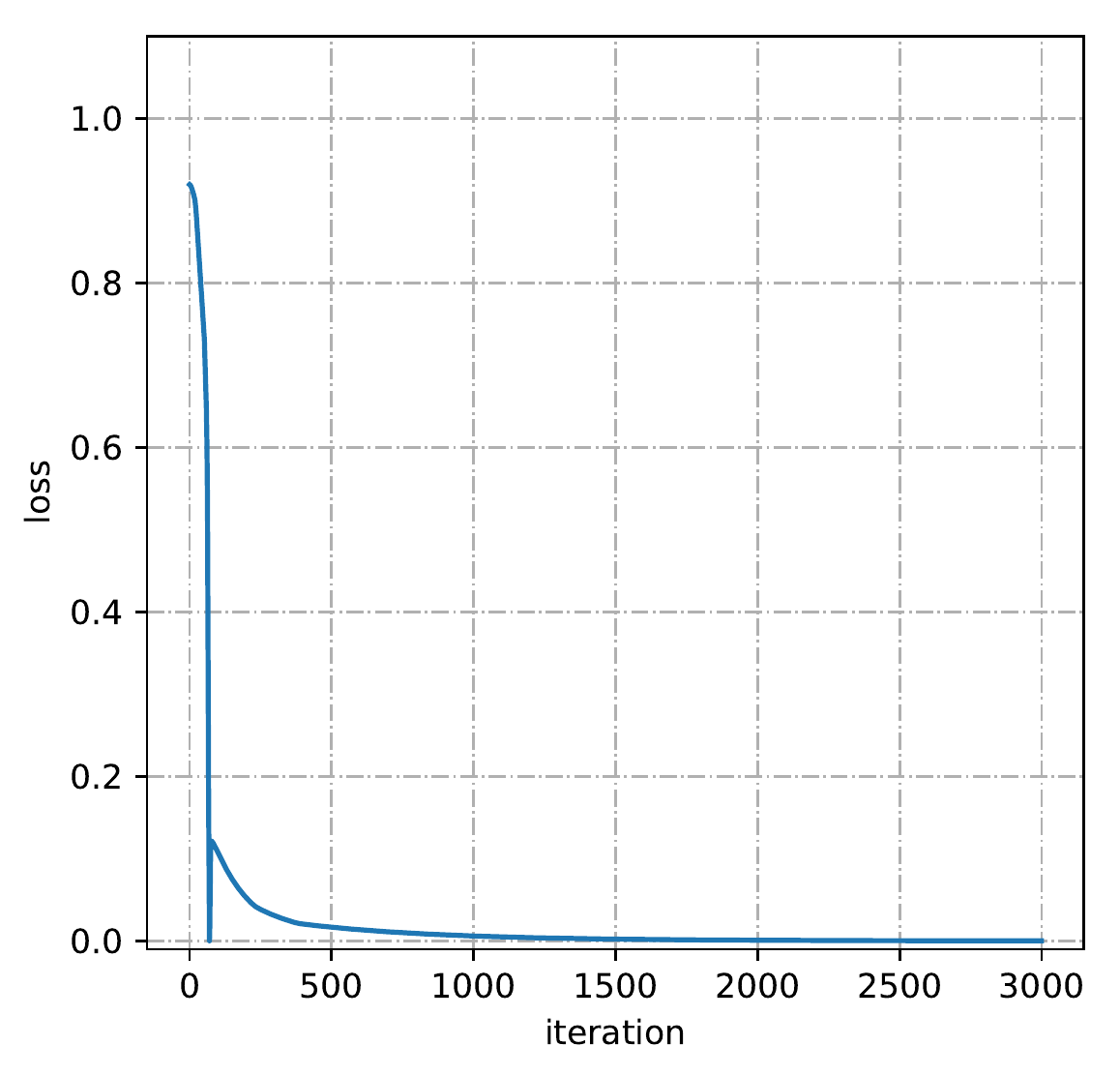}&\includegraphics[width=33mm]{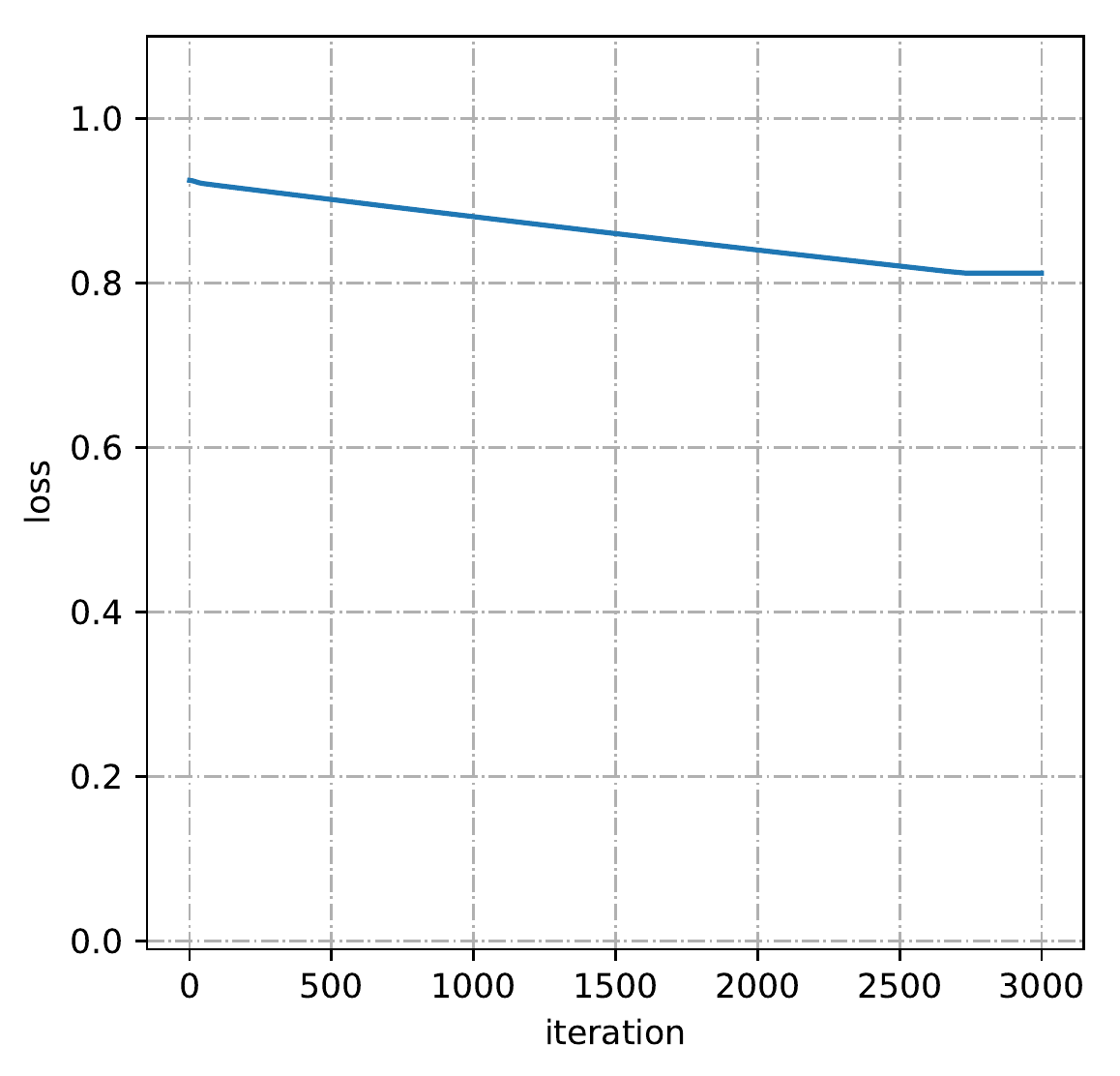}&\includegraphics[width=33mm]{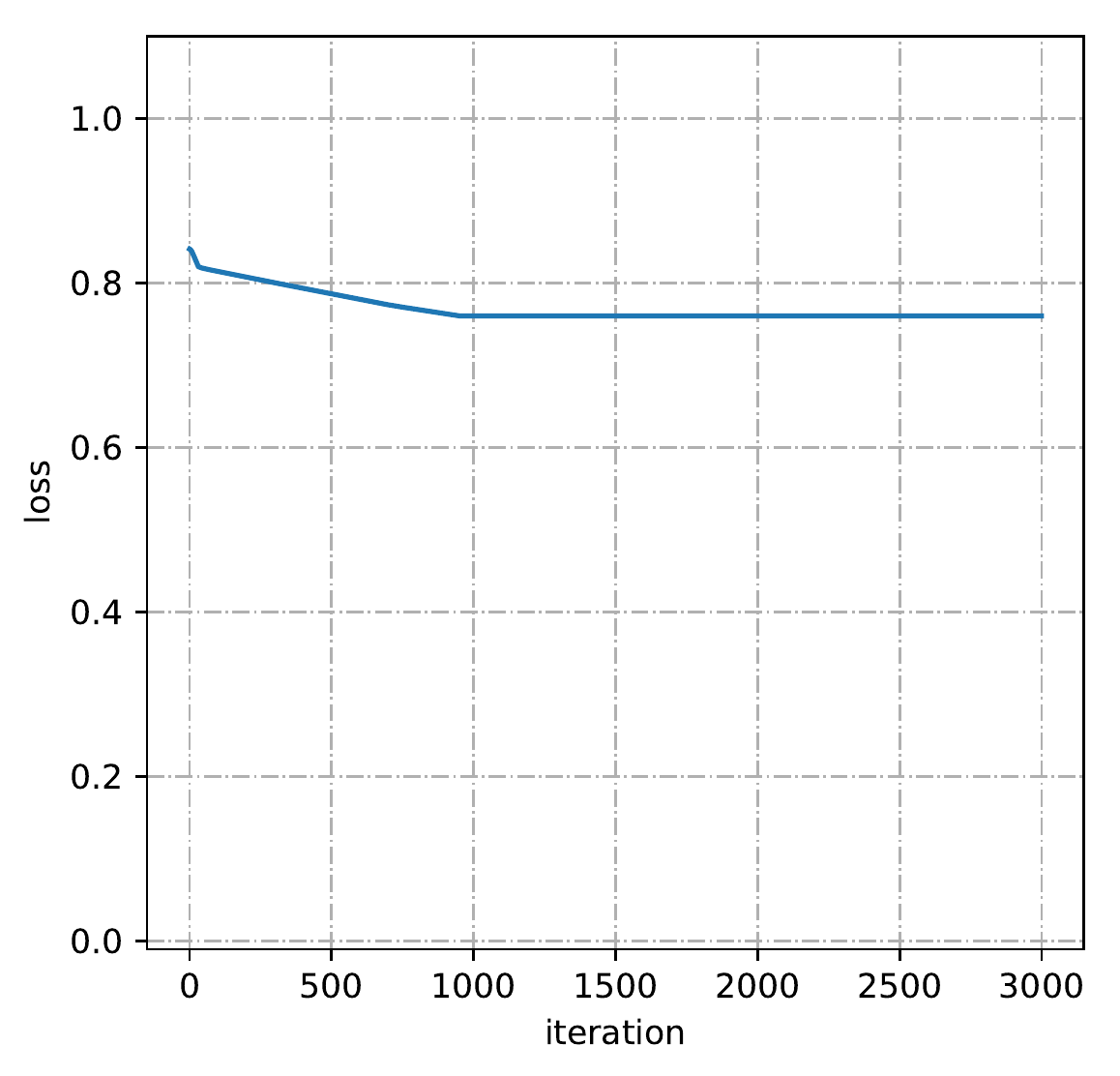} \\
\end{tabularx}
\caption{WGAN-GP's loss in $\cZ$ space and GD loss for reconstruction at 1000th, 5000th, 10000th and 50000th epoch.}
\label{fig:toysgddiff1}
\end{figure}

Sampling from a multi-modal distribution is challenging. Particularly if the modes are well separated it is important to adequately explore the domain in order not to get stuck in a single mode. To observe this  we study the ability to sample from a multi-modal distribution on our synthetic data. We use observation $x_o = x_1 = -1$ which retains an ambiguous $x_2 = 0.5$ or $x_2 = -0.5$. For this experiment, we use  the same 15k-th iteration generative model as employed in \figref{fig:toycomp}. Results for GD are shown in \figref{fig:ambigres}~(b,c) while AIS results are provided in \figref{fig:ambigres}~(d, e). 
We observe GD gets stuck in bad modes. In contrast, AIS  finds the optimal modes.

Following Fig. 1, Fig. 2 and Fig. 3 in the main paper, we compare the performances of baselines and our proposed AIS based HMC algorithm on WGAN-GP. The results illustrate the robustness of our algorithm on a variety of GAN losses and advantages over the baselines. 

\begin{figure}[t]
\centering
 \begin{tabularx}{\textwidth} {c@{\hskip1pt}c@{\hskip3pt}c@{\hskip3pt}c@{\hskip3pt}c@{\hskip3pt}c}

  \rotatebox{90}{\hspace{0.8cm}{\small 1000}} & \includegraphics[width=25mm]{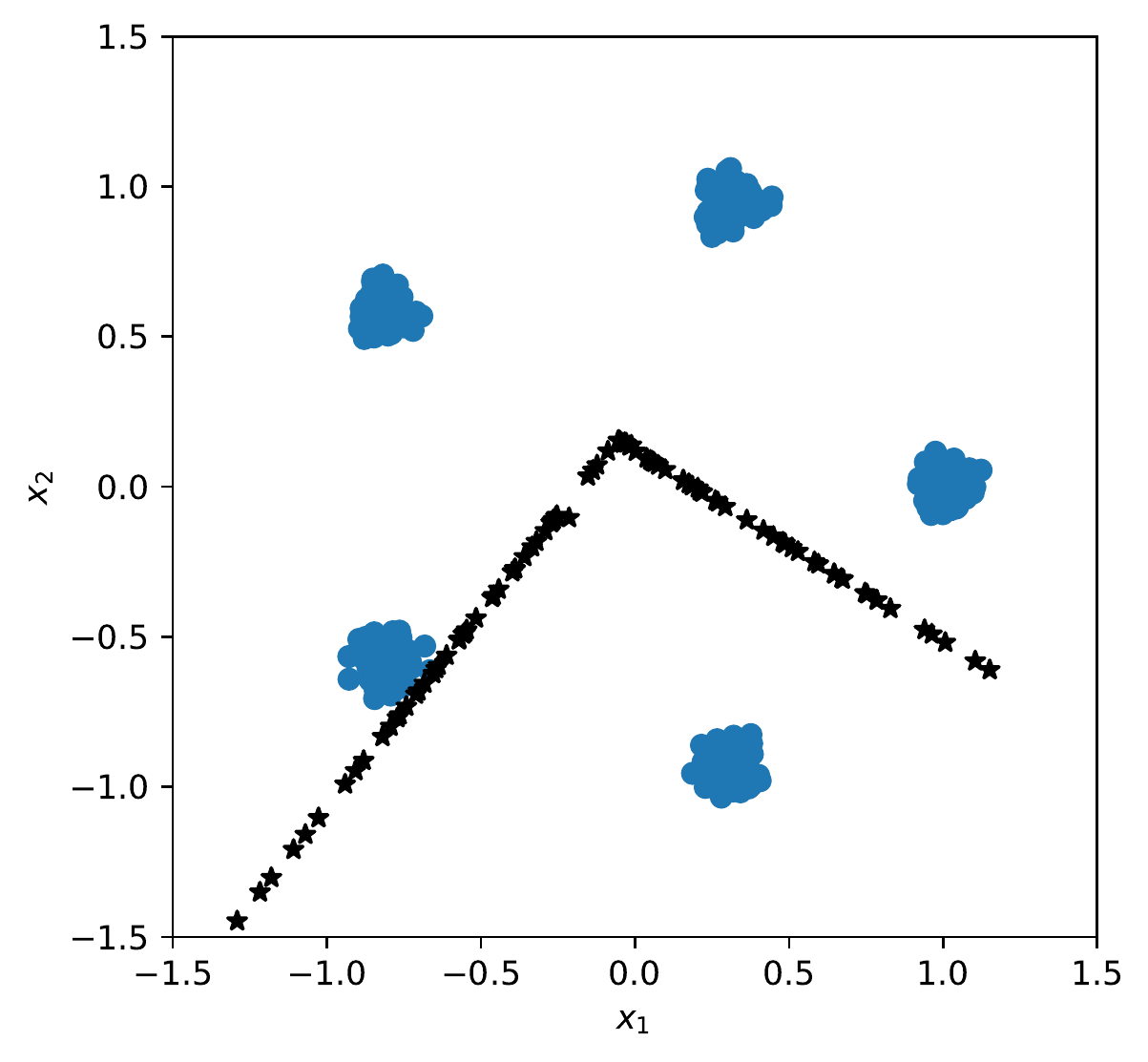}&\includegraphics[width=25mm]{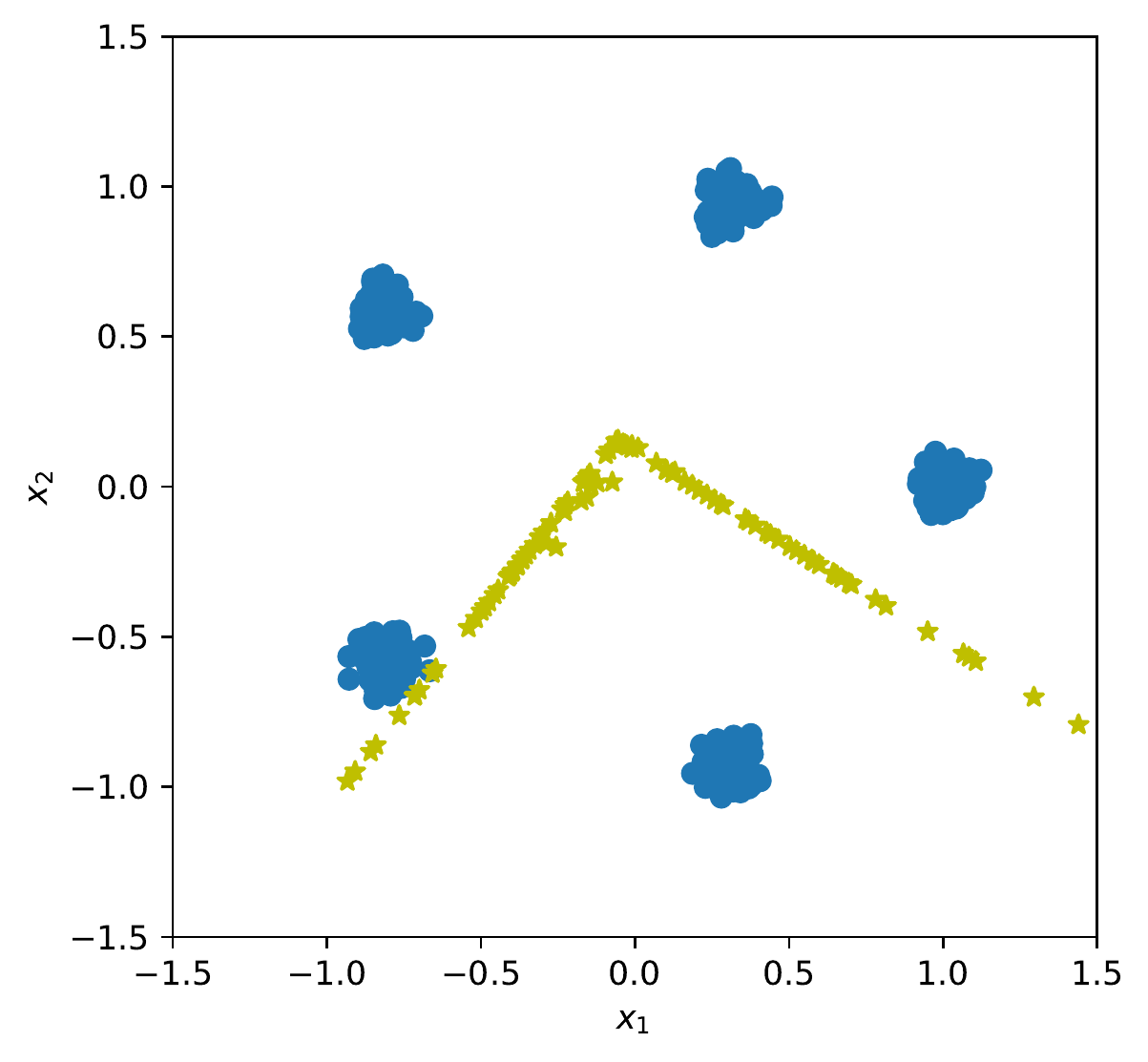}&\includegraphics[width=25mm]{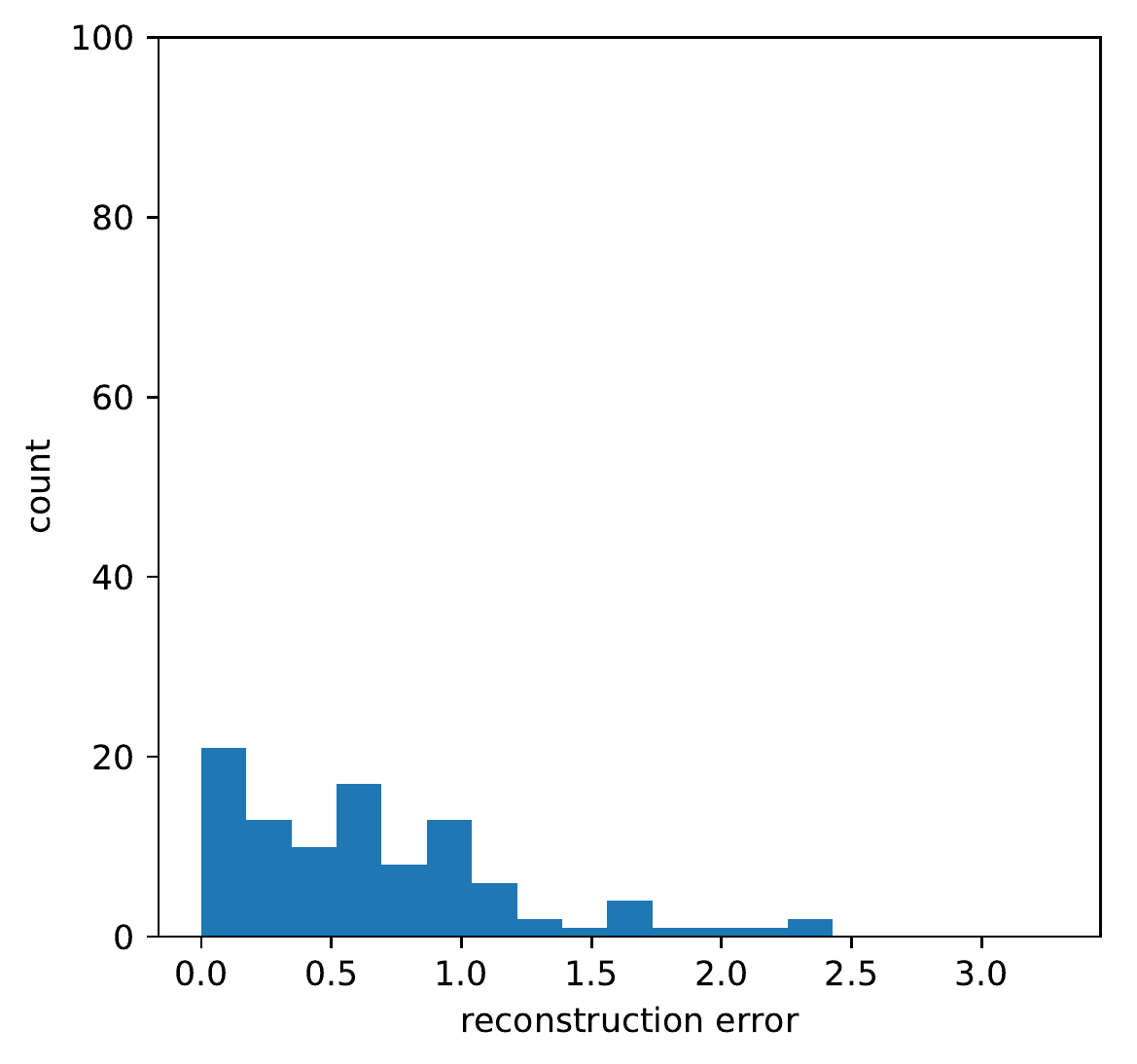}&\includegraphics[width=25mm]{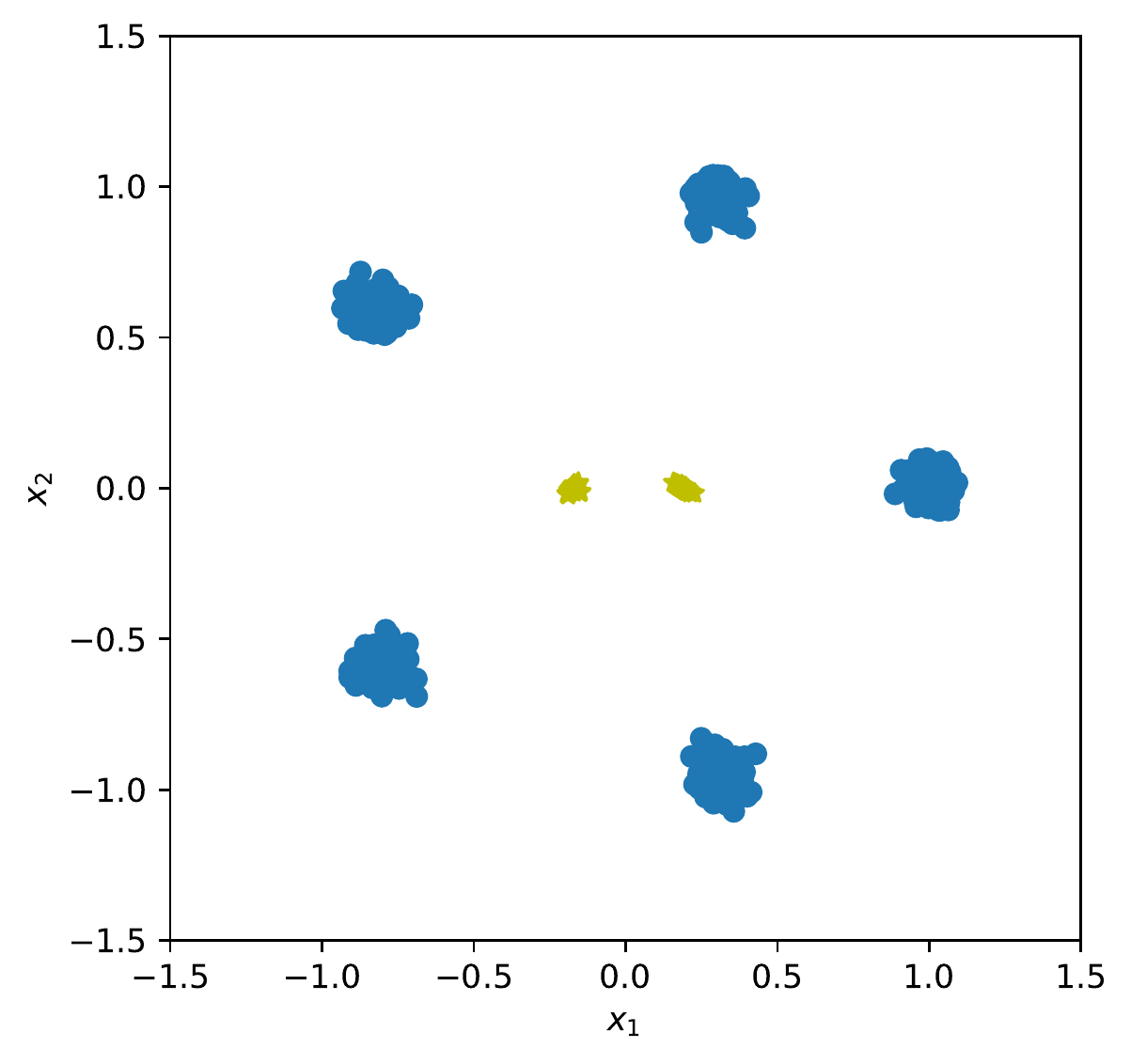}&\includegraphics[width=25mm]{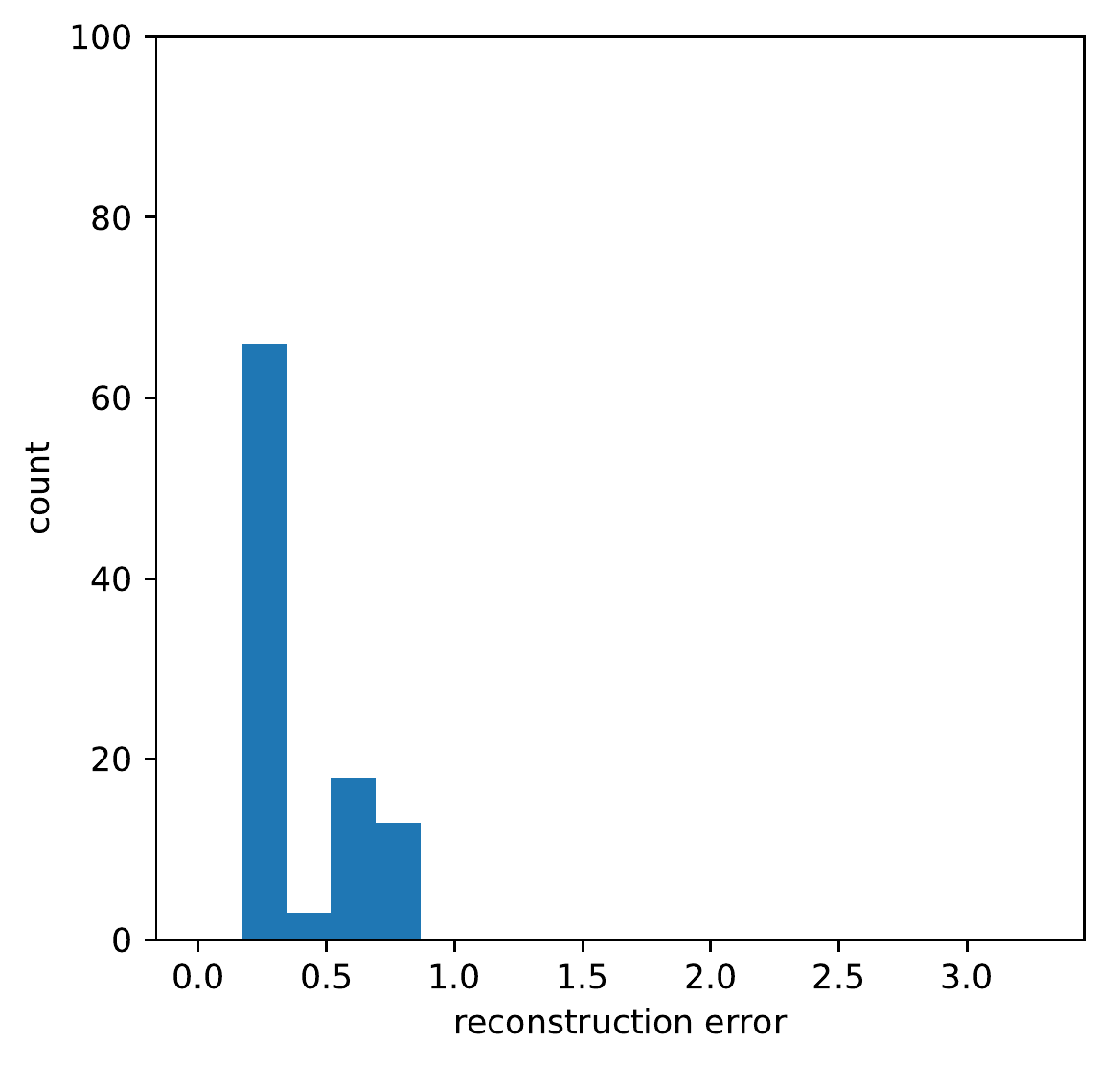}\\
   \rotatebox{90}{\hspace{0.8cm}{\small 5000}} & \includegraphics[width=25mm]{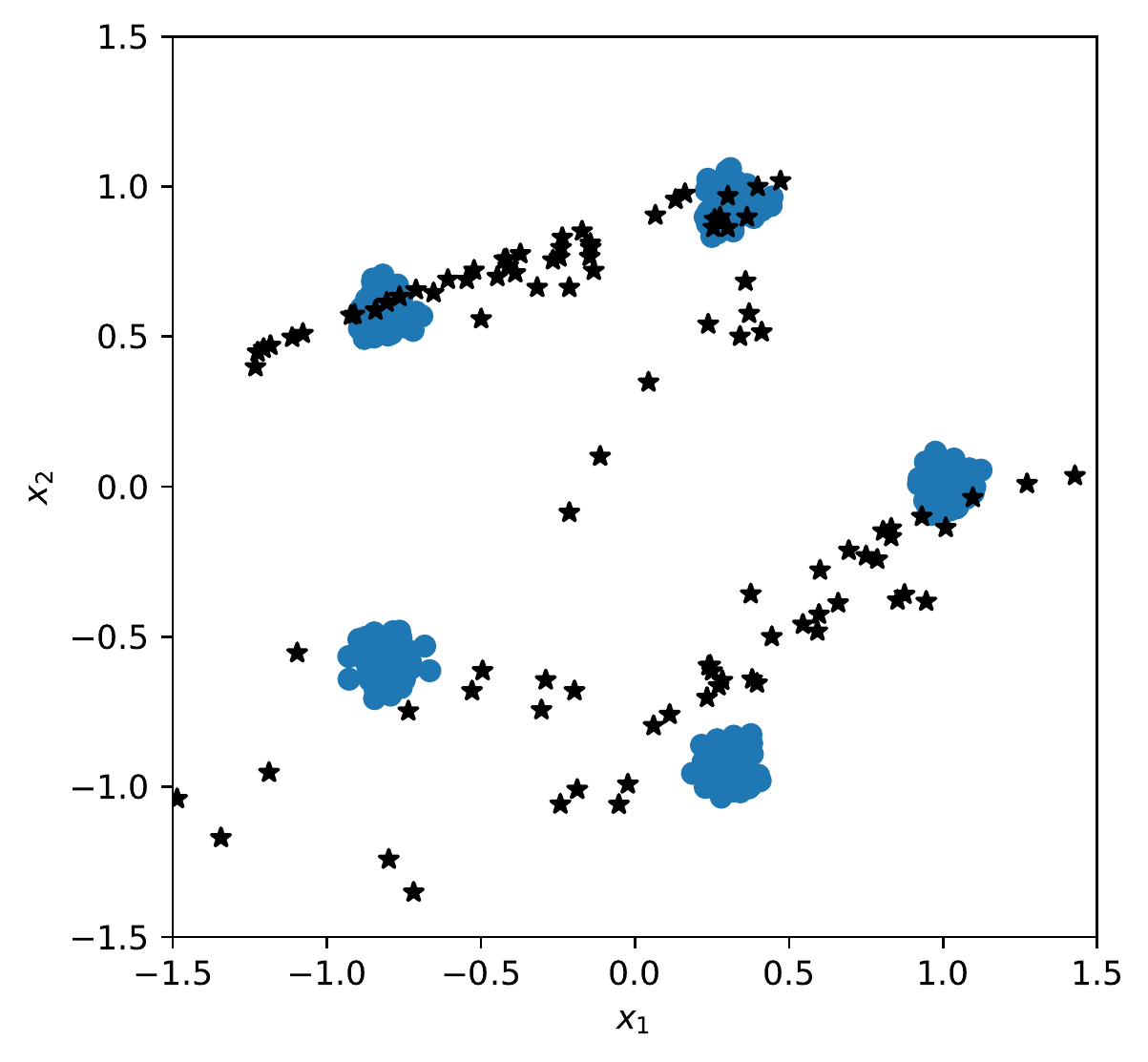}&\includegraphics[width=25mm]{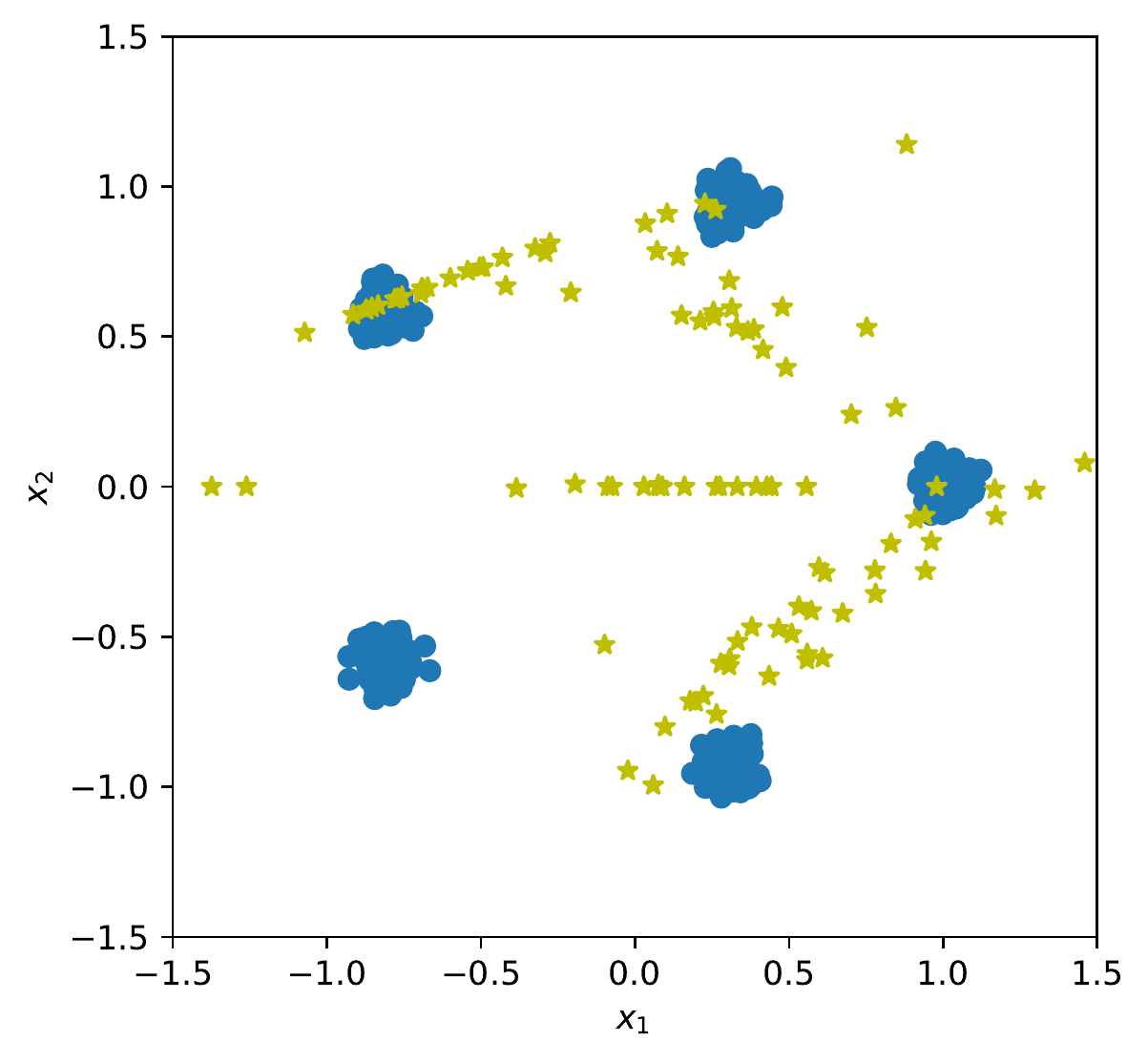}&\includegraphics[width=25mm]{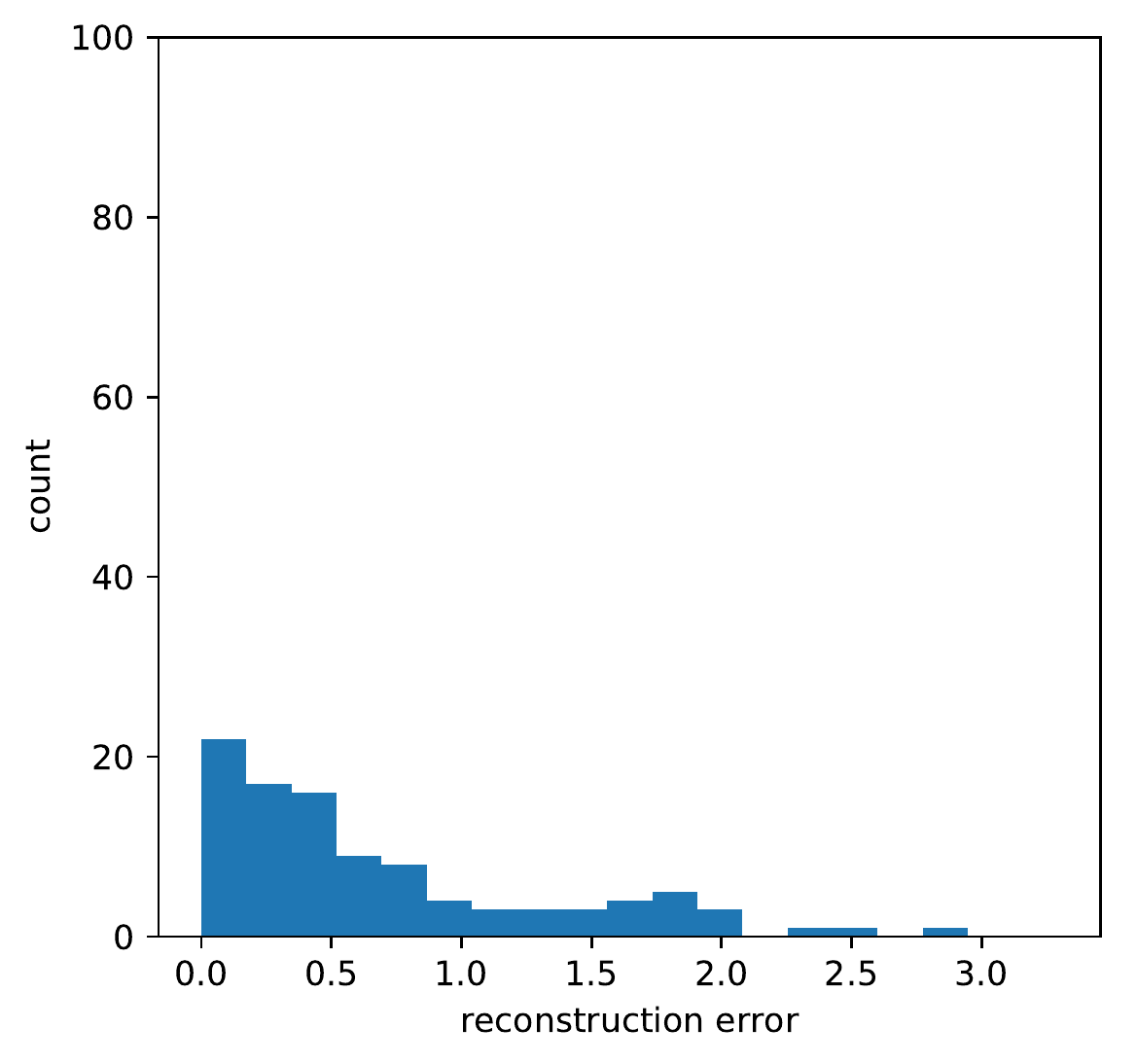}&\includegraphics[width=25mm]{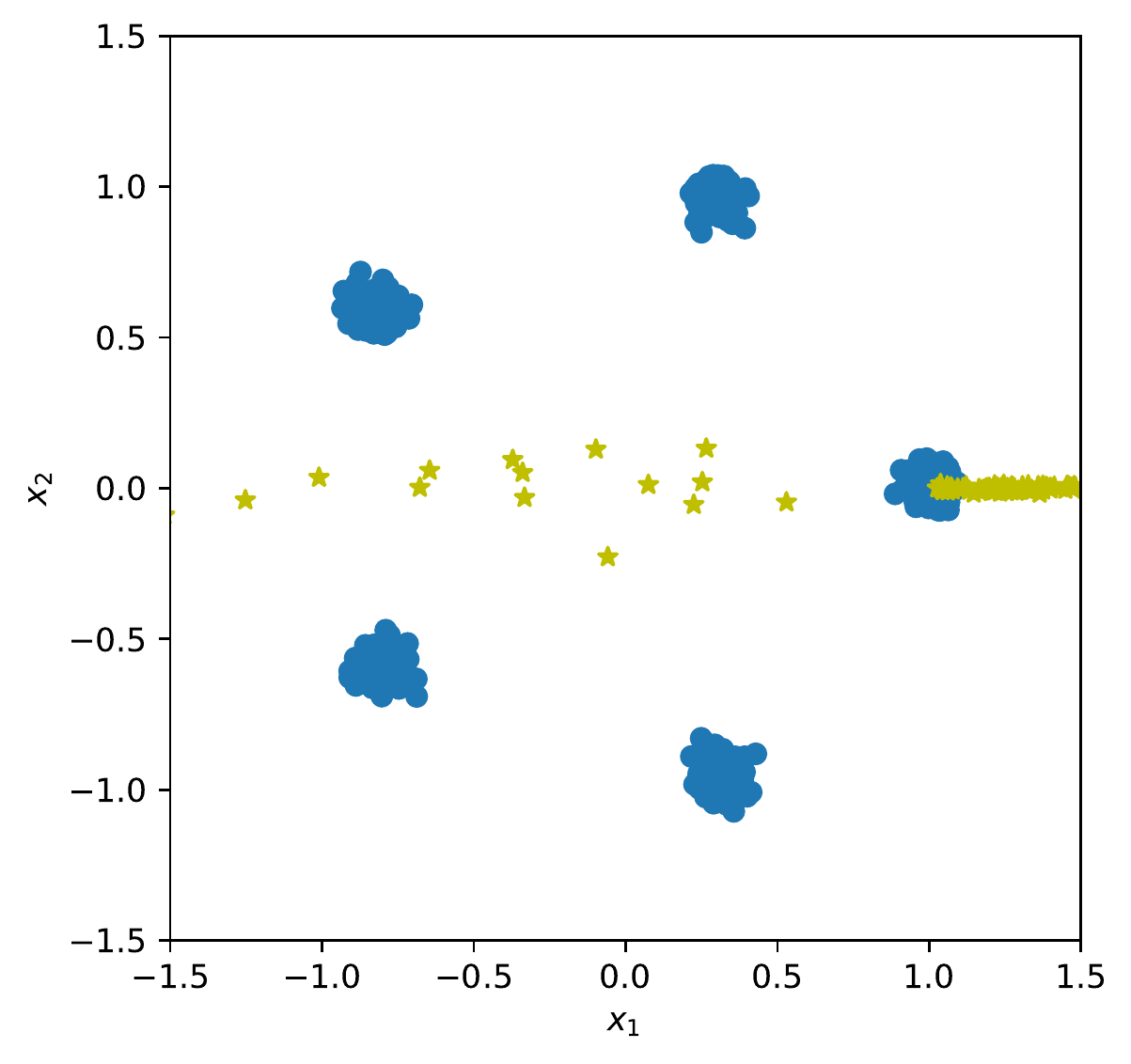}&\includegraphics[width=25mm]{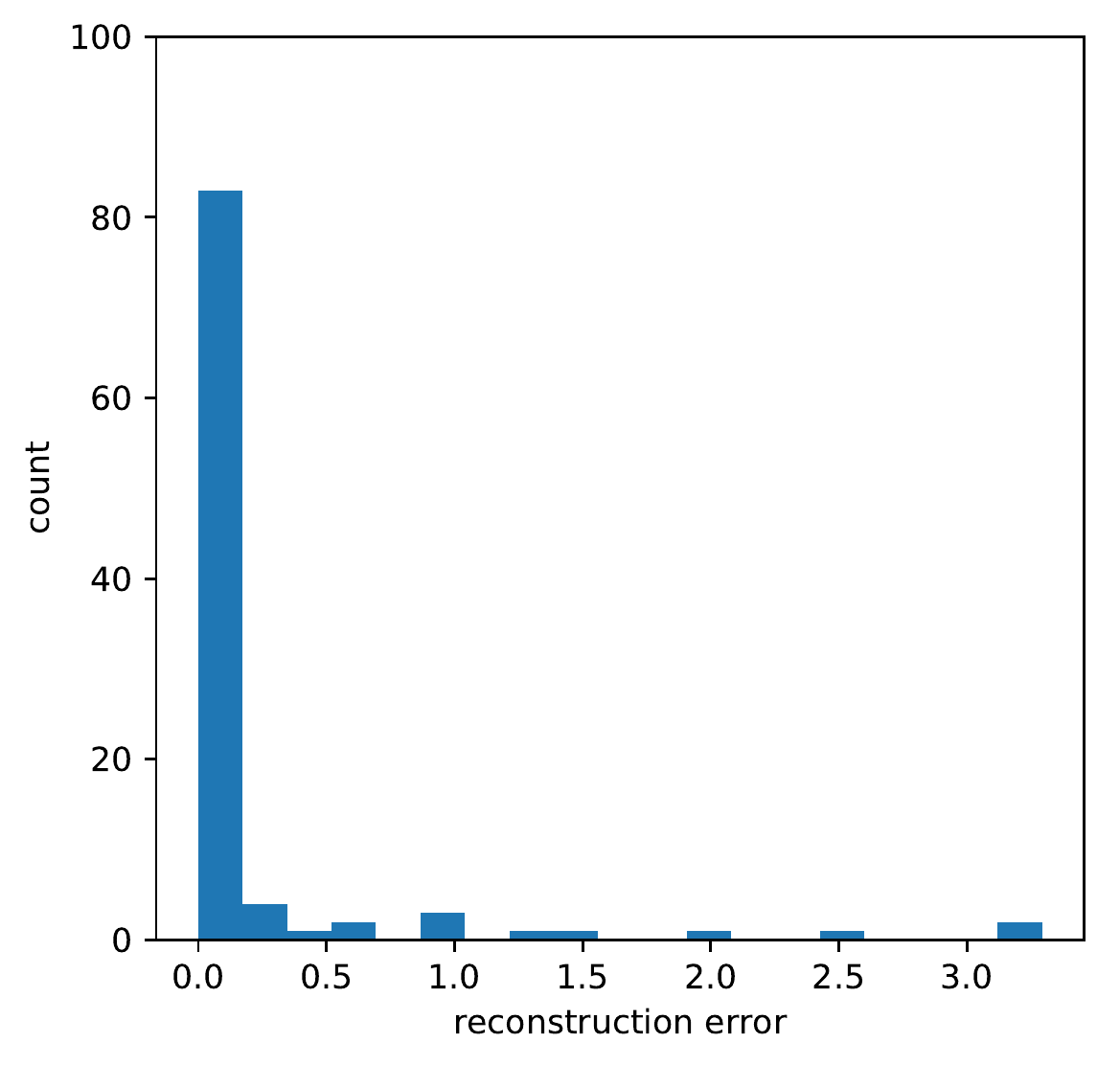}\\
   \rotatebox{90}{\hspace{0.8cm}{\small 10000}} & \includegraphics[width=25mm]{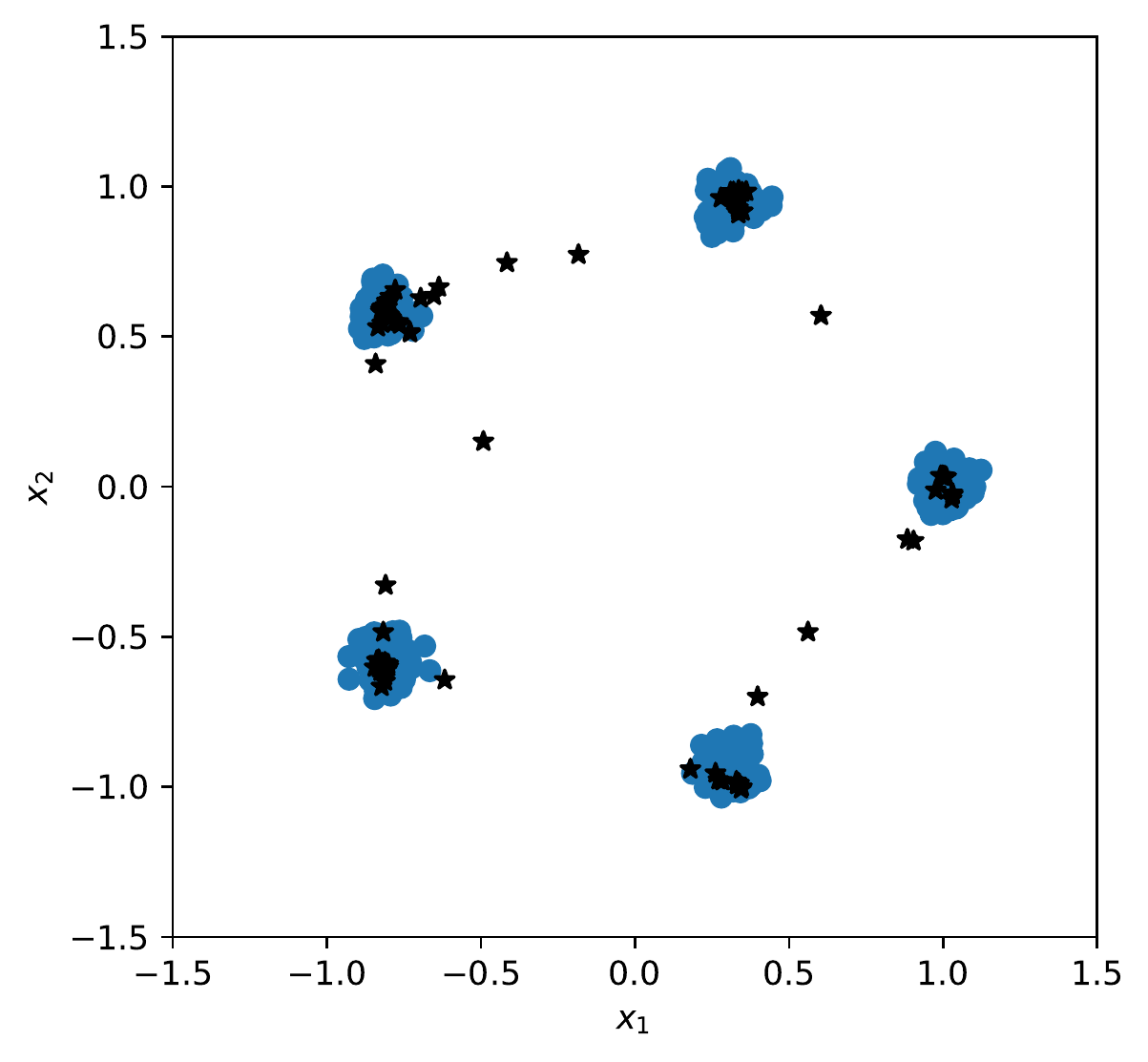}&\includegraphics[width=25mm]{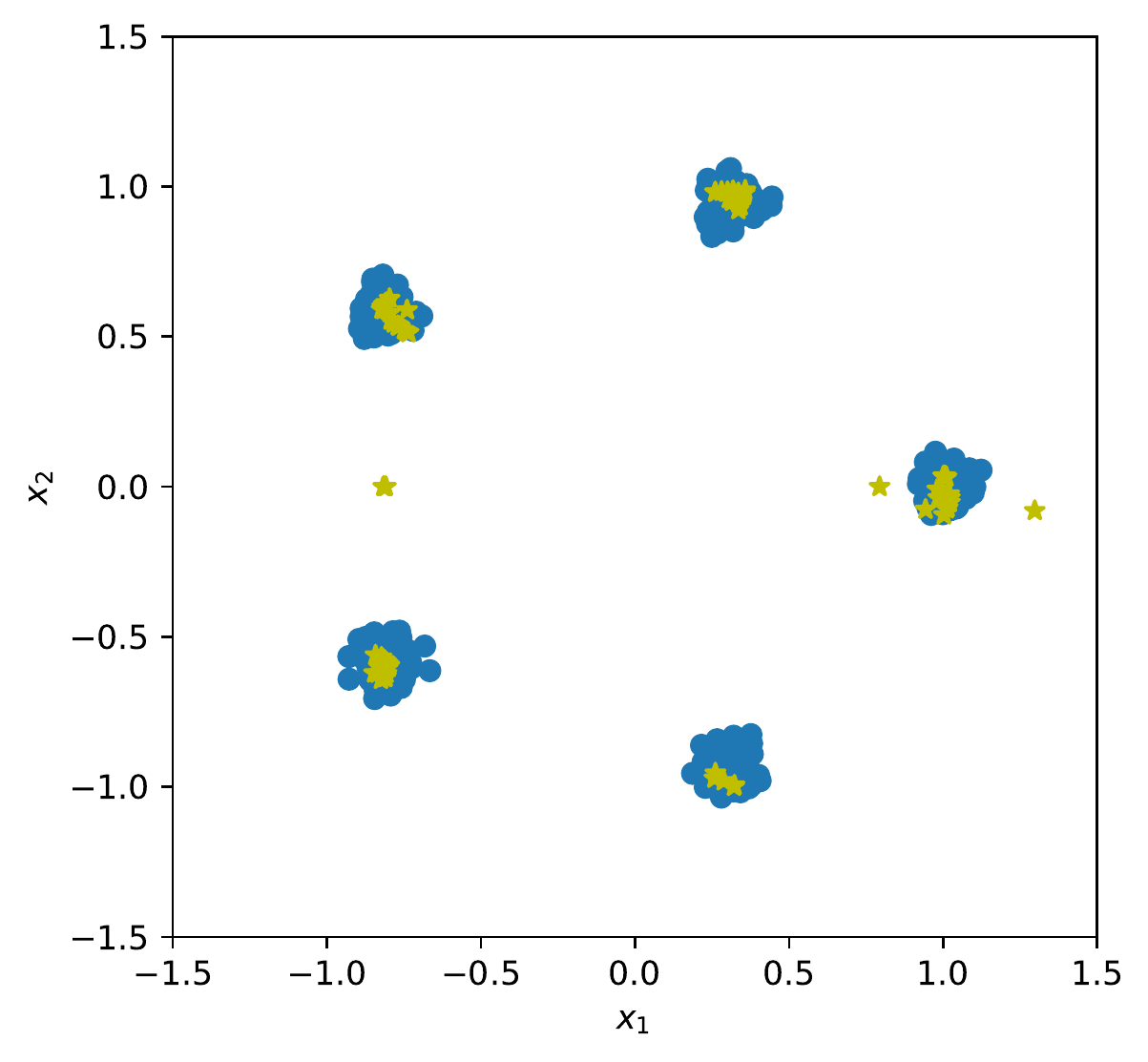}&\includegraphics[width=25mm]{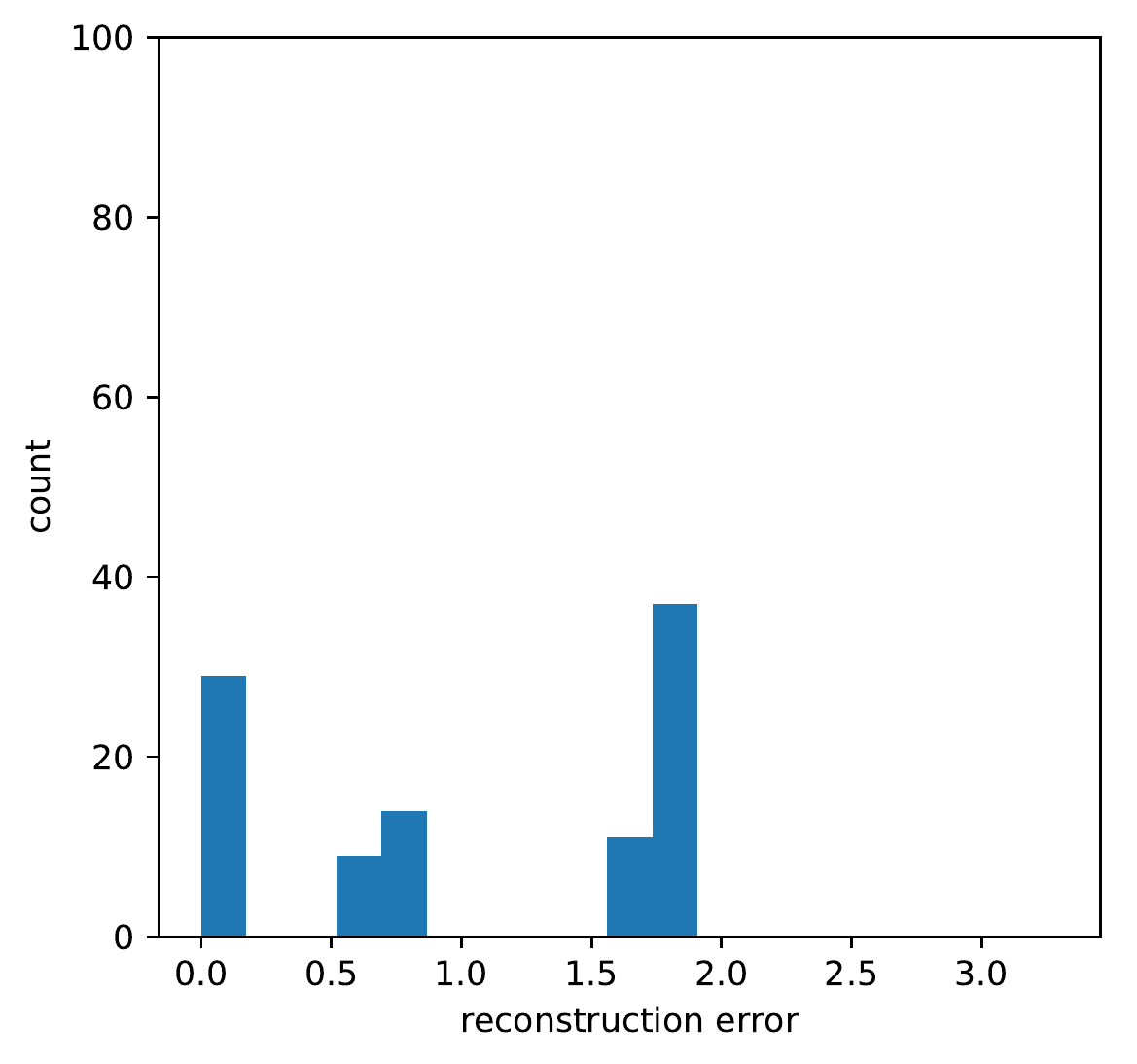}&\includegraphics[width=25mm]{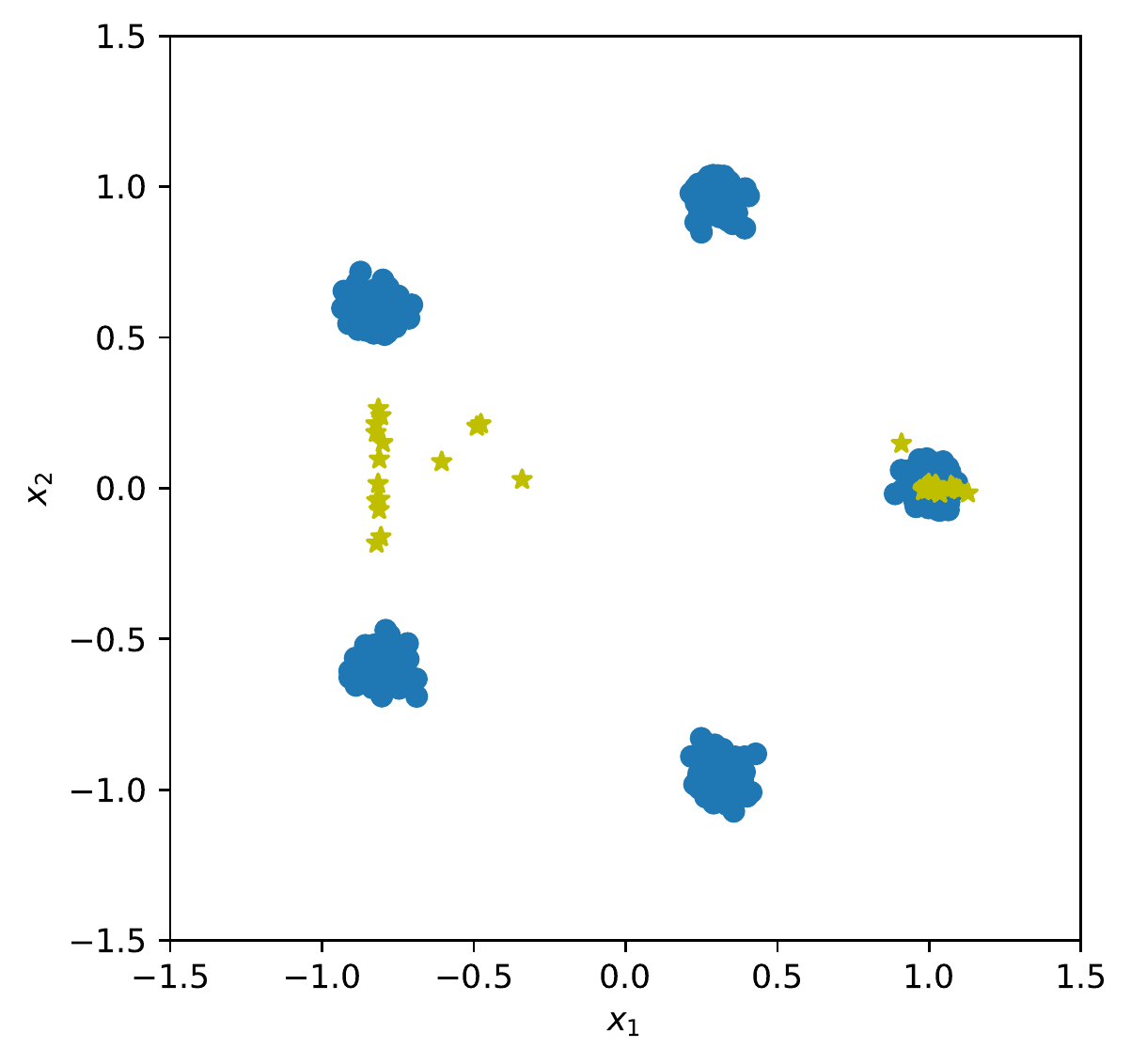}&\includegraphics[width=25mm]{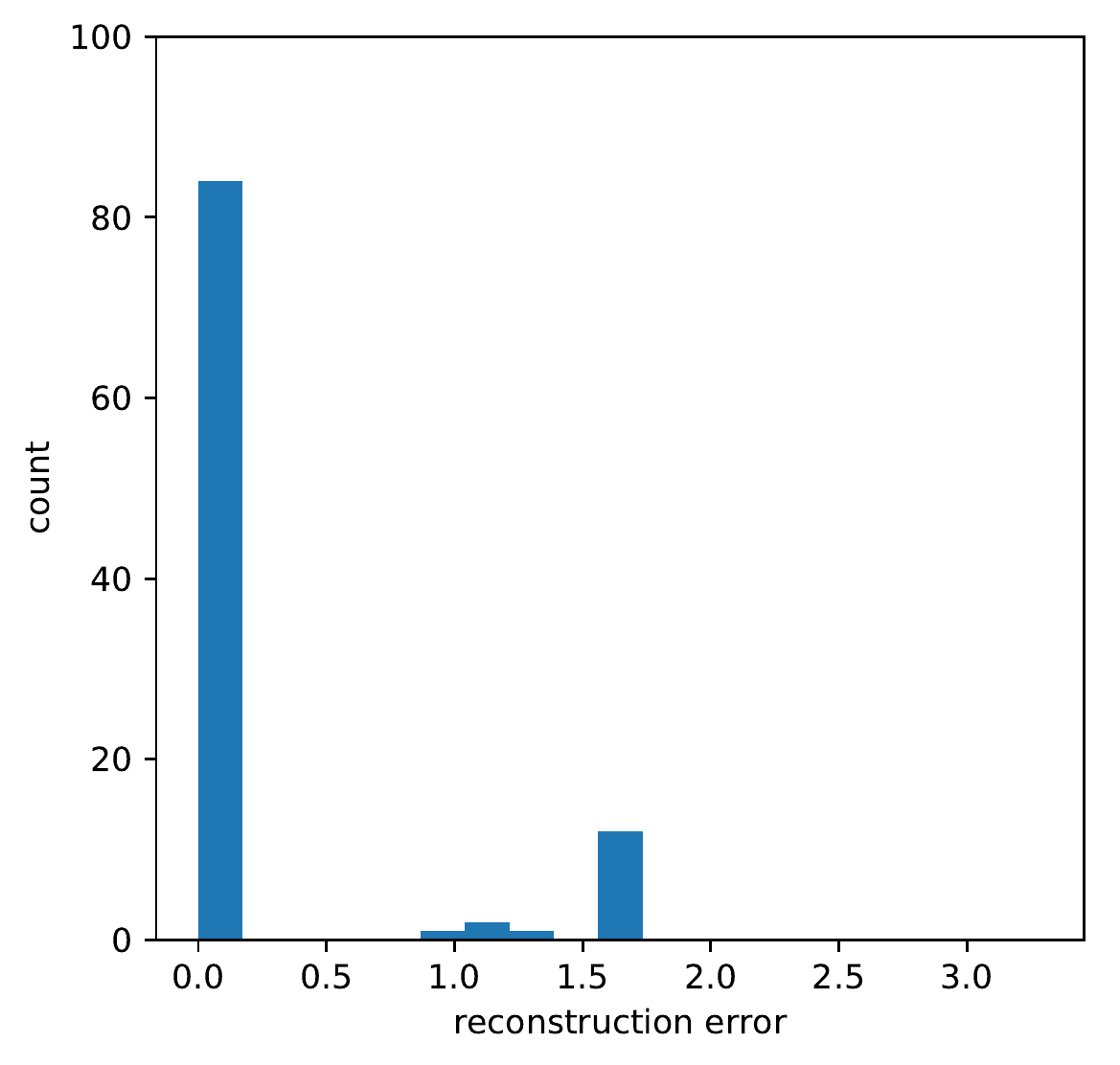}\\
   \rotatebox{90}{\hspace{0.8cm}{\small 50000}} & \includegraphics[width=25mm]{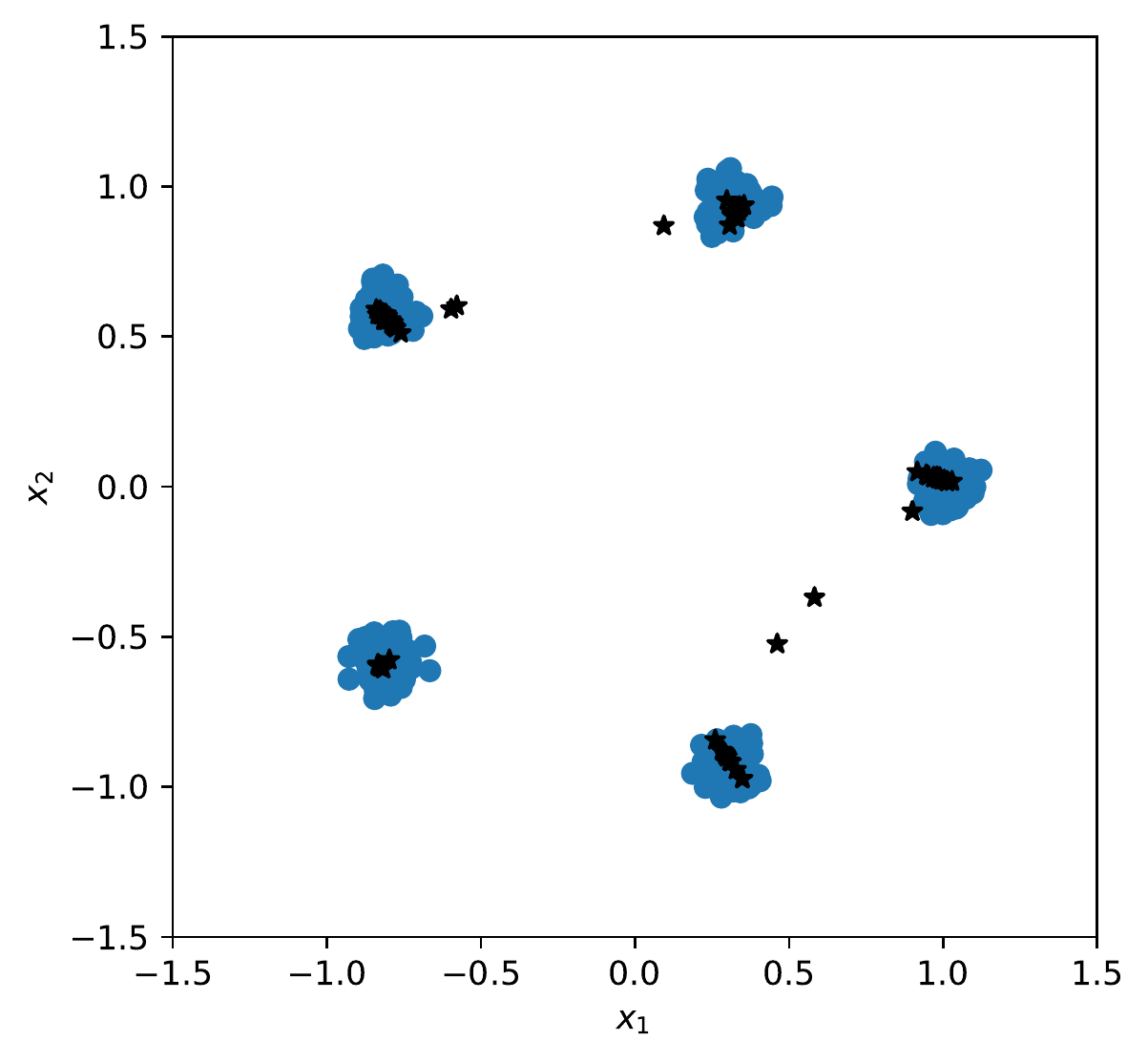}&\includegraphics[width=25mm]{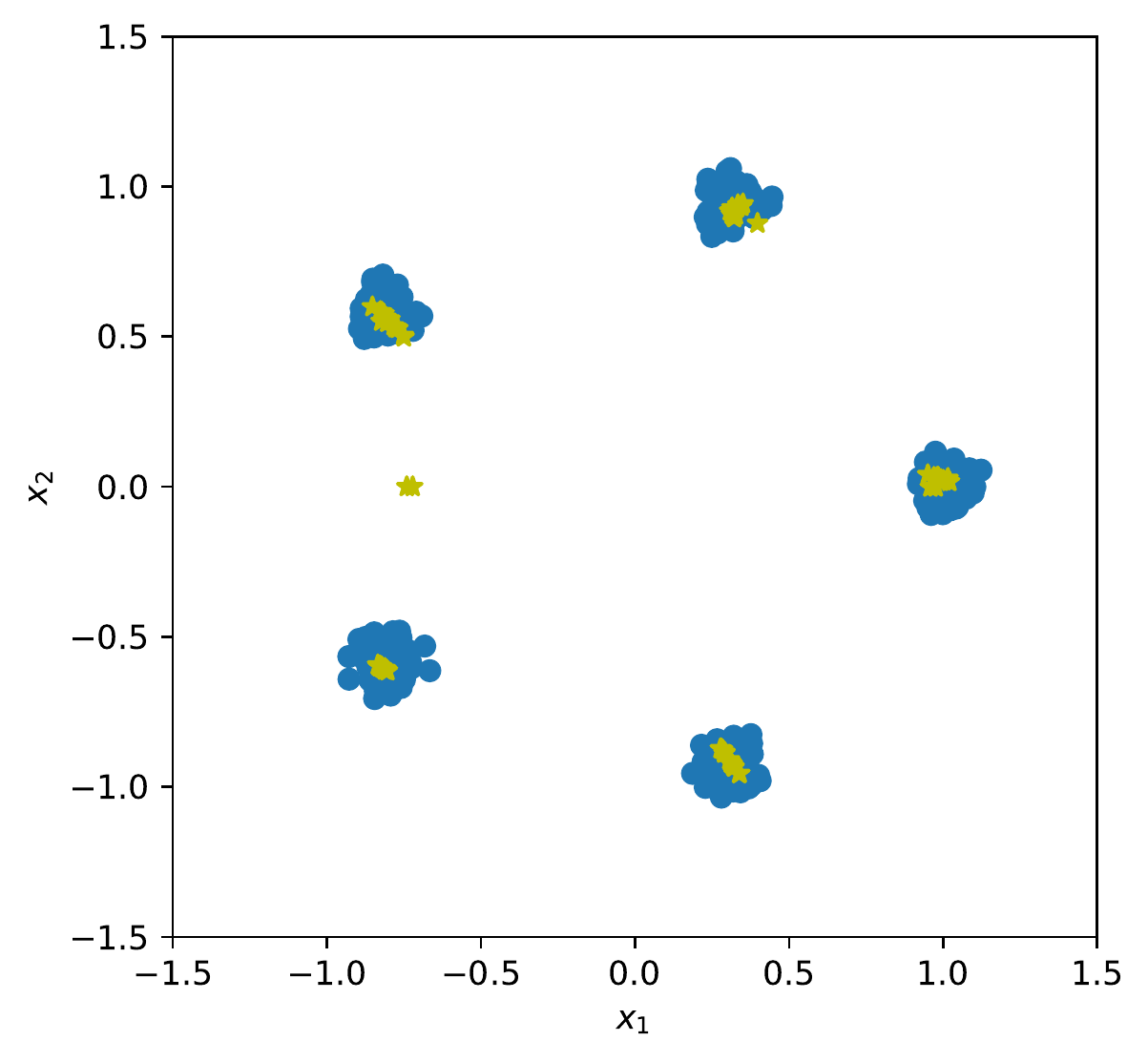}&\includegraphics[width=25mm]{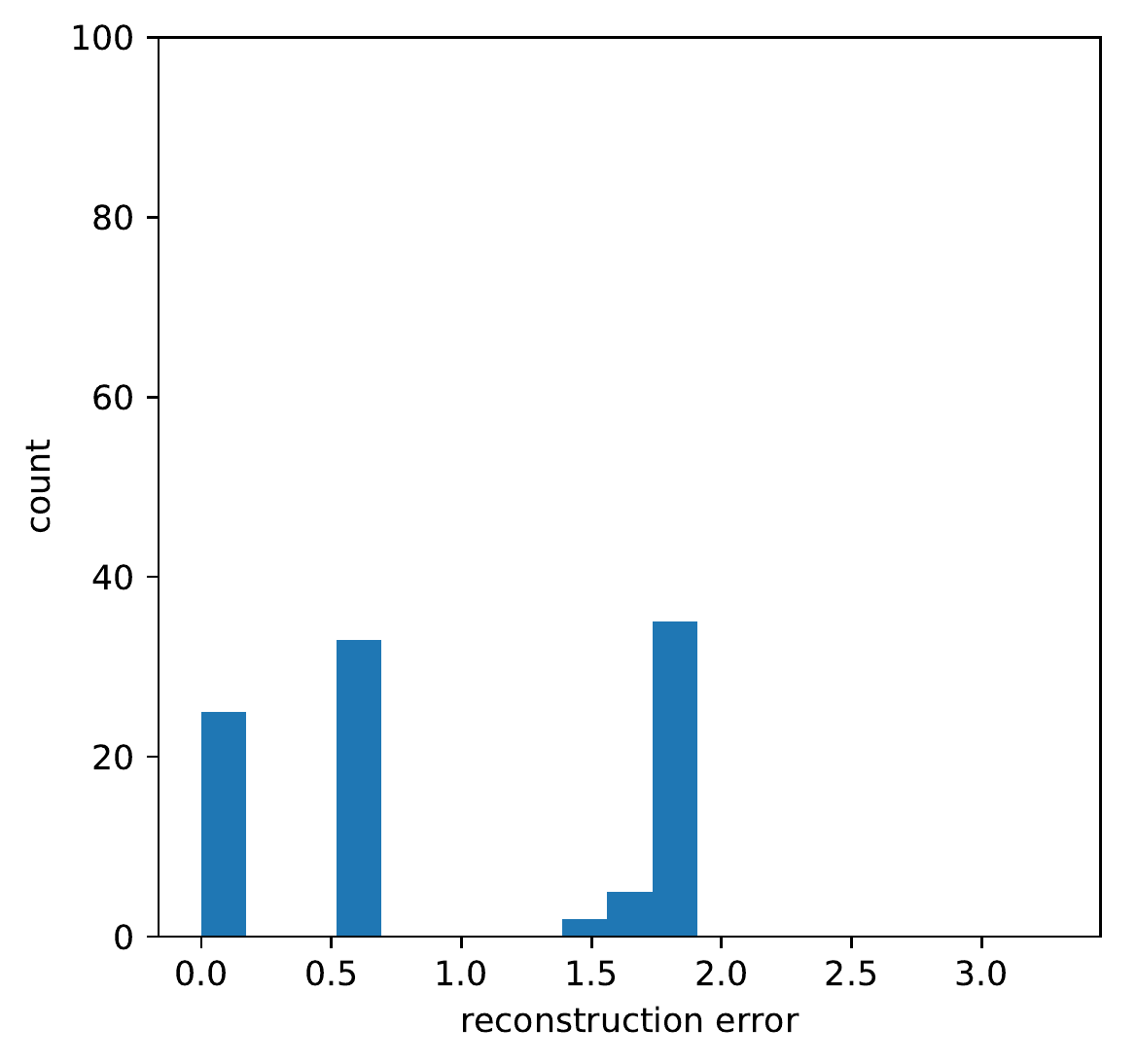}&\includegraphics[width=25mm]{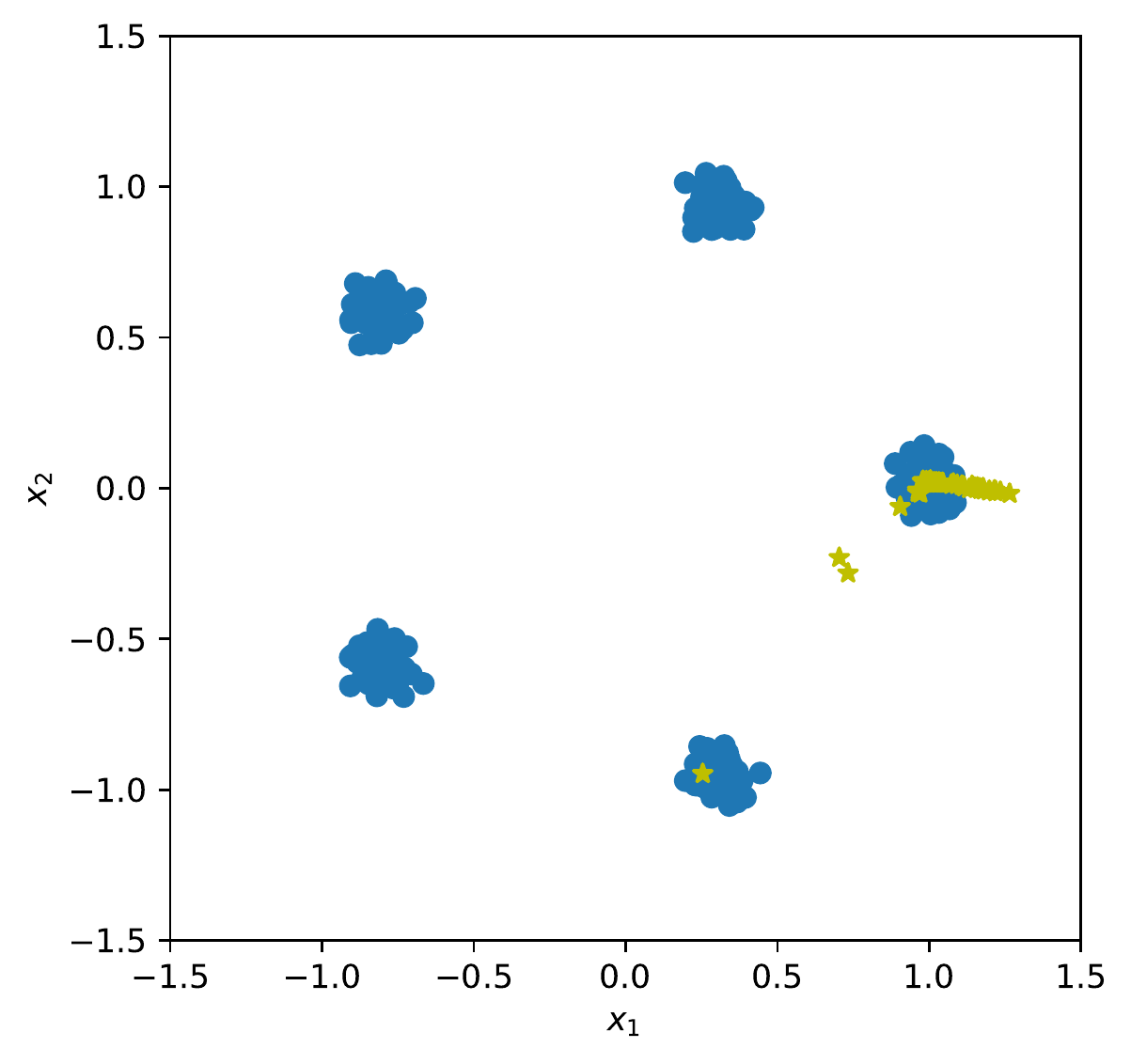}&\includegraphics[width=25mm]{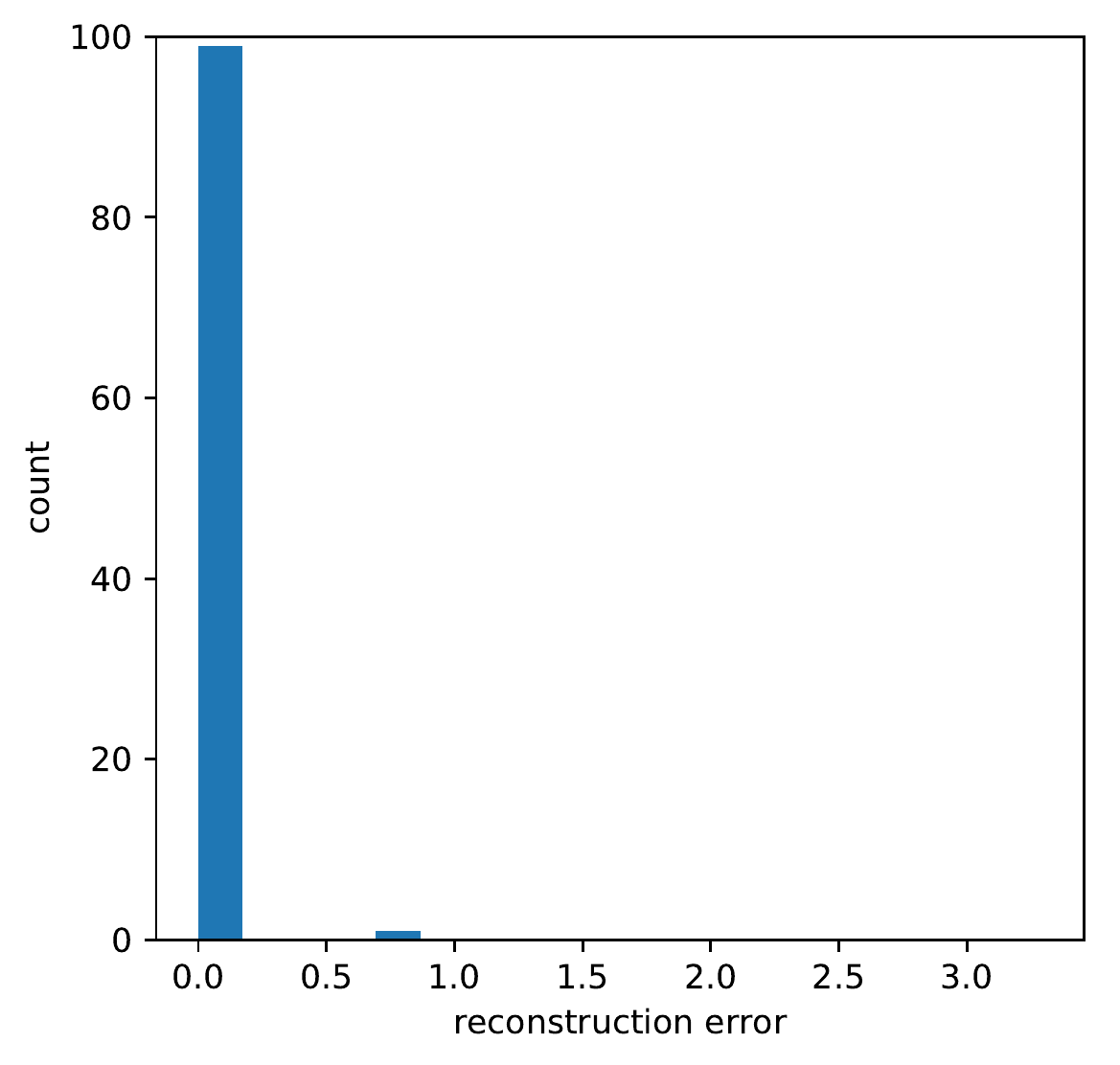}\\
  &\multicolumn{1}{c}{{\small (a)}} & \multicolumn{1}{c}{{\small (b)}} & \multicolumn{1}{c}{{\small (c)}} & \multicolumn{1}{c}{{\small (d)}} & \multicolumn{1}{c}{{\small (e)}}
   
\end{tabularx}
\caption{Rows correspond to generators trained for a different number of epochs as indicated (left). The columns illustrate: (a) Samples generated with a WGAN-GP (black); (b) GD reconstructions from 100 random initializations; (c)  Reconstruction error bar plot for the result in column (b); (d) Reconstructions recovered with our proposed AIS based HMC algorithm; (e) Reconstruction error bar plot for the results in column (d). 
}
\label{fig:toycomp1}
\end{figure}

\begin{figure}[t]
\centering
 \begin{tabularx}{\textwidth} {@{\hskip3pt}c@{\hskip3pt}c@{\hskip3pt}c@{\hskip3pt}c@{\hskip3pt}c}
  \multicolumn{1}{c}{$t = 100$} & \multicolumn{1}{c}{$t = 3700$} & \multicolumn{1}{c}{$t = 3800$} & \multicolumn{1}{c}{$t = 3900$} & \multicolumn{1}{c}{$t = 4000$} \\
  \includegraphics[width=27mm]{figs/fig3_iter100.pdf}&\includegraphics[width=27mm]{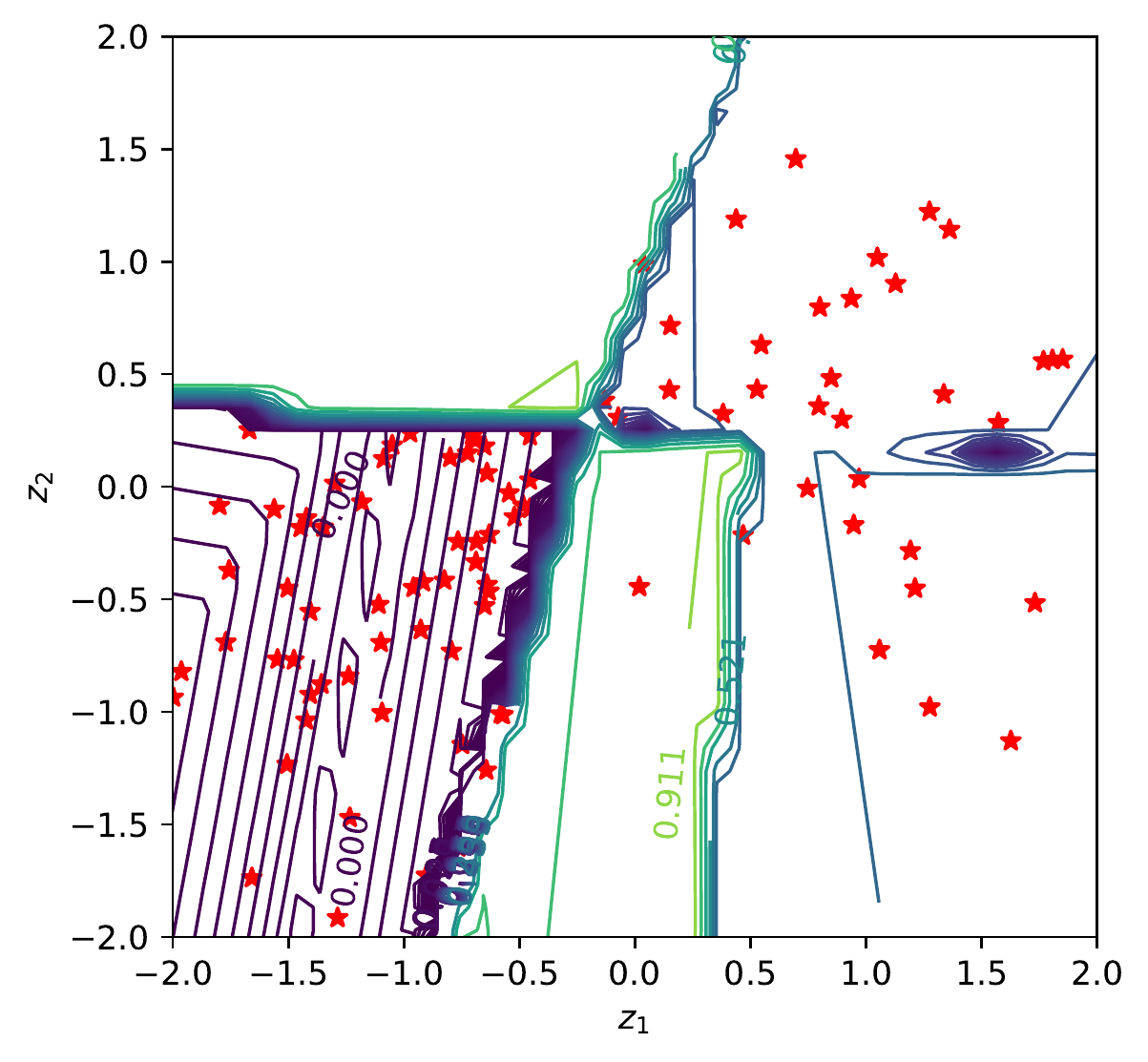}&\includegraphics[width=27mm]{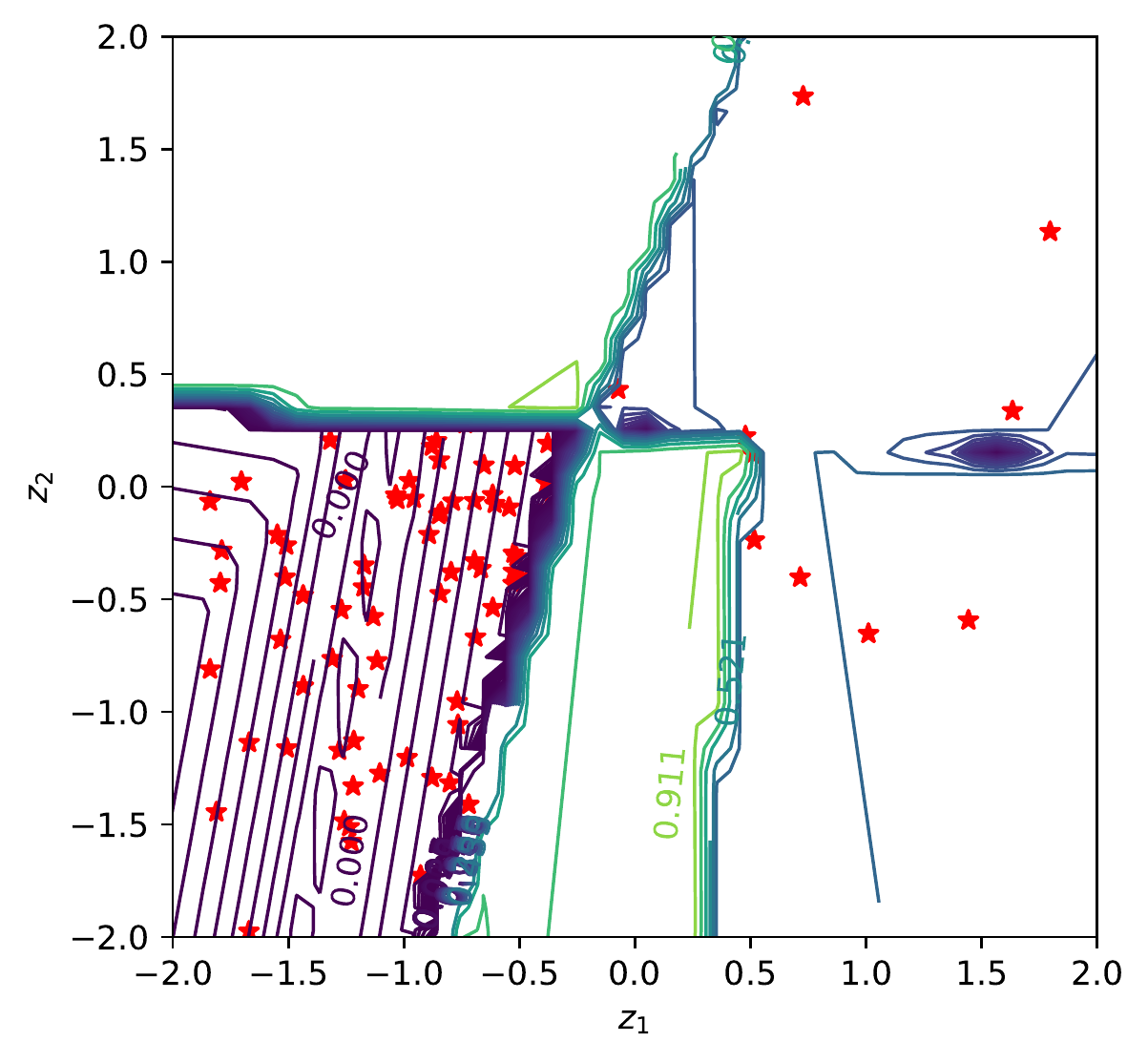}&\includegraphics[width=27mm]{figs/fig3_iter2000.pdf} &\includegraphics[width=27mm]{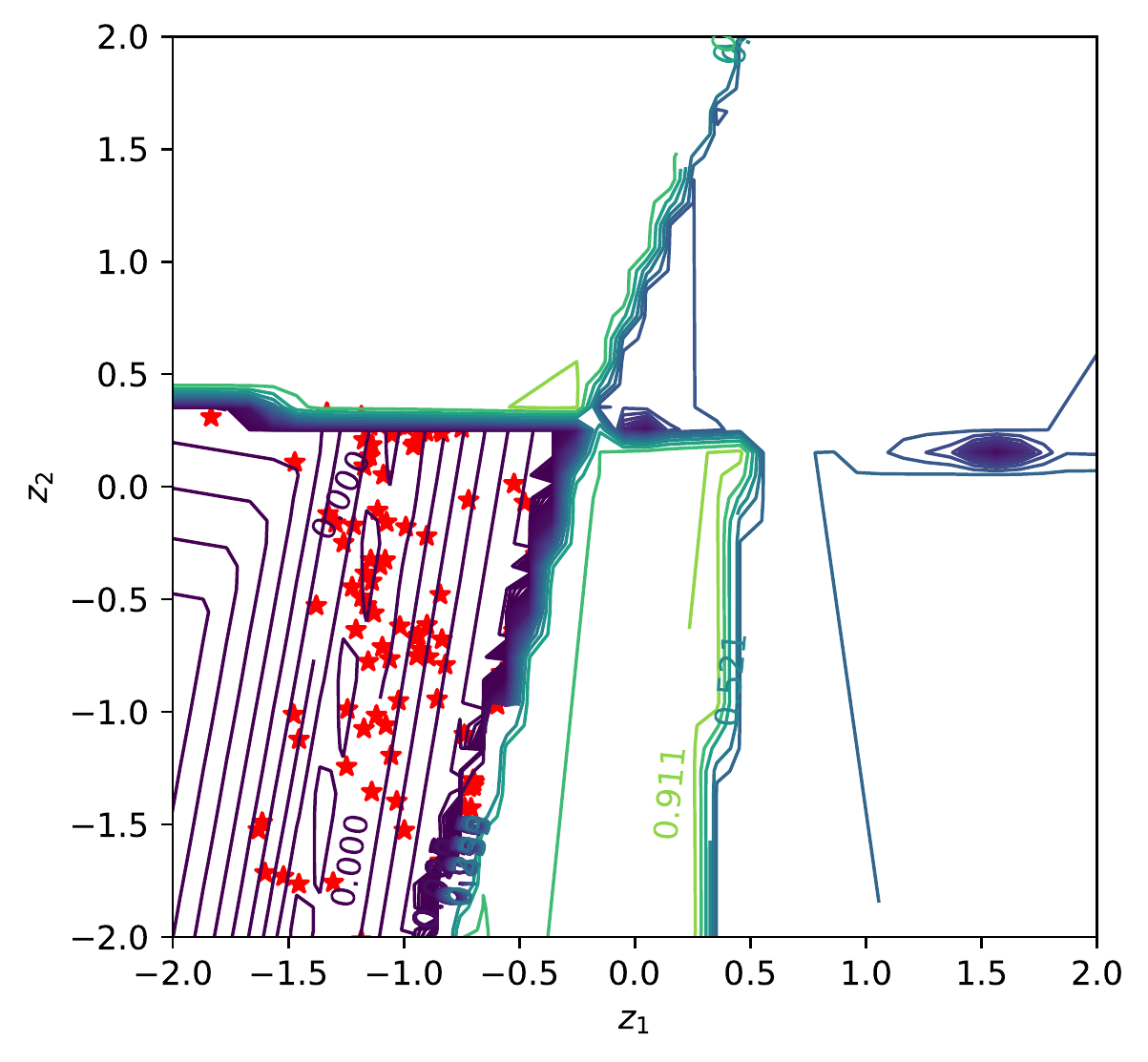} 
  
\end{tabularx}
\caption{$z$ state in $\cZ$ space during the AIS procedure after the 100th, 3700th, 3800th, 3900th and 4000th AIS loop.}
\label{fig:aisprocedure1}
\end{figure}

The optimal solution of this experiment is $(x_1, x_2) = (1, 0)$. \figref{fig:toysgddiff1} shows  energy barriers which challenge optimization \wrt  $z$ particularly for well trained generators. \figref{fig:toycomp1} shows that baselines  get stuck in a local optimum easily. 
In contrast, the reconstructions obtained by our proposed algorithm  overcome the energy barriers and find good solutions within 4,000 AIS loops. We observe that the reconstruction accuracy (\figref{fig:toycomp1} (d, e)) increases with  generator model improvements. \figref{fig:aisprocedure1} shows how 100 $z$ samples move in $\cZ$-space and cross the barriers during the AIS procedure.

\section{Appendix: Additional Real Data Examples}

We show additional results for real data experiments. We observe our proposed algorithm to recover masked images more accurately than baselines and to generate better high-resolution images given low-resolution images. \\

We show masked CelebA  (\figref{fig:celebacomp1}) and LSUN (\figref{fig:lsuncomp1}) recovery results for baselines and our method, given a Progressive GAN generator. Note that our algorithm is pretty robust to the position of the $z$ initialization, since the generated results are consistent in \figref{fig:celebacomp1}. In \figref{fig:celebahqcomp1}, we compare our results with baselines on generating a $1024\times1024$ high-resolution image from a given $128\times128$ low-resolution image.

\begin{figure}[t]
\centering

\begin{tabular}{cc} 
  \includegraphics[width=40mm]{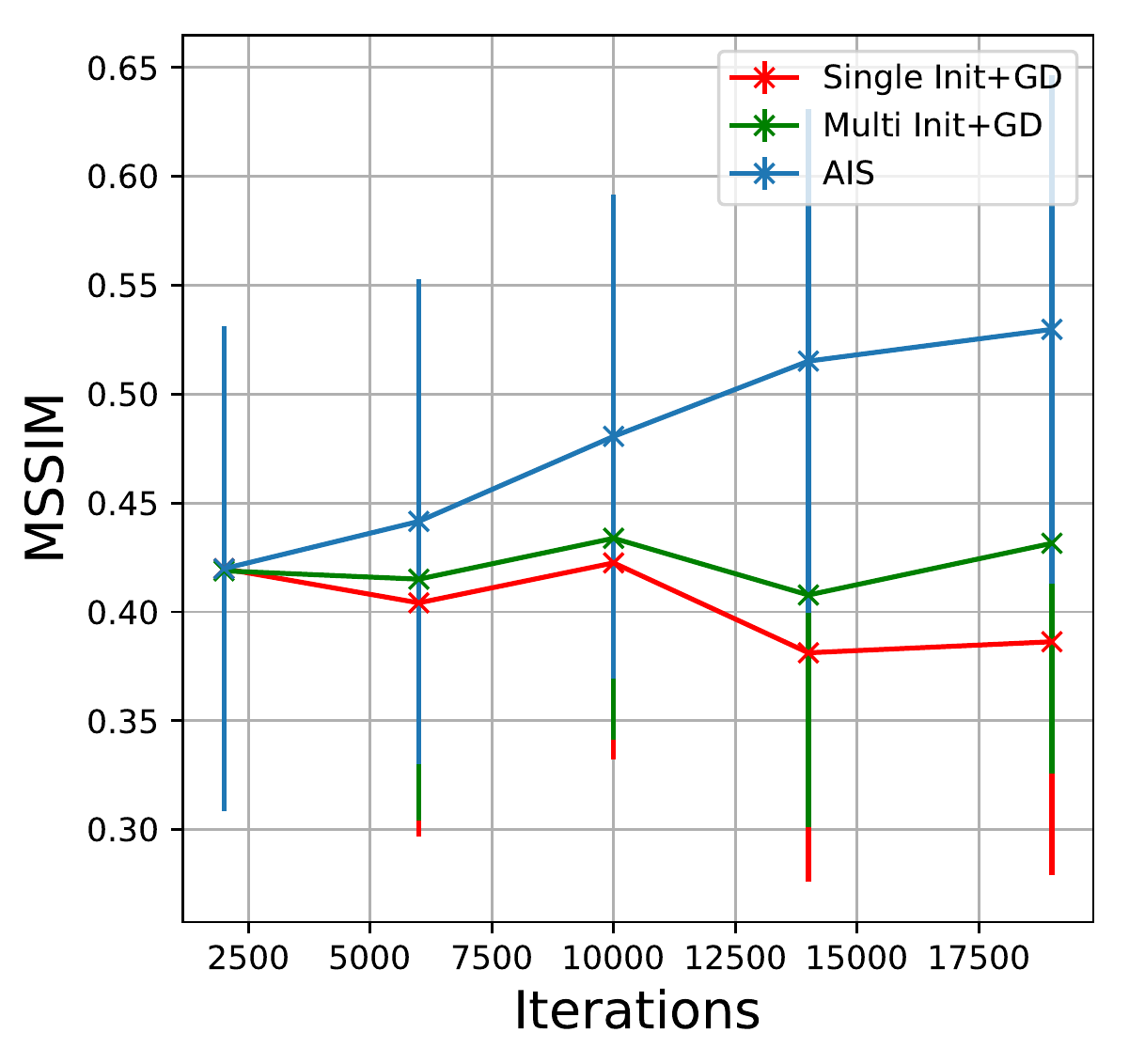}&\includegraphics[width=40mm]{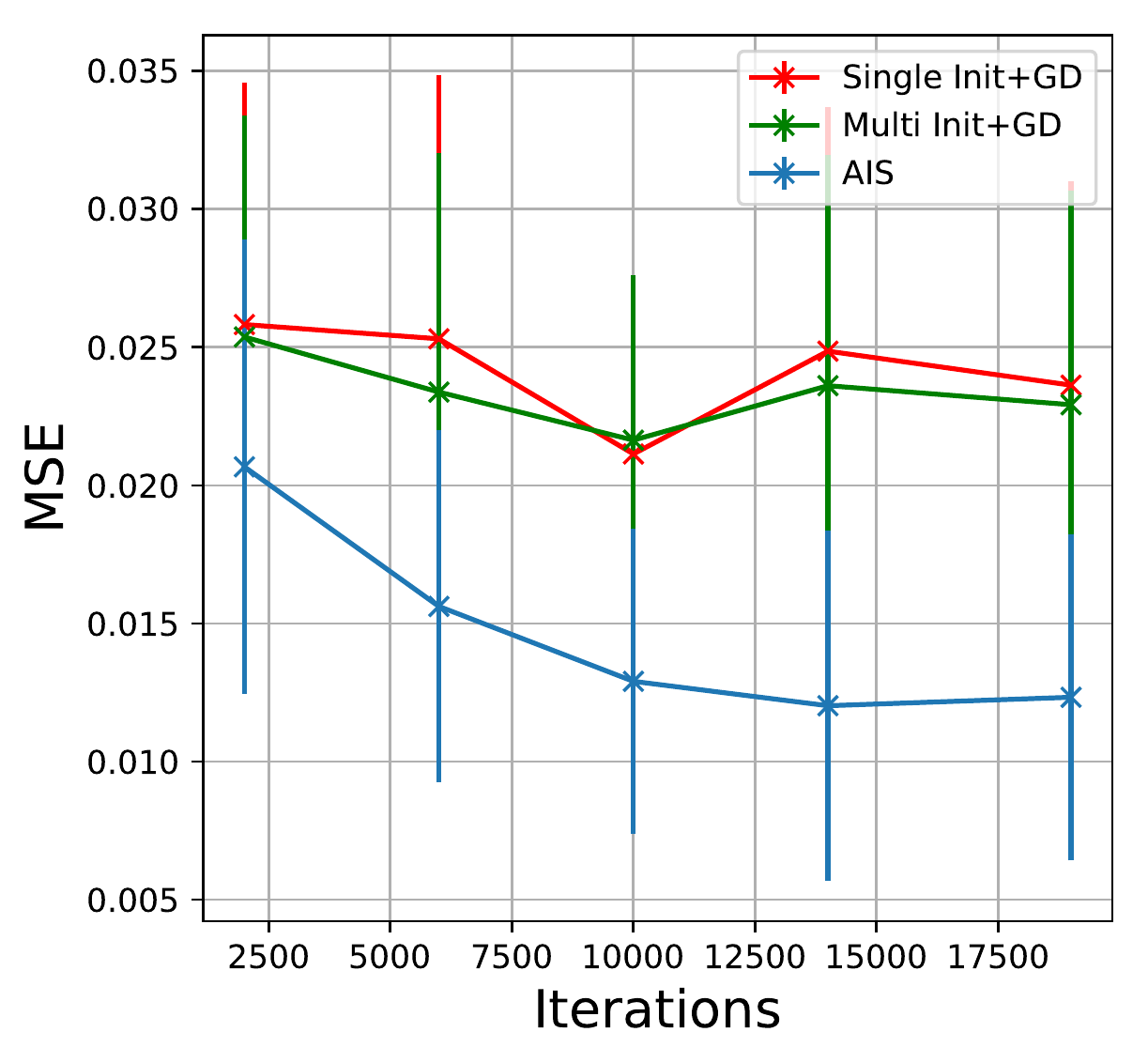}\\
  \multicolumn{1}{c}{{\small (a)}} & \multicolumn{1}{c}{{\small (b)}} 
\end{tabular}
\vspace{-0.3cm}
\caption{Reconstructions errors over the number of progressive GAN training iterations. (a) MSSIM on LSUN test data; (b) MSE on LSUN test data.}
\label{fig:lsuntestmetric}
\vspace{-0.5cm}
\end{figure}

\begin{figure}[t]
\centering
 \begin{tabularx}{\textwidth} {ccc}
   \multicolumn{3}{c}{\includegraphics[width=120mm]{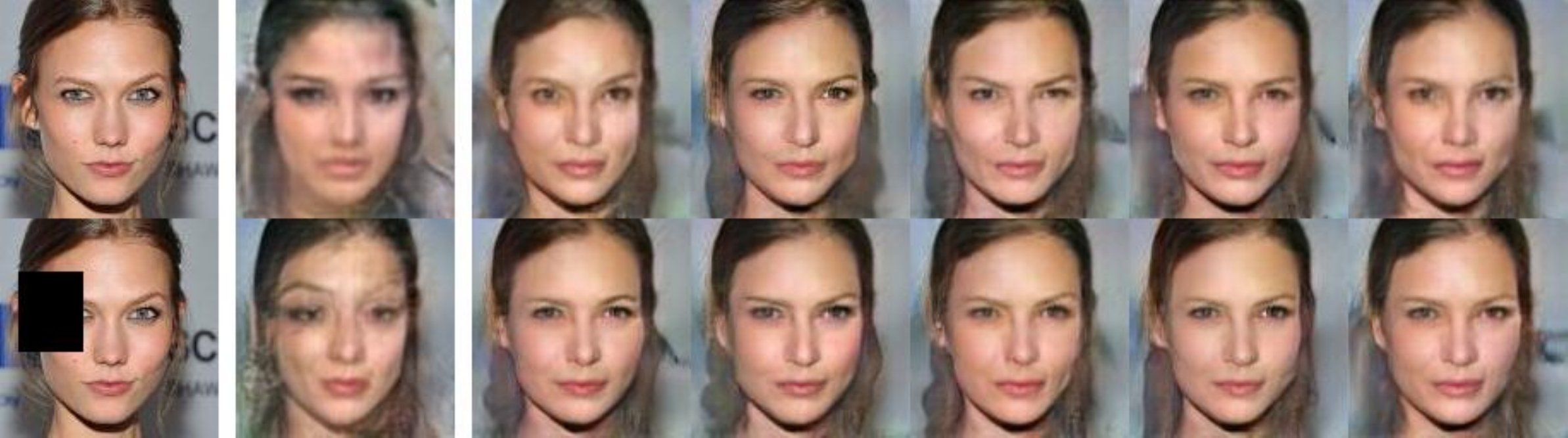}}\\
   \multicolumn{3}{c}{\includegraphics[width=120mm]{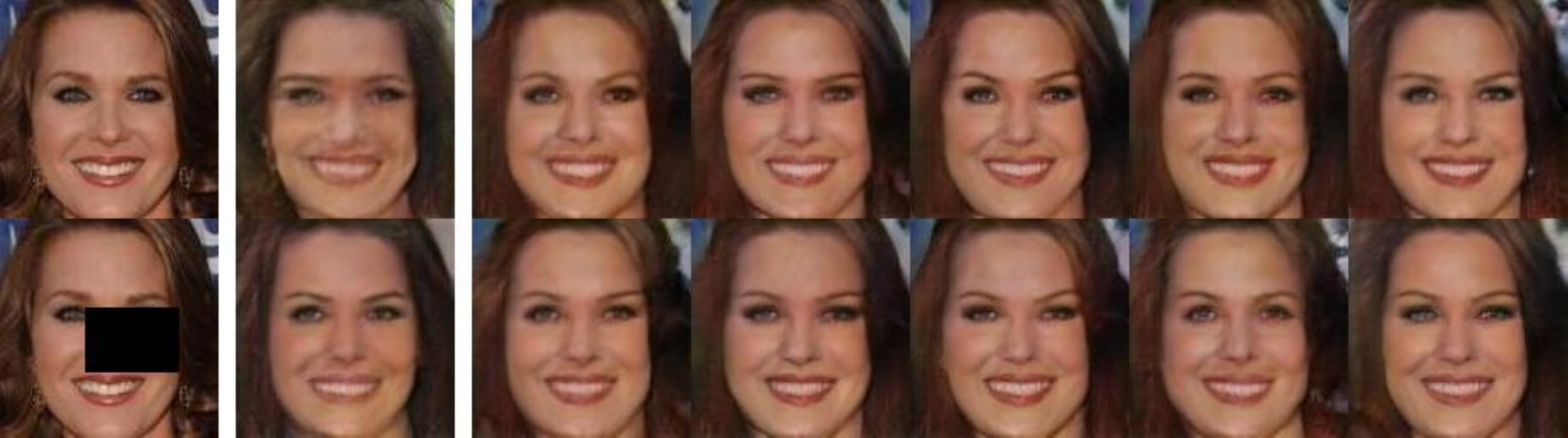}}\\
   \multicolumn{3}{c}{\includegraphics[width=120mm]{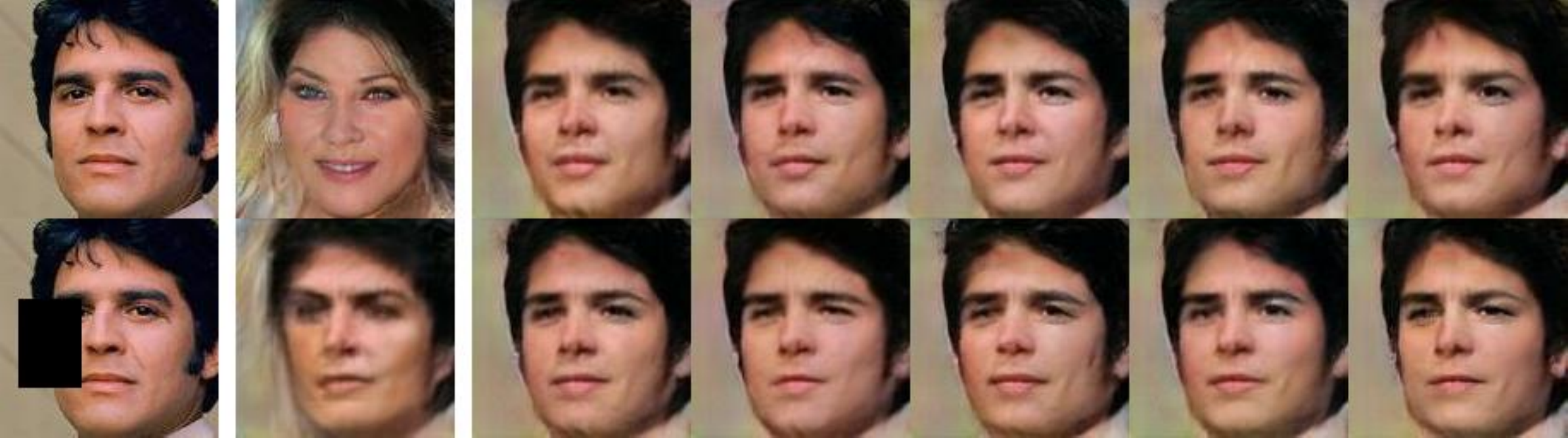}} \\
   \multicolumn{3}{c}{\includegraphics[width=120mm]{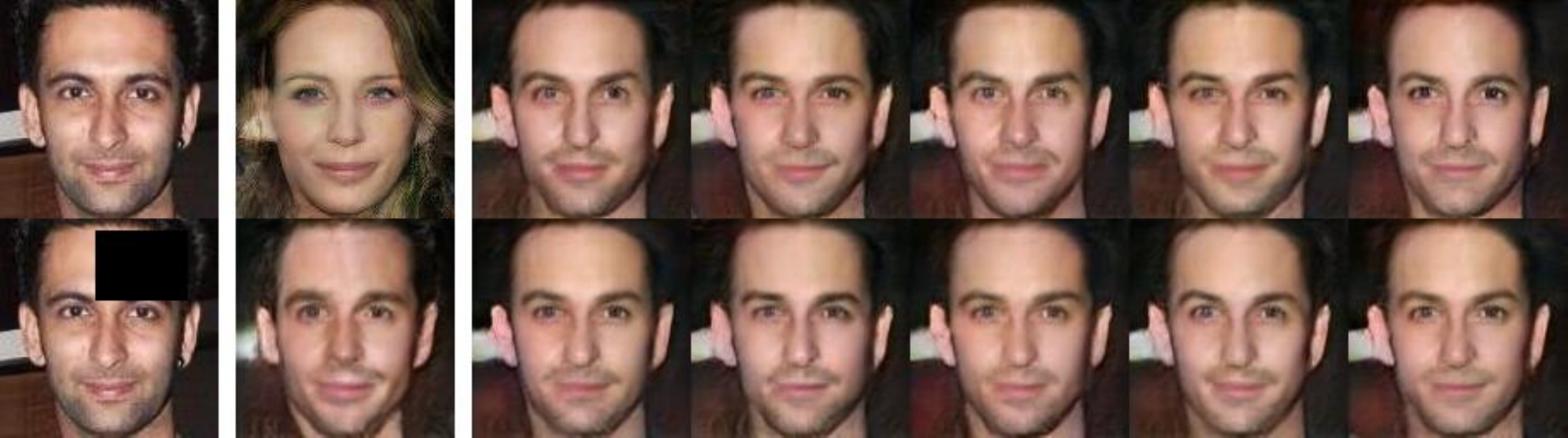}}\\
   \hspace{0.6cm}(a)\hspace{1cm} & (b) & (c)
\end{tabularx}
\caption{Reconstructions on 128$\times$128 CelebA images for a trained progressive GAN at 10k-th iteration. (a) Ground truth and masked (observed) images (top to bottom); (b)  The result obtained by optimizing  the best $z$ picked from 5,000 initializations (top to bottom); (c) Result generated by our algorithm.}
\label{fig:celebacomp1}
\end{figure}

\begin{figure}[t]
\centering
 \begin{tabularx}{\textwidth} {ccccc}
  \multicolumn{5}{c}{\includegraphics[width=120mm]{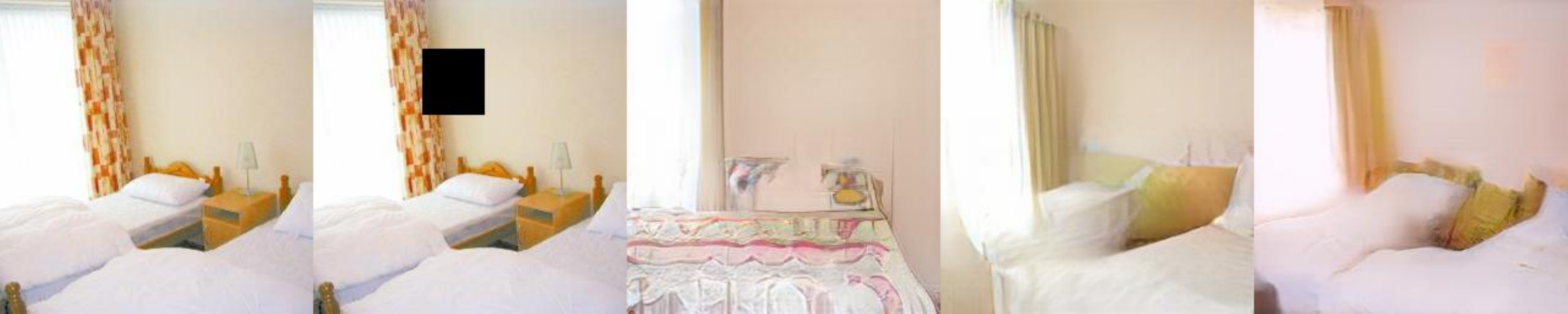}} \\
   \multicolumn{5}{c}{\includegraphics[width=120mm]{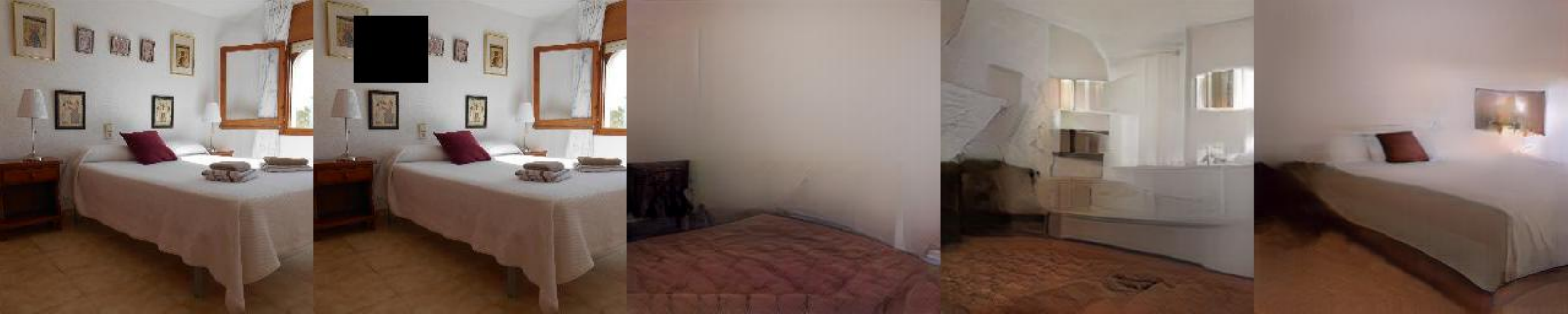}}\\
   \multicolumn{5}{c}{\includegraphics[width=120mm]{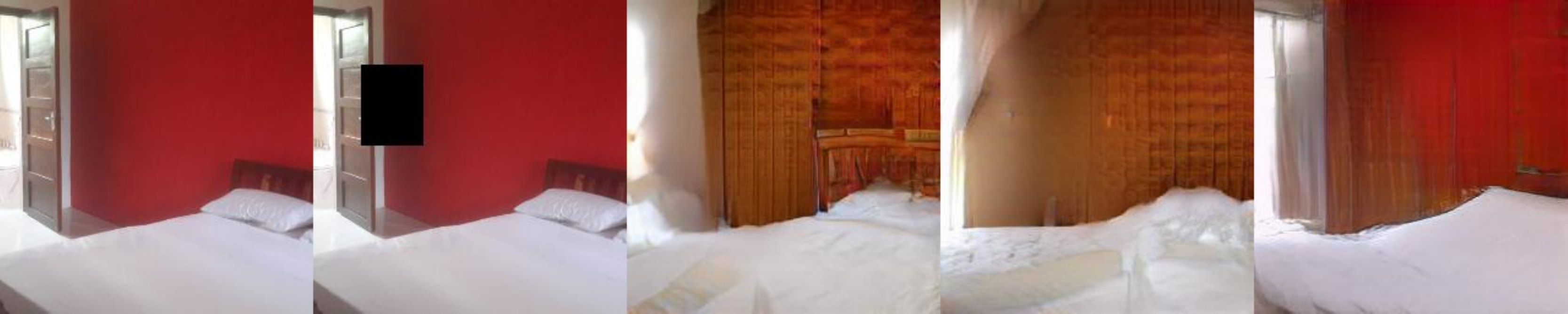}}\\
   \multicolumn{5}{c}{\includegraphics[width=120mm]{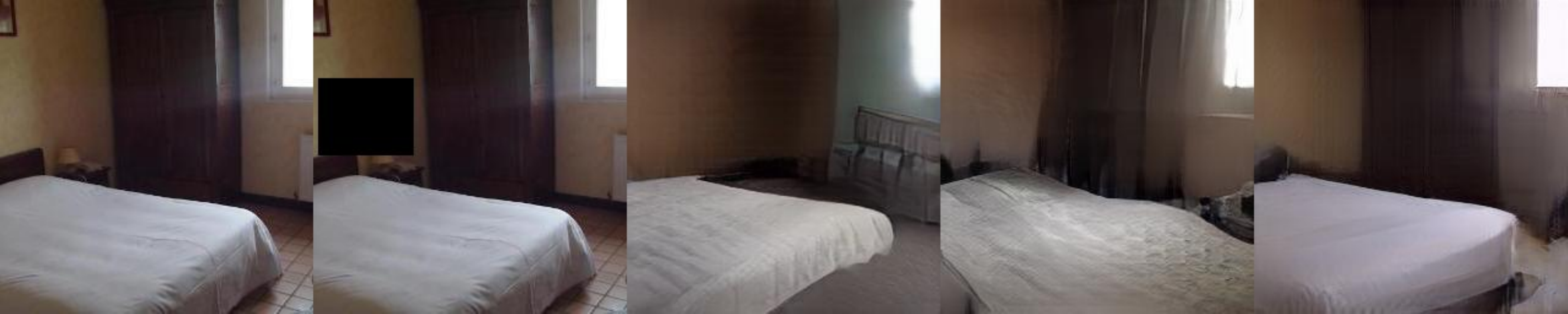}}\\
   \multicolumn{5}{c}{\includegraphics[width=120mm]{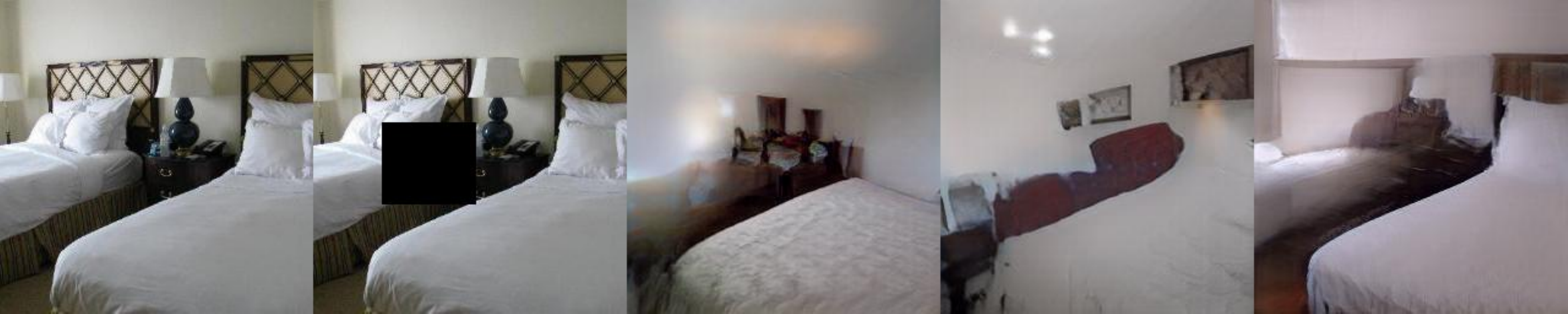}}\\
   \multicolumn{5}{c}{\includegraphics[width=120mm]{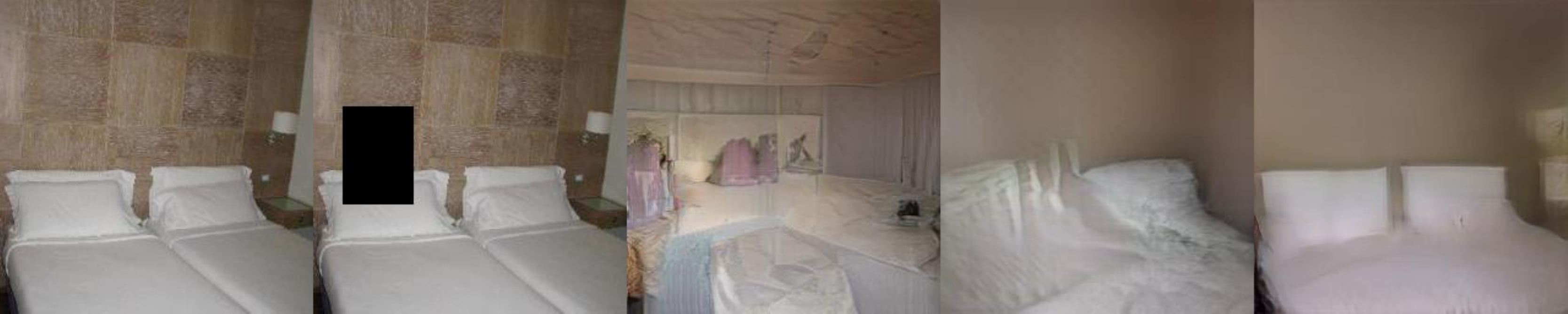}}\\
   \hspace{1cm}(a)\hspace{1.5cm} & (b)\hspace{1.6cm} & (c)\hspace{1.6cm} & (d)\hspace{0.6cm} & \hspace{0cm}(e) \\
\end{tabularx}
\caption{Reconstructions on $256\times 256$ LSUN images for a trained progressive GAN at 10k-th iteration. (a) Ground truth; (b) Masked (observed) images; (c) The result obtained whenn optimizing a  single $z$; (d)  The result obtained by optimizing  the best $z$ picked from 5,000 initializations; (e) Result of our algorithm. }
\label{fig:lsuncomp1}
\end{figure}

\begin{figure}[t]
\centering
 \begin{tabularx}{\textwidth} {cccc}
  \multicolumn{4}{c}{\includegraphics[width=120mm]{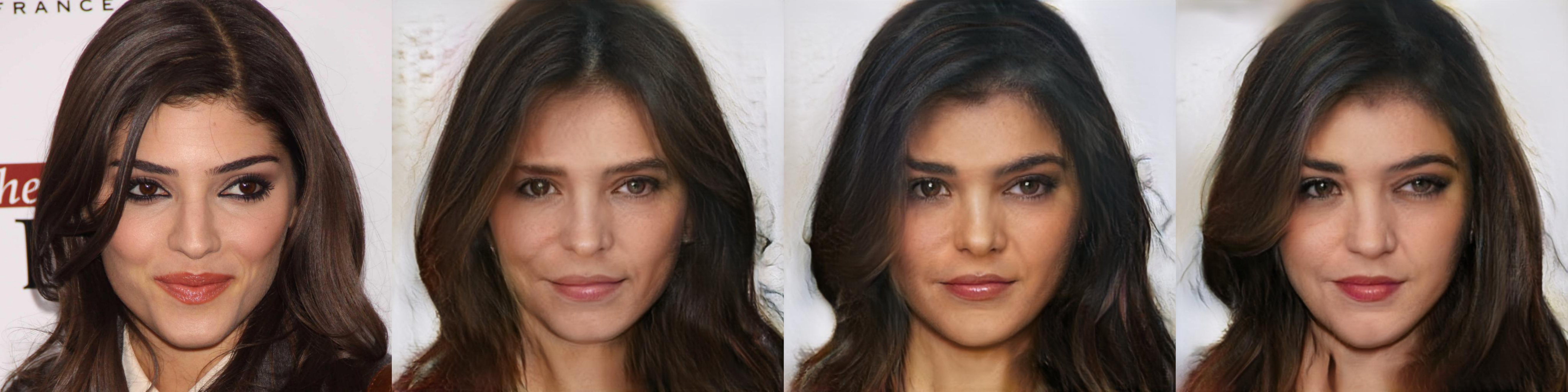}} \\
  \multicolumn{4}{c}{\includegraphics[width=120mm]{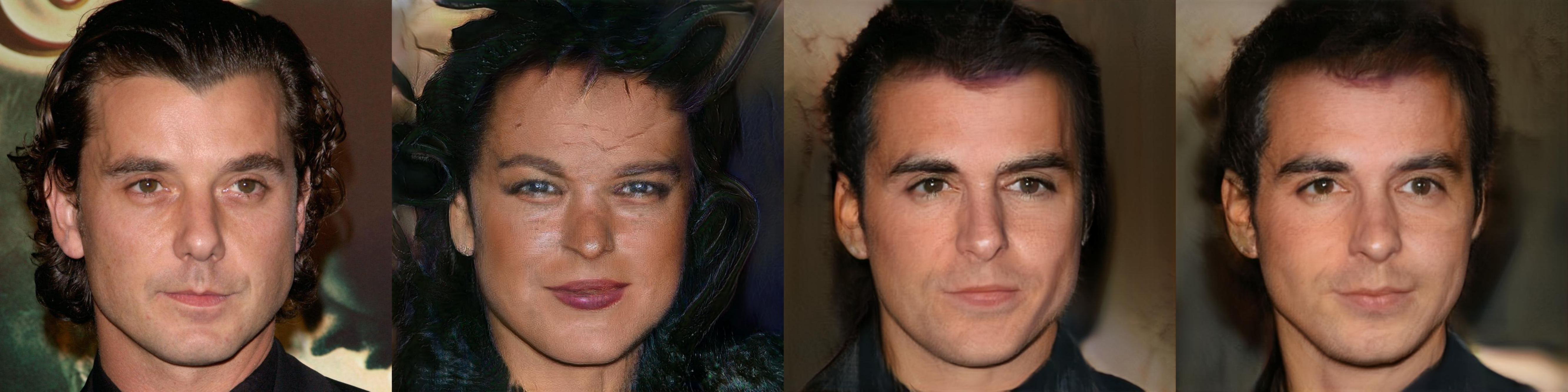}}\\
  \multicolumn{4}{c}{\includegraphics[width=120mm]{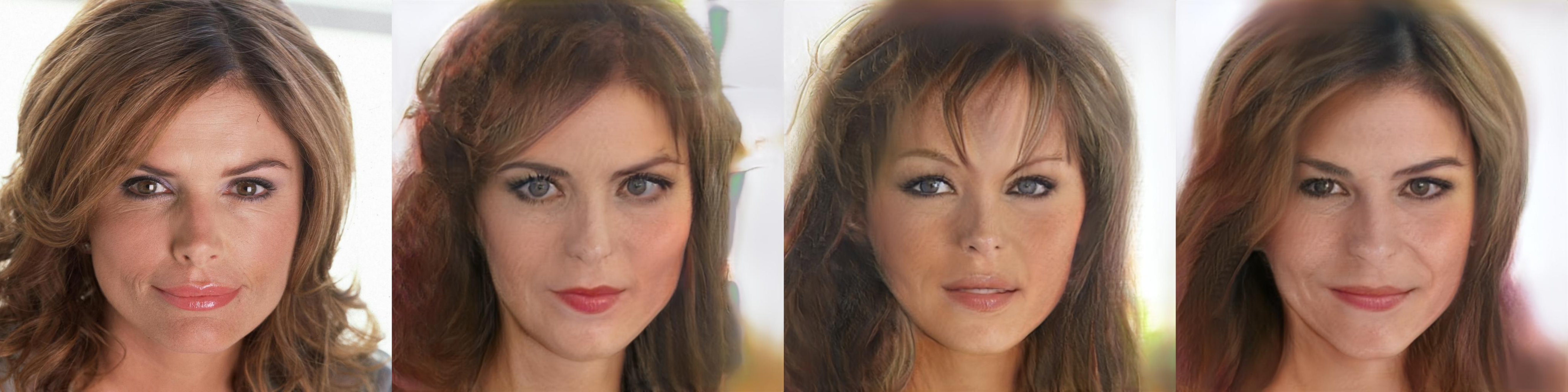}}\\
  \multicolumn{4}{c}{\includegraphics[width=120mm]{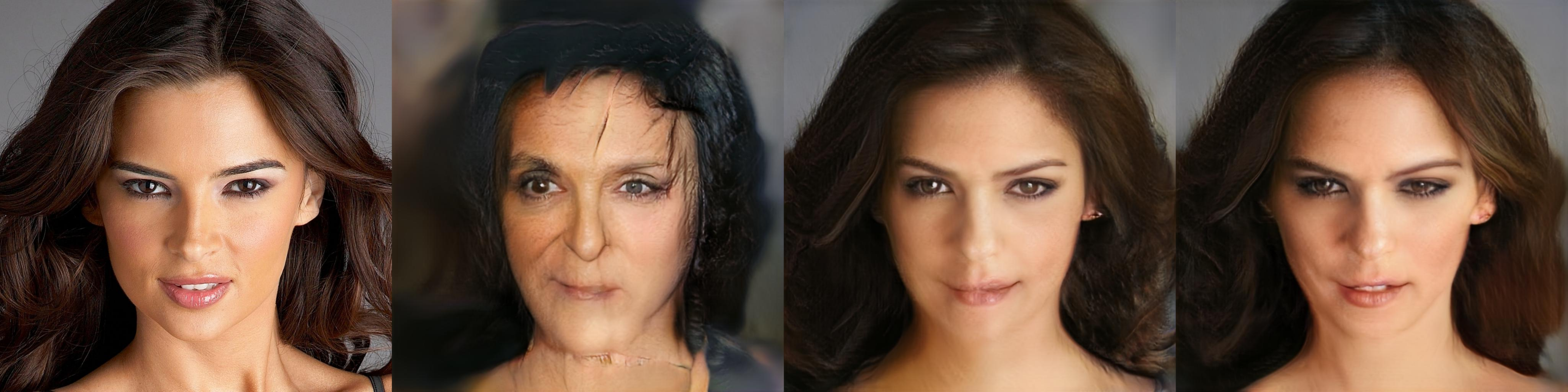}}\\
  \multicolumn{4}{c}{\includegraphics[width=120mm]{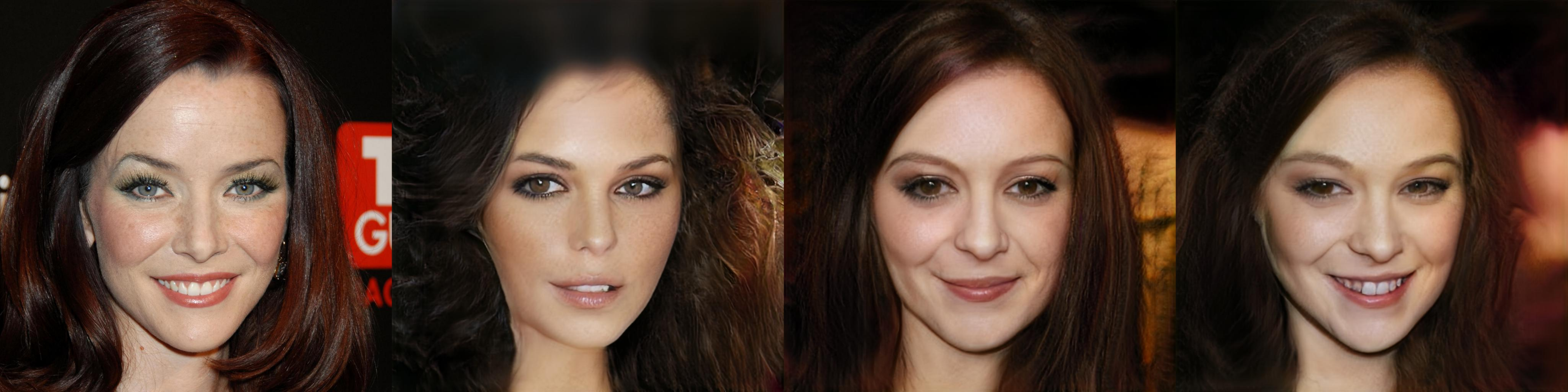}}\\
  \multicolumn{4}{c}{\includegraphics[width=120mm]{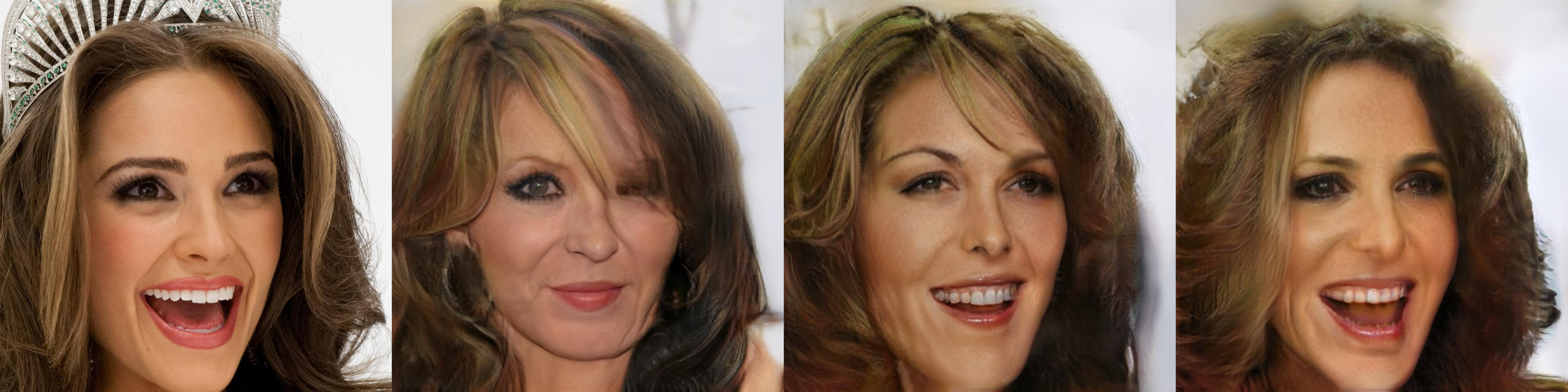}}\\
    \hspace{1.3cm}(a)\hspace{2.2cm} & (b)\hspace{2.2cm} & (c)\hspace{1cm} & (d) \\
\end{tabularx}
\caption{Simple Super Resolution Task result from  a 128$\times$128 to a  1024$\times$1024 image for a trained progressive GAN at 19k-th iteration. (a) Ground truth; (b) The result obtained  by optimizing a single $z$; (c) The result obtained by optimizing  the best $z$ picked from 5,000 initializations; (d) Results of our algorithm. }
\label{fig:celebahqcomp1}
\end{figure}

We also run both baselines and AIS on LSUN test data. The result is shown in \figref{fig:lsuntestmetric}. However, for both CelebA and CelebA-HQ, Progressive GANs are trained on the whole dataset, we are unable to do the experiments on these two datasets.

\end{document}